\RequirePackage[svgnames]{xcolor}

\documentclass[11pt,letterpaper]{mystyle}

\usepackage[all]{hypcap}
\usepackage[svgnames]{xcolor}
\usepackage[comma,authoryear,compress]{natbib}
\usepackage{hyperref}[citecolor=lightblue]

\hypersetup{
    colorlinks = true,
    citecolor = {YaleBlue},
}

\usepackage{algorithm}
\usepackage{algorithmicx}
\usepackage{algpseudocode}
\usepackage{microtype}
\usepackage{graphicx}
\expandafter\def\csname ver@subfig.sty\endcsname{}
\usepackage{booktabs} %
\usepackage{float}
\usepackage{bigstrut}

\usepackage{amsmath}
\usepackage{amssymb}
\usepackage{mathtools}
\usepackage{amsthm}
\usepackage{mathrsfs}
\usepackage{nicefrac}
\usepackage{dsfont}
\usepackage{enumitem}
\usepackage{subcaption}
\usepackage{graphicx,subfig}
\usepackage{cleveref}
\usepackage{bxcoloremoji}

\usepackage{float}

\usepackage[utf8]{inputenc} %
\usepackage[T1]{fontenc}    %
\usepackage{hyperref}       %
\usepackage{url}            %
\usepackage{booktabs}       %
\usepackage{amsfonts}       %
\usepackage{nicefrac}       %
\usepackage{microtype}      %
\usepackage{graphicx}
\usepackage{subcaption}
\usepackage{amssymb}
\usepackage{fdsymbol}
\usepackage{wrapfig}
\usepackage{lipsum}
\usepackage{enumitem}
\usepackage{stackengine}
\usepackage[font=small,labelfont=bf]{caption}
\usepackage{color}
\usepackage{adjustbox}

\usepackage{rotating}
\usepackage{makecell}

\newcommand{\x}{\mathbf{x}}

\DeclareMathOperator*{\E}{\mathbb{E}}

\definecolor{blanchedalmond}{rgb}{1.0, 0.92, 0.8}
\definecolor{carmine}{rgb}{0.59, 0.0, 0.09}
\definecolor{lightblue}{rgb}{0.22,0.45,0.70}%

\renewcommand{\mathbf}{\boldsymbol}

\newcommand{\msf}{\mathsf}

\makeatletter
\def\Ddots{\mathinner{\mkern1mu\raise\p@
\vbox{\kern7\p@\hbox{.}}\mkern2mu
\raise4\p@\hbox{.}\mkern2mu\raise7\p@\hbox{.}\mkern1mu}}
\makeatother

\newcommand{\clip}{\msf{clipped}}

\definecolor{amaranth}{rgb}{0.9, 0.17, 0.31}
\definecolor{antiquebrass}{rgb}{0.8, 0.58, 0.46}
\definecolor{antiquefuchsia}{rgb}{0.57, 0.36, 0.51}
\definecolor{chromeyellow}{rgb}{0.31, 0.47, 0.26}

\newcommand{\github}{\raisebox{-1.5pt}{\includegraphics[height=1.05em]{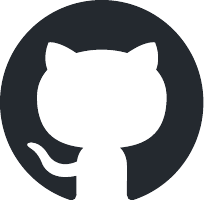}}}
\newcommand{\wnb}{\raisebox{-1.5pt}{\includegraphics[height=1.05em]{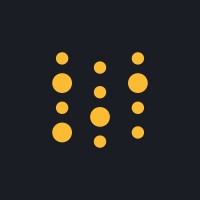}}}
\newcommand{\tina}{\raisebox{-1.5pt}{\includegraphics[height=1.05em]{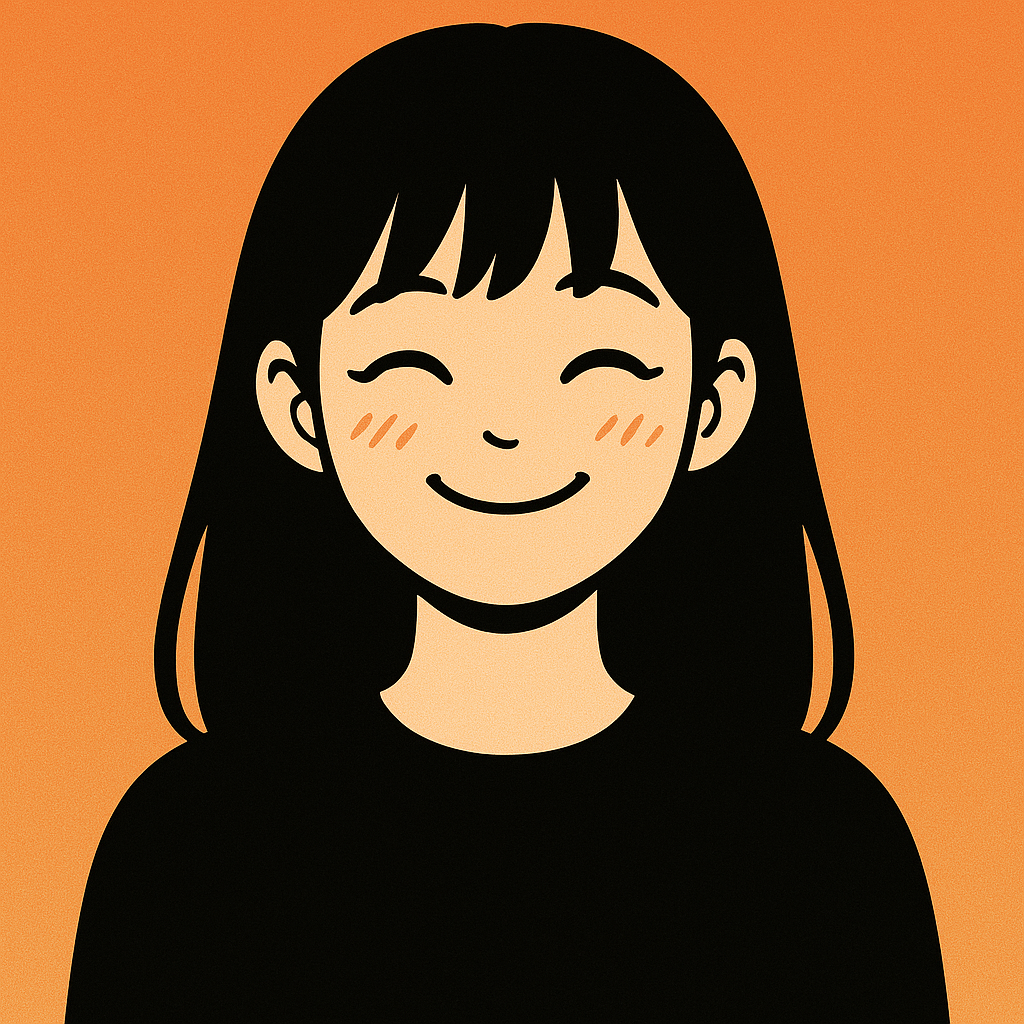}}}

\newtcolorbox{AIbox}[2][]{aibox,title=#2,#1}
\definecolor{lightblue}{rgb}{0.22,0.45,0.70}%
\definecolor{Gray}{gray}{0.95}
\definecolor{Cornsilk}{rgb}{1.0, 0.97, 0.86}

\usepackage{amsmath}

\usepackage[all]{hypcap}

\title{{\textbf{Tina}}: Tiny Reasoning Models via LoRA}

\runningtitle{\textbf{Tina}: Tiny Reasoning Models via LoRA}

\author{
  Shangshang Wang$^1$,  
  Julian Asilis$^1$, 
  Ömer Faruk Akgül$^1$,
  Enes Burak Bilgin$^1$, \\
  Ollie Liu$^1$, and 
  Willie Neiswanger
}

\affil[1]{University of Southern California}

\correspondingauthor{Shangshang Wang \href{https://shangshang-wang.github.io}{shangshangwang.github.io}; Willie Neiswanger \href{mailto:neiswang@usc.edu}{neiswang@usc.edu}}

\begin{document}

\begin{abstract}
How cost-effectively can strong reasoning abilities be achieved in language models? Driven by this fundamental question, we present Tina, a family of \textit{tiny} reasoning models achieved with high cost-efficiency.
Notably, Tina demonstrates that substantial reasoning performance can be developed using only minimal resources, by applying parameter-efficient updates during reinforcement learning (RL), using low-rank adaptation (LoRA), to an already \textit{tiny} 1.5B parameter base model.
This minimalist approach produces models that achieve reasoning performance which is competitive with, and sometimes surpasses, SOTA RL reasoning models built upon the same base model. Crucially, this is achieved at a \textit{tiny} fraction of the computational post-training cost employed by existing SOTA models. In fact, the best Tina model achieves a >20\% reasoning performance increase and 43.33\% Pass@1 accuracy on AIME24, at only \$9 USD post-training and evaluation cost (\emph{i.e.,} an estimated 260x cost reduction). Our work reveals \textit{the surprising effectiveness of efficient RL reasoning via LoRA}. We validate this across multiple open-source reasoning datasets and various ablation settings starting with a single, fixed set of hyperparameters. Furthermore, we hypothesize that this effectiveness and efficiency stem from LoRA rapidly adapting the model to the structural format of reasoning rewarded by RL, while largely preserving the base model's underlying knowledge. In service of accessibility and open research, we fully open-source all code, training logs, and model weights \& checkpoints.
\vspace{5mm}

\tina{} \textbf{Notion Blog}: \href{https://shangshangwang.notion.site/tina}{https://shangshangwang.notion.site/tina}

\github{} \textbf{Code Repository}: \href{https://github.com/shangshang-wang/Tina}{https://github.com/shangshang-wang/Tina}

\wnb{} \textbf{Training Logs}: \href{https://wandb.ai/upup-ashton-wang-usc/Tina}{https://wandb.ai/upup-ashton-wang-usc/Tina}

\coloremojicode{1F917} \textbf{Model Weights \& Checkpoints}: \href{https://huggingface.co/Tina-Yi}{https://huggingface.co/Tina-Yi}
\end{abstract}

\maketitle
\vspace{3mm}
\vspace{-4mm}
\section{Introduction}

Language models (LMs) demonstrate increasing proficiency across a variety of tasks, but achieving robust, multi-step reasoning remains a frontier challenge~\citep{wangllmreasoning2025, xu2025towards}. Notably, such reasoning abilities are crucial for applications demanding complex problem-solving, from scientific discovery to intricate planning. Enhancing complex reasoning via supervised fine-tuning (SFT) is a well-adopted technique, often utilizing a distillation process~\citep{min2024imitateexploreselfimprovereproduction, huang2024o1replicationjourney} by which the model learns to mimic reasoning traces (\emph{e.g.,} step-by-step thinking) generated by more advanced models such as o1~\citep{openai2024openaio1card}. This approach, while effective, relies upon the quality and availability of such expert demonstrations, which can be costly to obtain. Furthermore, it can run the risk of instilling a shallow form of imitation in the learning model, rather than fostering dynamic exploration of reasoning paths. In contrast, reinforcement learning (RL) enables models to learn directly and flexibly from verifiable reward signals derived from curated data~\citep{deepseekai2025deepseekr1incentivizingreasoningcapability, lambert2025tulu3pushingfrontiers}. In doing so, RL can lead the model to explore a greater variety of logical paths and possibly discover more robust solutions. However, RL pipelines are often complex and notoriously resource-intensive, typically involving substantial compute. 
This raises a fundamental question anchoring our research: 
\begin{center}
\emph{
How cost-effectively can one perform RL to efficiently instill reasoning abilities in LMs?
}
\end{center}

\begin{figure}[h]
    \centering
    \includegraphics[width=\linewidth]{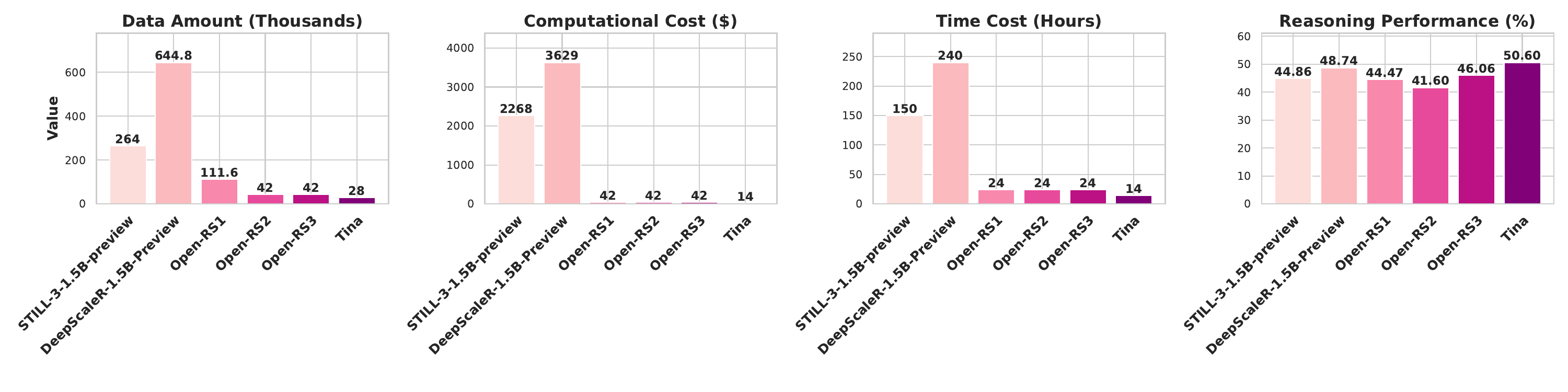}
    \vspace{-10mm}
    \caption{Overall comparison between Tina and baseline models. The Tina model in the figure corresponds to the best checkpoint in Table~\ref{tab:tina_openrs2_eval}. Reasoning performance denotes the average score across AIME24/25, AMC23, MATH500, GPQA, and Minerva, as described in Section~\ref{sec:training}. The calculation of each comparative metric is detailed in Appendix~\ref{sec:cost_breakdown}.}
    \label{fig:overall_comparison}
\end{figure}

Our pursuit of this question necessitates a deliberate move towards minimalism. Rather than utilizing models with tens of billions of parameters 
(such as \texttt{Qwen-7B/32B}, \texttt{QwQ-32B-preview}, and their variants~\citep{min2024imitateexploreselfimprovereproduction, sky_t1_2025, zeng2025simplerlzooinvestigatingtamingzero, muennighoff2025s1simpletesttimescaling, cui2025process, lyu2025exploring, openthoughts, OpenReasonerZero2025}), we instead direct our attention to tiny models. In particular, we use the 1.5B parameter model, \texttt{DeepSeek-R1-Distill-Qwen-1.5B}~\citep{deepseekai2025deepseekr1incentivizingreasoningcapability}. Our choice of this base model aligns with common practices in recent research~\citep{Slow_Thinking_with_LLMs_3_Preview, deepscaler2025, dang2025reinforcementlearningreasoningsmall}: we begin with a foundation that, owing to its specific lineage (DeepSeek/Qwen) and distillation process, likely possesses stronger initial reasoning aptitude compared to a generic pre-trained model of equivalent size. This strategic starting point allows us to more-rigorously evaluate the incremental reasoning enhancements imparted by RL, thereby isolating and measuring the effectiveness of the technique itself over a competent baseline.
More importantly, selecting such an architecture dramatically lowers the computational and financial threshold for experimentation. Complementing the choice of a compact base model, we further amplify efficiency during the RL phase and integrate parameter-efficient post-training by employing low-rank adaptation (LoRA)~\citep{hu2021loralowrankadaptationlarge}. Notably, LoRA enables the modification of a model's behavior by training only an exceptionally small number of new parameters. This dovetails with our central motivation: achieving reasoning capabilities through the most economical means possible.

Integrating the previous two components---a ``tiny'' model architecture and a ``tiny'' post-training via LoRA-based RL---we release the Tina (\underline{Tin}y Reasoning Models via LoR\underline{A}) family of models, which attain substantial reasoning performance at strikingly low cost. In total, we summarize our contributions as follows: 
\begin{itemize}[leftmargin=7mm,itemsep=2mm, topsep=0em]
    \item \textbf{Surprising Effectiveness of Efficient RL Reasoning.} We show that our Tina models achieve performance competitive with, and in some cases even superior to, SOTA baseline models built on the same base model with full-parameter training, as shown in Figure~\ref{fig:overall_comparison} and in more detail in Table~\ref{tab:tina_eval}. In particular, the best Tina model achieves a >20\% performance increase and 43.33\% Pass@1 accuracy on AIME24.
    
    \item \textbf{Rapid Reasoning Format Adaptation Hypothesis.} Based on our observations in post-training Tina, we hypothesize that LoRA's effectiveness and efficiency stem from rapidly adapting the reasoning format under RL while preserving base model knowledge—a likely more compute-efficient process than the deep knowledge integration of full-parameter training. Partial support comes from studies showing tiny LMs can reason effectively~\citep{openr1,deepseekai2025deepseekr1incentivizingreasoningcapability}, while large LMs can store broader world knowledge~\citep{allen-zhu2025physics}. This distinction suggests reasoning capabilities can be significantly enhanced by focusing on adapting the output format itself, consistent with our hypothesis about LoRA. To test this, we exclusively train LoRA parameters in RL settings, focusing on leveraging this format adaptation mechanism.
    
    \item \textbf{Democratizing RL Reasoning.} We provide a reproducible and highly cost-effective approach, enabling wider participation in the exploration of RL techniques without requiring extensive computational resources. Notably, the cost of reproducing the best Tina checkpoint stands at only \textbf{\$9}, and of reproducing all our experiments and everything presented in this paper \textit{from scratch} at \textbf{\$526}. Furthermore, in line with our goal of promoting accessible research, we release all code, training logs, evaluation scripts, and all Tina checkpoints.
\end{itemize}

\label{sec:intro}
\vspace{-1mm}

\section{Related Work}
\label{sec:related_work}

\begin{figure}
    \centering
    \includegraphics[width=.95\linewidth]{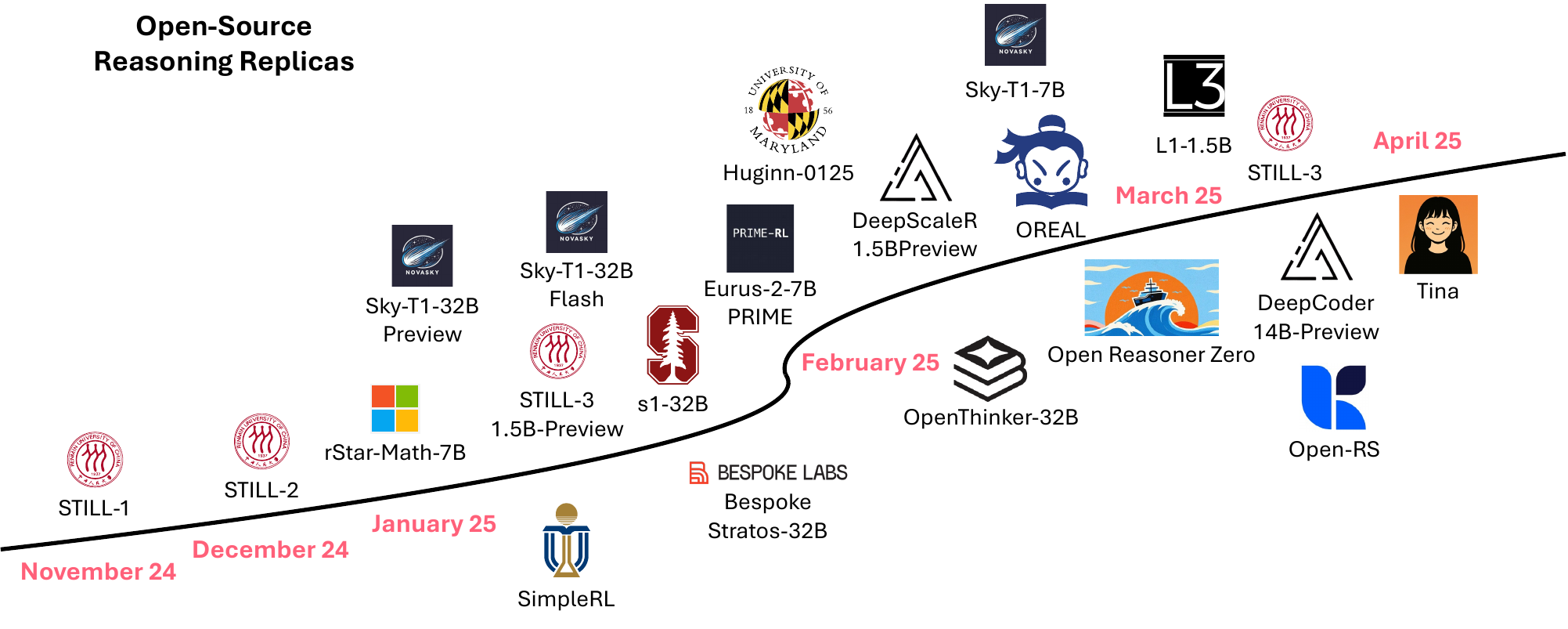}
    \caption{Release timeline of open-source models that aim to replicate the performance of advanced reasoning models like o1(-preview)~\citep{openai2024openaio1card} and R1~\citep{deepseekai2025deepseekr1incentivizingreasoningcapability}, which we refer to as open-source reasoning replicas.}
    \label{fig:timeline}
\end{figure}

\subsection{Open-Source Reasoning Replicas}
As shown in Figure~\ref{fig:timeline}, following the release of o1-preview~\citep{openai2024openaio1card}, a number of open-source models have emerged to replicate or exceed its reasoning capabilities. STILL~\citep{min2024imitateexploreselfimprovereproduction} introduced a minimal yet high-quality training recipe designed to elicit reasoning with modest compute, demonstrating that imitation learning from curated traces remains competitive. Sky-T1~\citep{sky_t1_2025} further explored scaling using open instruction-tuned checkpoints, while SimpleRL~\citep{zeng2025simplerlzooinvestigatingtamingzero} highlighted the potential of lightweight RL without requiring large-scale reward models. PRIME~\citep{cui2025process} and DeepScaleR~\citep{deepscaler2025} introduced process supervision and scaling experiments to isolate how reasoning quality evolves with model size and context length. s1~\citep{muennighoff2025s1simpletesttimescaling} showed that even strong base models such as Qwen2.5-32B-Instruct benefit from fine-tuning on only 1k high-quality and long chain-of-thought data, which is curated to elicit reasoning capabilities. 
L1~\citep{aggarwal2025l1controllinglongreasoning} combined prompt engineering with data curation for RL, resulting in models that can efficiently and adaptively control their response length. Meanwhile, OREAL~\citep{lyu2025exploring} and OpenThinker~\citep{openthoughts} investigated self-correction and latent structure emergence through unsupervised and hybrid paradigms. The release of Open Reasoner Zero~\citep{OpenReasonerZero2025} and Open-RS~\citep{dang2025reinforcementlearningreasoningsmall} further emphasized efficient RL-based strategies for reasoning with small models, completing a landscape of public alternatives for interpretability and reproducibility.

\subsection{RL with Verifiable Rewards}
Reasoning tasks are well-suited to RL paradigms, as the correctness or quality of the final output often provides verifiable reward signals (\emph{e.g.}, the validity of a logical deduction). Such signal can effectively guide the model towards learning more robust reasoning strategies. Consequently, various RL approaches have been explored within this domain. Certain methods introduce auxiliary reward models or critics to assess reasoning quality, such as ReFT~\citep{luong2024reftreasoningreinforcedfinetuning} and REFINER~\citep{paul2024refiner}. Other techniques employ explicit rule-based verification for self-correction~\citep{wu2024large}. Some leverage self-play dynamics and exploration, such as mutual reasoning~\citep{qi2024mutual}, or utilize inference-aware fine-tuning that optimizes performance under different sampling strategies~\citep{chow2024inference}. Notably, Group Relative Policy Optimization (GRPO) has been proposed as a variant of Proximal Policy Optimization (PPO) which removes the need for a separate value network by using a group-based baseline for advantage estimation,  improving training efficiency and leading to better reward alignment \citep{shao2024deepseekmathpushinglimitsmathematical}, as demonstrated by DeepSeek-R1~\citep{deepseekai2025deepseekr1incentivizingreasoningcapability}. Subsequently, Dr.GRPO~\citep{liu2025understandingr1zeroliketrainingcritical} introduced a subtle modification of GRPO addressing its bias to produce long responses. For completeness, we provide the standard formulation of GRPO in Appendix~\ref{app:tina_background}. 

\subsection{Low-Rank Adaptation}
While most existing open models that enable reasoning rely on the more expensive full-parameter training~\citep{min2024imitateexploreselfimprovereproduction, sky_t1_2025, zeng2025simplerlzooinvestigatingtamingzero, muennighoff2025s1simpletesttimescaling, aggarwal2025l1controllinglongreasoning, cui2025process, deepscaler2025, lyu2025exploring, openthoughts, OpenReasonerZero2025, dang2025reinforcementlearningreasoningsmall}, we investigate the use of LoRA for parameter-efficient post-training of reasoning models~\citep{hu2021loralowrankadaptationlarge}. Our goal is to assess whether updating only a small fraction of parameters can still yield strong reasoning capabilities~\citep{han2024parameterefficientfinetuninglargemodels}. In addition to its computational efficiency, LoRA provides modularity: by training only a low-rank decomposition of the parameter updates, it becomes possible to toggle reasoning behavior without maintaining multiple full model copies. For completeness, we provide the standard formulation of LoRA in Appendix~\ref{app:tina_background}.

\section{Tina: Tiny Reasoning Models via LoRA}
\label{sec:training}

Tina is our family of models created by post-training the \texttt{DeepSeek-R1-Distill-Qwen-1.5B} base model using LoRA during RL (employing a GRPO-style algorithm). The ``Tiny'' designation encapsulates a deliberate focus on minimalism and efficiency across the entire framework. This encompasses not only the tiny base model architecture and the tiny parameter updates enabled by LoRA, but also extends to a tiny overall resource footprint. This minimized footprint is achieved through an efficient training pipeline leveraging accessible open-source datasets and codebase (detailed in Section~\ref{sec:tiny_pipeline}), and requires only minimal hardware and budget resources (described in Section~\ref{sec:tiny_setup}).

\subsection{Training Pipeline: Baselines \& Datasets}
\label{sec:tiny_pipeline}
To facilitate meaningful comparisons and enable precise ablations, we post-train our Tina models via RL using the datasets and setups from publicly available reasoning models. All Tina and baseline models adopt \texttt{DeepSeek-R1-Distill-Qwen-1.5B} as their base model checkpoint with default open-source weights.

\begin{itemize}[leftmargin=7mm,itemsep=2mm, topsep=0em]
    \item \textbf{STILL-3-1.5B-preview}~\citep{Slow_Thinking_with_LLMs_3_Preview} is a slow-thinking reasoning model developed through iterative RL on a curated dataset of 33k reasoning traces. The data originates from mathematics competitions and includes problems from MATH~\citep{hendrycks2021measuringmathematicalproblemsolving, lightman2023let}, NuminaMathCoT~\citep{numina_math_datasets}, and AIME (1983--2023)~\citep{aime2025}. \texttt{Tina-STILL-3-1.5B-preview} uses the same dataset and reward pipeline.
    
    \item \textbf{DeepScaleR-1.5B-Preview}~\citep{deepscaler2025} focuses on long-context mathematical reasoning via RL, and is trained over approximately 40k problem-answer pairs drawn from the AIME~\citep{aime2025}, AMC~\citep{amc23}, OMNI-MATH~\citep{gao2024omnimathuniversalolympiadlevel}, and STILL~\citep{Slow_Thinking_with_LLMs_3_Preview} datasets. \texttt{Tina-DeepScaleR-1.5B-Preview} uses this dataset and mirrors the reward design.
    
    \item \textbf{Open-RS1/2/3}~\citep{dang2025reinforcementlearningreasoningsmall} are three models from the \texttt{Open-RS} project exploring reasoning performance in 1.5B models trained via RL. All Open-RS models are trained on small, high-quality datasets further curated from the s1~\citep{muennighoff2025s1simpletesttimescaling} (\emph{i.e.}, \textbf{Open-S1}) and DeepScaleR~\citep{deepscaler2025} (\emph{i.e.}, \textbf{Open-DeepScaleR}) datasets. The Tina models (\texttt{Tina-Open-RS1/2/3}) replicate these setups, using identical data splits and reward scaffolding.
\end{itemize}

\subsection{Training Setup: Infrastructure \& Budget}
\label{sec:tiny_setup}
\textbf{Training Codebase.} Our implementation builds upon \texttt{OpenR1},\footnote{\url{https://github.com/huggingface/open-r1}} a fully open reproduction of DeepSeek-R1~\citep{deepseekai2025deepseekr1incentivizingreasoningcapability}
which combines the \texttt{Accelerate}~\citep{accelerate} and \texttt{Trl}~\citep{vonwerra2022trl} libraries and the DeepSpeed ZeRO optimization~\citep{zero}.  It aims to transparently replicate and extend RL methods used for improving reasoning in language models, particularly focusing on aligning model behavior with reasoning-oriented objectives via verifiable reward signals. Our methodology inherits its scaffolding, training utilities, and reward interfaces.

\textbf{Training Hyperparameters.} We initiated parameter selection by replicating key parameters from \texttt{OpenR1}~\citep{openr1} and \texttt{OpenRS}~\citep{dang2025reinforcementlearningreasoningsmall}. For all experiments presented in this paper, we deliberately adopted the default or recommended hyperparameter configurations provided in their works. These settings were kept largely fixed across different runs (Table~\ref{tab:common_hyperparameter}). For the main Tina results (Section \ref{sec:exp_tine_eval}), only reward function parameters were adjusted per task, and for ablation studies (Section \ref{sec:exp_ablation}), only the specific factor under investigation (e.g., learning rate, LoRA rank/alpha, RL algorithm) was varied (Table~\ref{tab:varied_hyperparameter}). This approach intentionally circumvents costly hyperparameter search procedures for our specific setup, ensuring negligible tuning overhead and focusing on the efficacy of the core LoRA-based RL methodology.

\begin{table}[h]
\centering
\resizebox{0.95\textwidth}{!}{
\begin{tabular}{l|ccccc}
\toprule
\rowcolor{LightPink}
\textbf{\textsc{Experimental Task}} & \textbf{\textsc{Training Cost Est.}} & \textbf{\textsc{Evaluation Cost Est.}} & \textbf{\textsc{Total Cost Est.}} \\
\midrule
\textbf{Baseline: Model Re-Evaluation} & - & \$6 & \$6 \\
\midrule
\midrule
\textbf{Main: Tina-STILL-3-1.5B-preview} & \$59 & \$7 & \$66 \\
\midrule
\textbf{Main: Tina-DeepScaleR-1.5B-Preview} & \$84 & \$10 & \$94 \\
\midrule
\textbf{Main: Tina-Open-RS1} & \$40 & \$11 & \$51 \\
\midrule
\textbf{Main: Tina-Open-RS2} & \$15 & \$17 & \$32 \\
\midrule
\textbf{Main: Tina-Open-RS3} & \$15 & \$17 & \$32 \\
\midrule
\midrule
\textbf{Ablation: OpenThoughts Dataset} & \$84 & \$10 & \$94 \\
\midrule
\textbf{Ablation: OpenR1 Dataset} & \$59 & \$7 & \$66 \\
\midrule
\textbf{Ablation: LIMR Dataset} & \$4 & \$4 & \$8 \\
\midrule
\textbf{Ablation: DrGRPO Algorithm} & \$15 & \$17 & \$32 \\
\midrule
\textbf{Ablation: Learning Rate} & \$7 & \$8 & \$15 \\
\midrule
\textbf{Ablation: LoRA Rank/Alpha} & \$14 & \$16 & \$30 \\
\midrule
\midrule
\textbf{Total: All Tasks} & \textbf{\$396} & \textbf{\$130} & \textbf{\$526} \\
\midrule
\textbf{Total: Main Tasks} & \textbf{\$213} & \textbf{\$62} & \textbf{\$275} \\
\midrule
\textbf{Total: Best Ckpt.\ in Each Main Task} & \textbf{\$80} & \textbf{\$5} & \textbf{\$85} \\
\midrule
\textbf{Total: All Ckpt.\ in Best-Performance Task} & \textbf{\$14} & \textbf{\$17} & \textbf{\$31} \\
\midrule
\textbf{Total: Best Ckpt.\ in Best-Performance Task} & \textbf{\$8} & \textbf{\$1} & \textbf{\$9} \\
\bottomrule
\end{tabular}
}
\caption{\textbf{Computational cost breakdown.} Costs for all experimental tasks in this paper, measured in USD.
The row ``Best Ckpt.\ in Each Main Task'' denotes the cost of reproducing the best checkpoint in each of Table~\ref{tab:tina_still_eval},~\ref{tab:tina_deepscaler_eval},~\ref{tab:tina_openrs3_eval},~\ref{tab:tina_openrs2_eval},~\ref{tab:tina_openrs1_eval}.
The row ``All Ckpt.\ in Best-Performance Task'' denotes the cost of reproducing all checkpoints in Table~\ref{tab:tina_openrs2_eval}. ``Best Ckpt.\ in Best-Performance Task'' denotes the cost of reproducing the best checkpoint in Table~\ref{tab:tina_openrs2_eval}, \emph{i.e.}, the checkpoint at step 450.}
\label{tab:cost_breakdown}
\vspace{-3mm}
\end{table}

\textbf{Training Hardware.} A key element of our low-cost approach was minimizing the hardware footprint. While distributed RL training algorithms like GRPO often benefit from using three or more GPUs (\emph{e.g.,} dedicating one GPU to an inference engine such as \texttt{vLLM} for faster sample generation), we deliberately targeted a minimal setup using only two NVIDIA L40S GPUs.\footnote{Occasionally, NVIDIA RTX 6000 Ada GPUs were used instead, which is reflected in the system configuration metadata on Weights \& Biases. From our practical experience, these two GPU types are similar in terms of cost and computational performance. For consistency, we report costs and compute metrics based on the L40S.} To enable this, we co-located the RL training process and the \texttt{vLLM} on the same two GPUs by constraining \texttt{vLLM}'s GPU memory usage. The training itself utilized data parallelism across both GPUs. While running inference and training concurrently on two GPUs might result in a longer wall-clock training time compared to a setup with dedicated inference GPUs, it significantly reduces the hardware requirement.

\textbf{Training Budget.} The NVIDIA L40S GPUs we use are accessible via commercial cloud platforms at an approximate rate of \$1 USD per GPU hour, including 300 GB storage, based on pricing observed at the time of writing~\citep{pricing}. The RL training process for our LoRA models proved highly efficient, with a single RL step typically completing within one minute on this hardware. Evaluating a model checkpoint across our entire suite of six reasoning benchmarks required approximately 1 L40S GPU hours on average. To ensure cost control, we initially established a conservative maximum budget of \$100 USD for each complete experimental run, encompassing all stages from training to evaluation and miscellaneous tasks. As detailed in Table \ref{tab:cost_breakdown}, our actual expenditures were significantly below this ceiling. Our calculation is based on the full Tina model evaluation performance in Appendix~\ref{app:full_tina_eval}. We believe this low cost makes our setup an accessible testbed for the research community.

\vspace{-2mm}
\section{Surprising Effectiveness of Efficient RL Reasoning via LoRA}

\subsection{Experiments Stage I: Baseline Model Re-Evaluation}
\label{sec:exp_baseline_eval}

Before presenting Tina's performance, it is crucial to establish fair and reliable comparisons against existing SOTA reasoning models. We note that performance scores reported in the literature for relevant models often stem from evaluations using disparate frameworks (\emph{e.g.,} \texttt{verl}~\citep{Sheng_2025}, \texttt{lighteval}~\citep{lighteval}, \texttt{lm-eval-harness}~\citep{eval-harness}) and inconsistent inference settings (such as different generation hyperparameters or varying numbers of GPUs). These variations can significantly influence reported metrics, creating potential inconsistencies and hindering reliable comparisons between models.

\begin{table}[h]
\centering
\resizebox{0.9\textwidth}{!}{
\begin{tabular}{l|ccccccc|cc}
\toprule
\rowcolor{LightPink}
\textbf{\textsc{Baseline Model}} & \textbf{\textsc{AIME24}} & \textbf{\textsc{AIME25}} & \textbf{\textsc{AMC23}} & \textbf{\textsc{MATH500}} & \textbf{\textsc{GPQA}} & \textbf{\textsc{Minerva}} & \textbf{\textsc{Avg.}} \\
\midrule
\textbf{DeepSeek-R1-Distilled-Qwen-1.5B} & 23.33 & 16.67 & 62.50 & 82.60 & 31.82 & 30.15 & 41.18 \\
\midrule
\textbf{STILL-3-1.5B-preview} & 26.67 & \textbf{26.67} & 67.50 & 86.40 & 34.34 & 27.57 & 44.86 \\
\midrule
\textbf{DeepScaleR-1.5B-Preview} & 36.67 & \textbf{26.67} & \textbf{77.50} & \textbf{87.80} & 31.82 & \textbf{31.99} & \textbf{48.74} \\
\midrule
\textbf{Open-RS1} & 26.67 & 20.00 & 72.50 & 83.60 & \textbf{35.35} & 28.68 & 44.47 \\
\midrule
\textbf{Open-RS2} & 26.67 & 13.33 & 62.50 & 85.40 & 34.85 & 26.84 & 41.60 \\
\midrule
\textbf{Open-RS3} & \textbf{43.33} & 20.00 & 67.50 & 83.00 & 33.84 & 28.68 & 46.06 \\
\bottomrule
\end{tabular}
}
\caption{\textbf{Baseline model re-evaluation.} Performance evaluation of baseline models on six reasoning tasks.}
\label{tab:baseline_eval}
\vspace{-3mm}
\end{table}

To mitigate these confounding factors, we performed a comprehensive re-evaluation of key baseline models using a single, consistent methodology throughout this paper. All baseline evaluations reported herein utilize the \texttt{lighteval} framework integrated with the \texttt{vLLM}~\citep{kwon2023efficient} inference engine for efficient generation. For comparability with prior work such as \texttt{OpenR1}, we maintained a fixed hardware configuration (two L40S GPUs) and applied a standardized set of \texttt{vLLM} inference parameters across all evaluated baseline models. All scores are zero-shot pass@1 performance. The exact command structure employed for these evaluations is provided in Appendix~\ref{app:eval_command} for transparency and reproducibility. The results stemming from this consistent re-evaluation protocol are presented in Table~\ref{tab:baseline_eval}.

Particularly, we evaluate the reasoning capabilities of our Tina models and the baselines across a diverse suite of six challenging benchmarks, primarily focused on mathematical and scientific reasoning:
\begin{itemize}[leftmargin=7mm,itemsep=2mm, topsep=0em]
    \item \textbf{AIME24/25}~\citep{aime2025} contains 30 high-school-level math problems in algebra, geometry, number theory, and combinatorics from the 2024/2025 American Invitational Mathematics Examination. Each problem demands precise multi-step reasoning. 
    \item \textbf{AMC23}~\citep{amc23} includes 40 problems from the 2023 American Mathematics Competition, offering a mix of logic and symbolic manipulation tasks.
    \item \textbf{MATH500}~\citep{hendrycks2021measuringmathematicalproblemsolving, lightman2023let} is a benchmark comprising 500 competition mathematics problems derived from various sources, covering different difficulty levels and often necessitating multi-step derivation and calculation. 
    \item \textbf{GPQA Diamond}~\citep{rein2024gpqa}, hereafter referred to as GPQA, consists of 198 PhD-level science questions across biology, chemistry, and physics. Each question is multiple-choice with subtle distractors.
    \item \textbf{Minerva}~\citep{NEURIPS2022_18abbeef} includes 272 quantitative reasoning problems generally at the undergraduate level. The questions span multiple STEM fields, including physics, biology, chemistry, and economics, often requiring mathematical modeling or calculation steps. Includes tasks such as calculating enzyme kinetics from reaction data.
\end{itemize}

\vspace{-2mm}
\subsection{Experiments Stage II: Tina Model Evaluation}
\label{sec:exp_tine_eval}

We now present the core evaluation results for our Tina models. These experiments assess the reasoning capabilities attained by post-training the \texttt{DeepSeek-R1-Distill-Qwen-1.5B} with minimal parameter updates via LoRA-based RL. The results presented in Table \ref{tab:tina_eval} demonstrate that significant reasoning performance can be achieved efficiently, yielding models that are competitive with, or outperform, relevant baselines despite the inherent resource constraints of using parameter-efficient tuning.\footnote{Tables~\ref{tab:tina_eval} and~\ref{tab:tina_ablation_eval} adopt a consistent naming pattern where ``Tina-\texttt{X}'' denotes our model is the LoRA counterpart of a baseline model \texttt{X} or is trained on a dataset \texttt{X} (possibly followed with an extra ablation setup). This can reflect the model origin and serve as a direct reference to the public checkpoints for reproducibility.}

\begin{table}[h]
\centering
\resizebox{\textwidth}{!}{
\begin{tabular}{lc|cccccc|cc}
\toprule
\rowcolor{LightPink}
\textbf{\textsc{Tina Model}} & \textbf{\textsc{Steps (\% of 1 Epoch)}} & \textbf{\textsc{AIME24}} & \textbf{\textsc{AIME25}} & \textbf{\textsc{AMC23}} & \textbf{\textsc{MATH500}} & \textbf{\textsc{GPQA}} & \textbf{\textsc{Minerva}} & \textbf{\textsc{Avg.}} & \textbf{\textsc{Baseline}} \\
\midrule
\textbf{Tina-STILL-3-1.5B-preview} & 53\% & 36.67 & \textbf{30.00} & 77.50 & 84.60 & 33.33 & 26.84 & 48.16 & 44.86 \\
\midrule
\textbf{Tina-DeepScaleR-1.5B-Preview} & 19\% & \textbf{43.33} & 26.67 & 67.50 & 86.20 & \textbf{37.88} & 28.68 & 48.38 & \textbf{48.74} \\
\midrule
\textbf{Tina-Open-RS1} & 34\% & \textbf{43.33} & 20.00 & 80.00 & 84.00 & 35.35 & 28.68 & 48.56 & 44.47 \\
\midrule
\textbf{Tina-Open-RS2} & 51\% & \textbf{43.33} & 26.67 & 77.50 & \textbf{87.00} & 36.36 & \textbf{32.72} & \textbf{50.60} & 41.60 \\
\midrule
\textbf{Tina-Open-RS3} & 57\% & 36.67 & 23.33 & \textbf{82.50} & 85.20 & 37.37 & 31.62 & 49.45 & 46.06 \\
\bottomrule
\end{tabular}
}
\caption{\textbf{Tina model evaluation.} Performance comparison between Tina models and corresponding full-parameter-trained SOTA models on six reasoning tasks. The value in the \textit{Steps} column indicates the training steps of the best model checkpoint within one epoch, the full model checkpoint evaluation is shown in Appendix~\ref{app:full_tina_eval}. The \textit{Baseline} column represents the average score achieved by baseline model with full-parameter RL in Table~\ref{tab:baseline_eval}.}
\label{tab:tina_eval}
\end{table}

Table \ref{tab:tina_eval} summarizes the performance of five distinct Tina models across a suite of six reasoning tasks: AIME24/25, AMC23, MATH500, GPQA, and Minerva. For each Tina model, we report the extent of training completed (as a percentage of a predefined training stpes within 1 epoch) and the percentage scores achieved on each task. 
The results compellingly demonstrate the efficacy of our economical LoRA-based RL strategy. All Tina models exhibit substantial reasoning aptitude, achieving average scores in the range of 48.16\% to 50.60\%. Significantly, nearly all Tina models notably outperform their corresponding baseline average scores , indicating marked improvements instilled by the parameter-efficient RL. The \texttt{Tina-Open-RS2} model yielded the highest average performance observed at 50.60\%. Furthermore, these strong results were achieved with remarkably limited training durations, ranging from just 19\% to 57\% of a full training epoch, highlighting the efficiency and rapid adaptation enabled by the Tina approach. These findings strongly support our central hypothesis: robust reasoning capabilities can be effectively and economically cultivated in small language models through the targeted application of LoRA and RL.

\vspace{-2mm}
\subsection{Experiments Stage III: Tina Ablation Variants}
\label{sec:exp_ablation}

\begin{table}[h]
\centering
\resizebox{\textwidth}{!}{
\begin{tabular}{lc|cccccc|c}
\toprule
\rowcolor{LightPink}
\textbf{\textsc{Ablation on Datasets}} & \textbf{\textsc{Steps (\% of 1 Epoch)}} & \textbf{\textsc{AIME24}} & \textbf{\textsc{AIME25}} & \textbf{\textsc{AMC23}} & \textbf{\textsc{MATH500}} & \textbf{\textsc{GPQA}} & \textbf{\textsc{Minerva}} & \textbf{\textsc{Avg.}} \\
\midrule
\textbf{Tina-OpenR1 (93.7k)} & 13\% & 36.67 & 26.67 & 75.00 & 86.80 & 39.90 & 30.51 & 49.26 \\
\midrule
\textbf{Tina-OpenThoughts (66.1k)} & 30\% & 36.67 & 26.67 & 72.50 & 84.80 & \textbf{41.41} & \textbf{33.09} & 49.19 \\
\midrule
\textbf{Tina-DeepScaleR (40.3k)} & 19\% & 43.33 & 26.67 & 67.50 & 86.20 & 37.88 & 28.68 & 48.38 \\
\midrule
\textbf{Tina-STILL-3 (33k)} & 53\% & 36.67 & \textbf{30.00} & 77.50 & 84.60 & 33.33 & 26.84 & 48.16 \\
\midrule
\textbf{Tina-Open-S1 (18.6k)} & 34\% & 43.33 & 20.00 & \textbf{80.00} & 84.00 & 35.35 & 28.68 & 48.56 \\
\midrule
\textbf{Tina-Open-RS (7k)} & 51\% & 43.33 & 26.67 & 77.50 & \textbf{87.00} & 36.36 & 32.72 & \textbf{50.60} \\
\midrule
\textbf{Tina-LIMR (1.39k)} & 58\% & \textbf{46.67} & 20.00 & 75.00 & 83.80 & 34.85 & 30.51 & 48.47 \\
\midrule
\midrule
\rowcolor{LightPink}
\textbf{\textsc{Ablation on Learning Rate}} & \textbf{\textsc{Steps (\% of 1 Epoch)}} & \textbf{\textsc{AIME24}} & \textbf{\textsc{AIME25}} & \textbf{\textsc{AMC23}} & \textbf{\textsc{MATH500}} & \textbf{\textsc{GPQA}} & \textbf{\textsc{Minerva}} & \textbf{\textsc{Avg.}} \\
\midrule
\textbf{Tina-LIMR-5e-6-lr} & 29\% & 36.67 & \textbf{26.67} & 75.00 & 83.60 & \textbf{35.86} & 29.41 & 47.87 \\
\midrule
\textbf{Tina-LIMR-1e-6-lr} & 58\% & \textbf{46.67} & 20.00 & 75.00 & 83.80 & 34.85 & \textbf{30.51} & \textbf{48.47} \\
\midrule
\textbf{Tina-LIMR-5e-7-lr} & 58\% & 43.33 & 16.67 & \textbf{77.50} & \textbf{84.60} & 34.85 & \textbf{30.51} & 47.91 \\
\midrule
\midrule
\rowcolor{LightPink}
\textbf{\textsc{Ablation on LoRA Rank}} & \textbf{\textsc{Steps (\% of 1 Epoch)}} & \textbf{\textsc{AIME24}} & \textbf{\textsc{AIME25}} & \textbf{\textsc{AMC23}} & \textbf{\textsc{MATH500}} & \textbf{\textsc{GPQA}} & \textbf{\textsc{Minerva}} & \textbf{\textsc{Avg.}} \\
\midrule
\textbf{Tina-LIMR-64-LoRA-rank} & 29\% & 20.00 & 30.00 & 77.50 & \textbf{84.20} & \textbf{38.38} & \textbf{31.62} & 46.95 \\
\midrule
\textbf{Tina-LIMR-32-LoRA-rank} & 58\% & \textbf{46.67} & 20.00 & 75.00 & 83.80 & 34.85 & 30.51 & 48.47 \\
\midrule
\textbf{Tina-LIMR-16-LoRA-rank} & 58\% & 43.33 & \textbf{33.33} & 70.00 & 83.20 & 35.35 & 28.31 & \textbf{48.92} \\
\midrule
\textbf{Tina-LIMR-8-LoRA-rank} & 29\% & 30.00 & 26.67 & 82.50 & 83.80 & 33.84 & 30.51 & 47.89 \\
\midrule
\textbf{Tina-LIMR-4-LoRA-rank} & 86\% & 36.67 & 20.00 & \textbf{85.00} & 83.80 & 31.82 & 29.04 & 47.72 \\
\midrule
\midrule
\rowcolor{LightPink}
\textbf{\textsc{Ablation on RL Algorithm}} & \textbf{\textsc{Steps (\% of 1 Epoch)}} & \textbf{\textsc{AIME24}} & \textbf{\textsc{AIME25}} & \textbf{\textsc{AMC23}} & \textbf{\textsc{MATH500}} & \textbf{\textsc{GPQA}} & \textbf{\textsc{Minerva}} & \textbf{\textsc{Avg.}} \\
\midrule
\textbf{Tina-Open-RS3-GRPO} & 57\% & 36.67 & \textbf{23.33} & \textbf{82.50} & \textbf{85.20} & \textbf{37.37} & \textbf{31.62} & 49.45 \\
\midrule
\textbf{Tina-Open-RS3-DrGRPO} & \textbf{17\%} & \textbf{43.33} & \textbf{23.33} & 80.00 & 85.00 & 35.35 & 30.15 & \textbf{49.53} \\
\bottomrule
\end{tabular}
}
\caption{\textbf{Tina ablation variants evaluation.} Performance evaluation of Tina's ablation variants on six reasoning tasks. The value in the \textit{Steps} column indicates the training steps of the best model checkpoint within one epoch, the full model checkpoint evaluation is shown in Appendix~\ref{app:full_tina_eval}. 
For the number in parentheses (the ablation on datasets), it means the data size of a dataset. During training, this number should be multiplied by the number of generation in GRPO-like algorithm (in our case, that multiplier is 4).
For the model names, \texttt{Tina-LIMR}, \texttt{Tina-LIMR-1e-6-lr} and \texttt{Tina-LIMR-32-LoRA-rank} are the same model, we duplicate them for better visualization. The same idea applies to \texttt{Tina-DeepScaleR} and \texttt{Tina-DeepScaleR-1.5B-Preview}, \texttt{Tina-STILL-3} and \texttt{Tina-STILL-3-1.5B-preview}, \texttt{Tina-Open-S1} and \texttt{Tina-Open-RS1}, \texttt{Tina-Open-RS} and \texttt{Tina-Open-RS2}, \texttt{Tina-Open-RS3-GRPO} and \texttt{Tina-Open-RS3}.}
\label{tab:tina_ablation_eval}
\vspace{-3mm}
\end{table}

To better understand the factors influencing the performance and efficiency of our Tina models within the proposed low-cost framework, we conducted a series of ablation studies. These studies systematically investigate the impact of key design choices and hyperparameter: the underlying training dataset, the learning rate for LoRA updates, the rank of the LoRA adapters, and the specific RL algorithm employed. In each study, we typically varied one factor while holding others constant, often based on a high-performing configuration identified in our main experiments or preliminary runs. The results, summarized in Table~\ref{tab:tina_ablation_eval}, provide valuable insights into the robustness and sensitivity of our economical approach.

\textbf{Impact of Training Dataset.}
The first section of Table \ref{tab:tina_ablation_eval} highlights the influence of the dataset used for RL. We compared seven distinct datasets, varying significantly in size (from $\approx$1.4k to $\approx$94k samples). Strikingly, the \texttt{Tina-Open-RS} model, trained on a concise dataset of merely 7k examples, achieved the highest average score (50.60\%). This outcome surpasses models trained on considerably larger datasets, such as \texttt{Tina-OpenR1} (93.7k samples, 49.26\% avg). This observation strongly supports our core ``Tiny'' premise and reflects the intuition that the quality and diversity of the dataset matter more than the data size.

\textbf{Sensitivity to Learning Rate.}
Using the \texttt{Tina-LIMR} configuration as a testbed (second section of Table \ref{tab:tina_ablation_eval}), we assessed sensitivity to the learning rate. Among the tested values ($5 \times 10^{-6}$, $1 \times 10^{-6}$, and $5 \times 10^{-7}$), a learning rate of $1 \times 10^{-6}$ yielded the optimal average performance (48.47\%) for this setup. While performance differences were not drastic, this indicates that learning rate selection remains a factor, although effective results were obtained without extensive tuning.

\textbf{Effect of LoRA Rank.}
The third ablation study investigated the impact of LoRA rank, which directly controls the number of trainable parameters. Testing ranks 4, 8, 16, 32, and 64 on the \texttt{Tina-LIMR} setup, we observed considerable robustness. Ranks 8, 16, and 32 all produced strong results, with average scores clustering between 47.89\% and 48.92\%. Notably, rank 16 achieved the peak performance (48.92\%) in this comparison, slightly outperforming rank 32 (48.47\%). Performance decreased slightly at the extremes (rank 4 and 64). This study validates that highly parameter-efficient configurations (low ranks like 16 or 32) are effective, further enhancing the cost-effectiveness and minimal overhead of the Tina approach.

\textbf{Comparison of RL Algorithms.}
Finally, we compared two RL algorithms, GRPO and Dr.GRPO~\citep{liu2025understandingr1zeroliketrainingcritical}, using the \texttt{Tina-Open-RS3} setup (final section of Table \ref{tab:tina_ablation_eval}). Both algorithms led to similar peak average performance levels (49.45\% for GRPO vs.\ 49.53\% for Dr.GRPO). However, Dr.GRPO reached its best checkpoint significantly earlier in the training process (17\% of an epoch vs.\ 57\% for GRPO). This suggests potential advantages in sample efficiency for Dr.GRPO in this context with an alternative normalization in loss calculation, offering potentially faster convergence and further reductions in training time and cost.

\vspace{-2mm}
\section{Hypothesis for Effective and Efficient LoRA: Rapid Format Adaptation}

\begin{figure}[h]
    \centering
    \includegraphics[width=\linewidth]{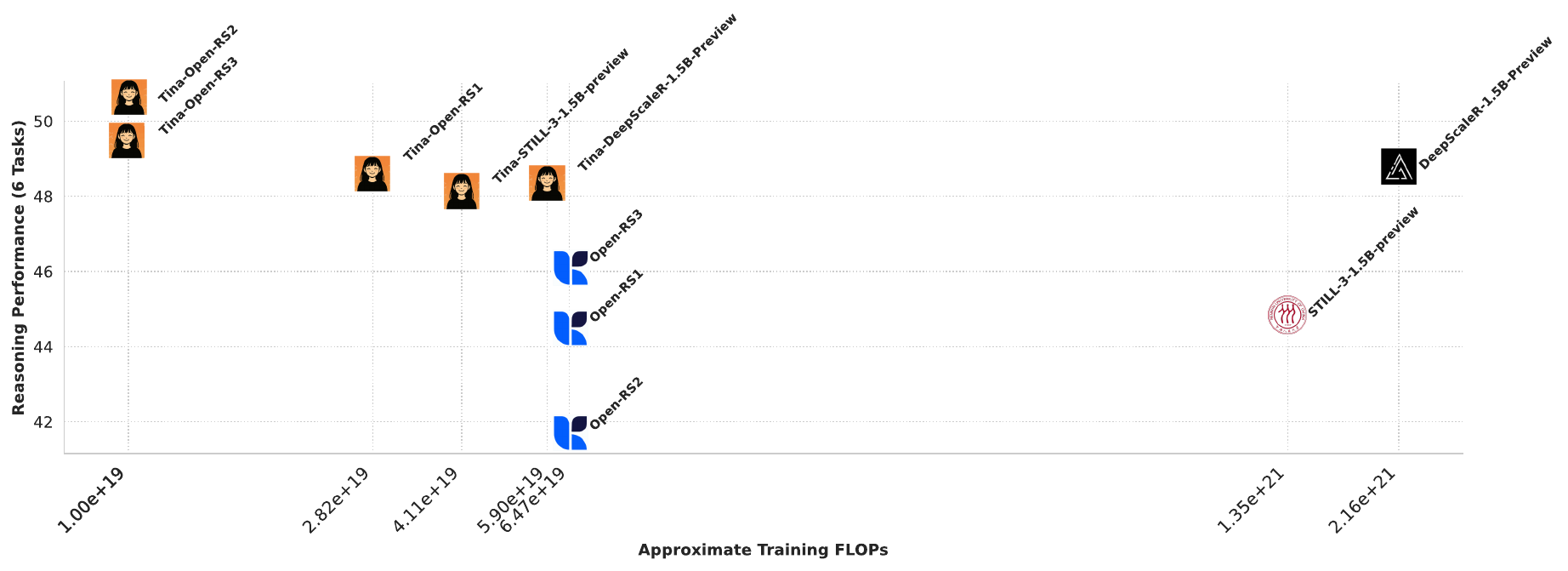}
    \caption{\textbf{Less is more LoRA-based RL}. Approximate training FLOPs vs reasoning performance comparison between Tina and baseline models. The calculation is detailed in Appendix~\ref{sec:cost_breakdown}.}
    \label{fig:flop}
\end{figure}

\textbf{Less is More LoRA-based RL.} To understand why LoRA facilitates both effective and efficient reasoning improvements via RL, we analyze the relationship between training compute and performance, alongside training dynamics. As illustrated in Figure~\ref{fig:flop}, plotting reasoning performance against approximate training FLOPs reveals a stark contrast between full-parameter and LoRA-based training regimes. First, our LoRA-based Tina models achieve reasoning scores comparable or superior to fully fine-tuned baselines while requiring (in some cases) orders of magnitude fewer training FLOPs. We observe that in LoRA models, increased training compute inversely affects performance, in contrast to full-parameter models. This observation highlights a ``less compute can yield more performance'' phenomenon.

This finding supports our hypothesis regarding how LoRA achieves such remarkable efficiency, which relates to the principle of ``learn structure/format, maintain knowledge.'' We posit that LoRA excels in this scenario because RL for reasoning heavily rewards the model's ability to generate outputs in a specific, verifiable format or structure (\emph{e.g.,} step-by-step reasoning chains). LoRA appears to be highly adept at learning these structural and stylistic patterns with minimal parameter changes, thus requiring very few FLOPs. At the same time, because LoRA modifies only a tiny fraction of the weights, it largely preserves the base model's vast pre-trained knowledge. Therefore, LoRA efficiently teaches the model how to format its existing knowledge into effective reasoning traces, rather than potentially imposing costly relearning of concepts or procedures that extensive full-parameter updates might entail. We hypothesize that this focus on structural adaptation allows Tina to achieve high reasoning performance with minimal computational investment.

\begin{figure}[h!]
    \centering
    \includegraphics[width=.45\linewidth]{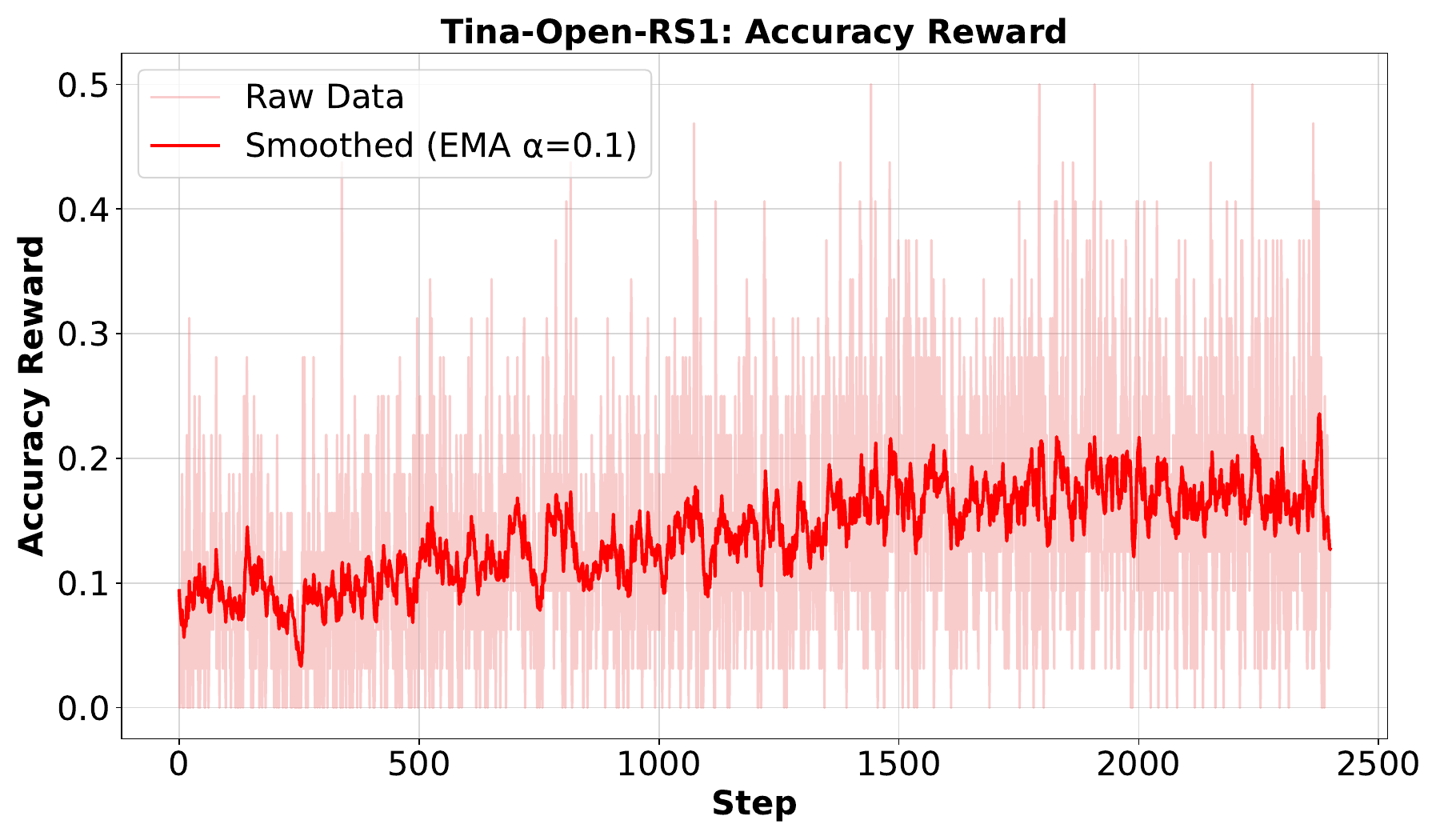}
    \includegraphics[width=.45\linewidth]{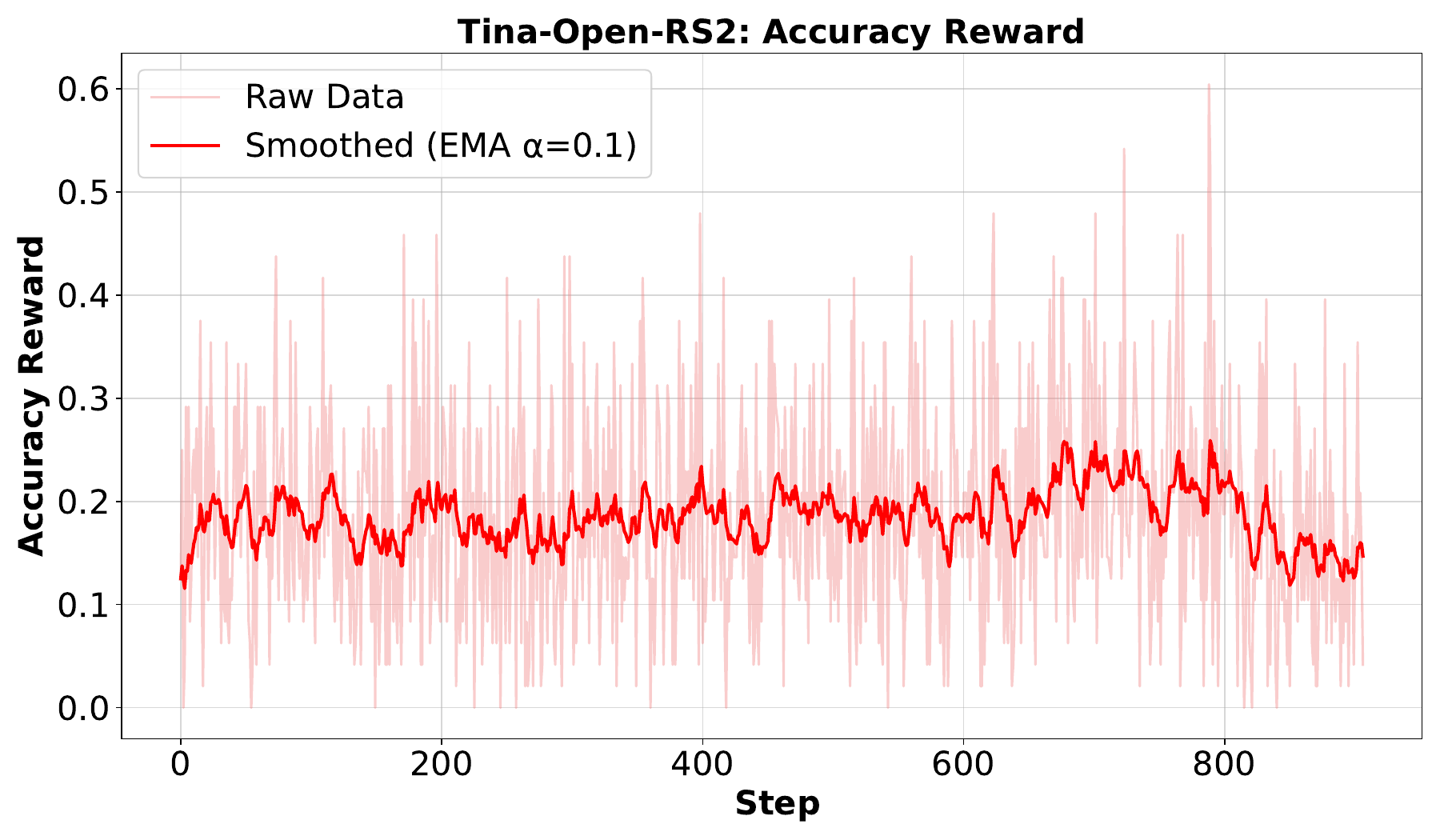}
    \includegraphics[width=.45\linewidth]{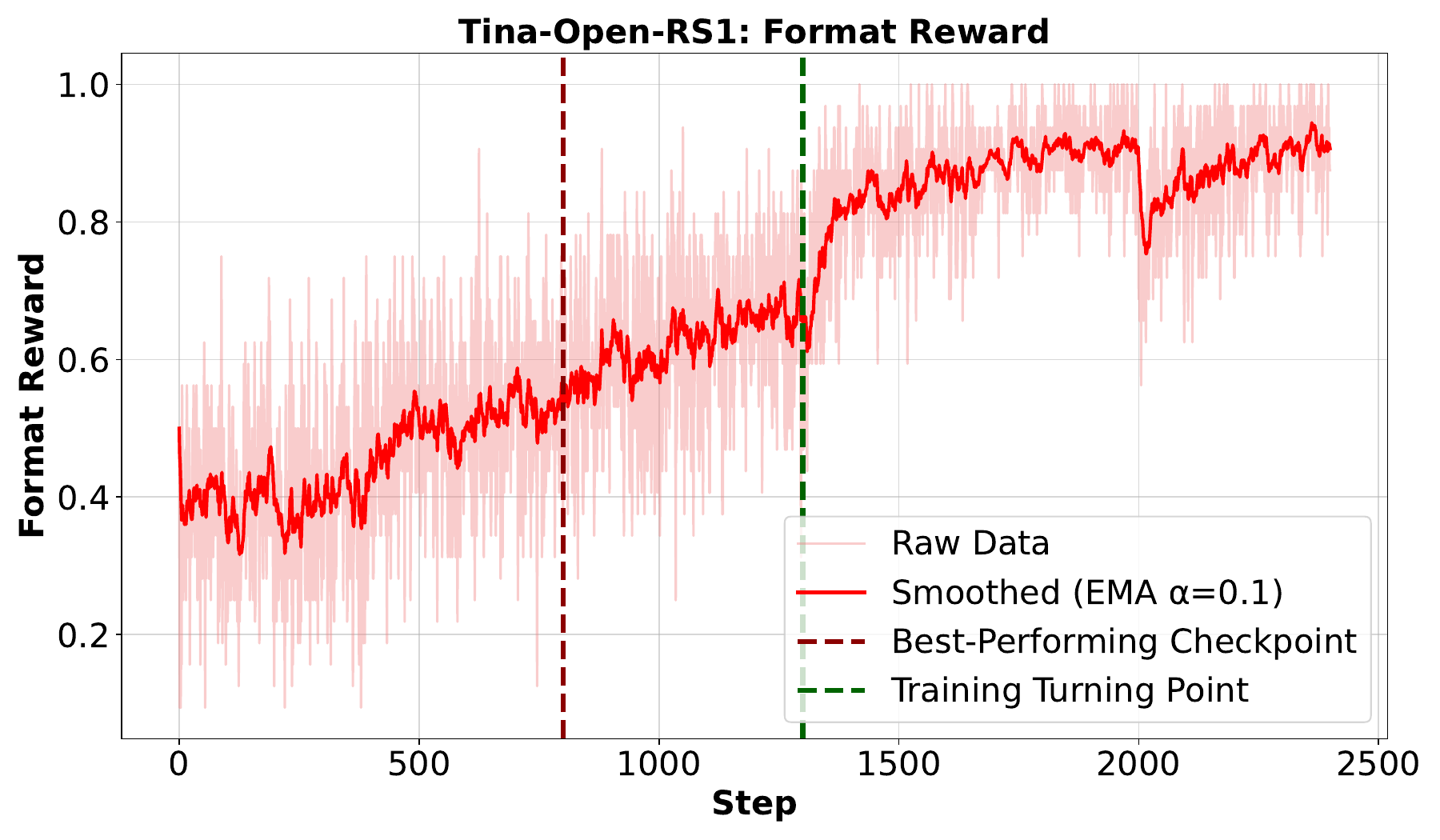}
    \includegraphics[width=.45\linewidth]{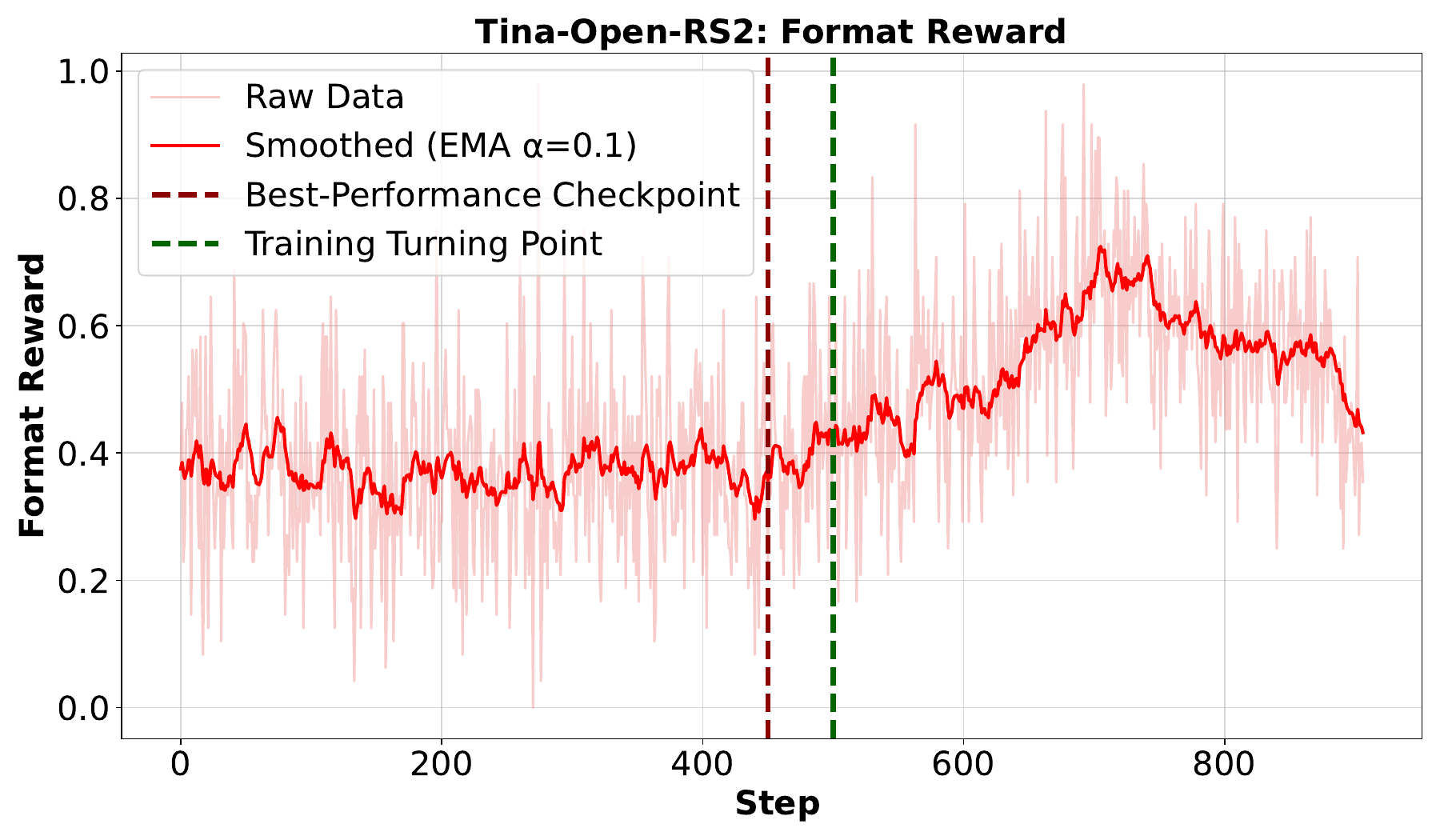}
    \includegraphics[width=.45\linewidth]{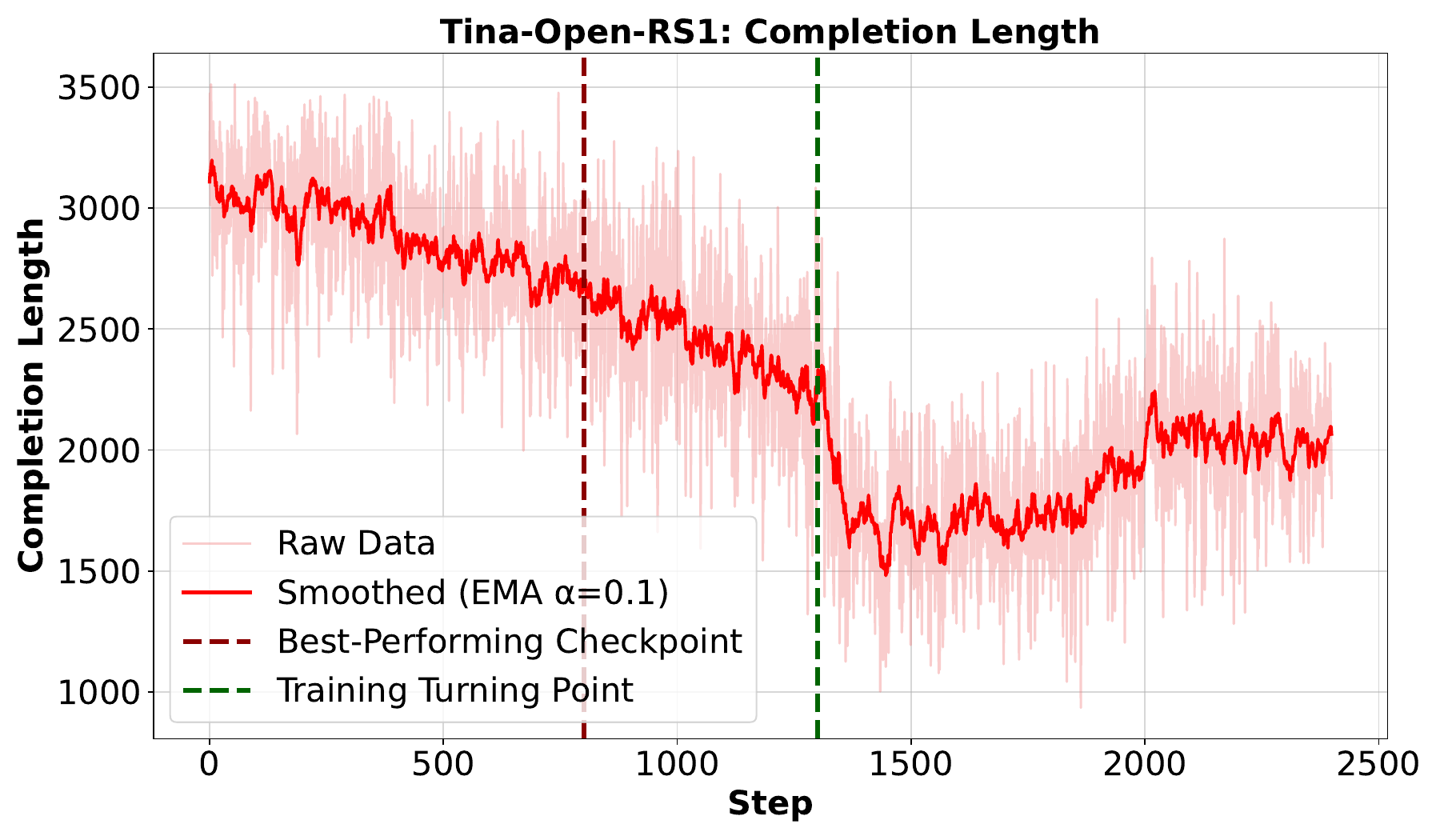}
    \includegraphics[width=.45\linewidth]{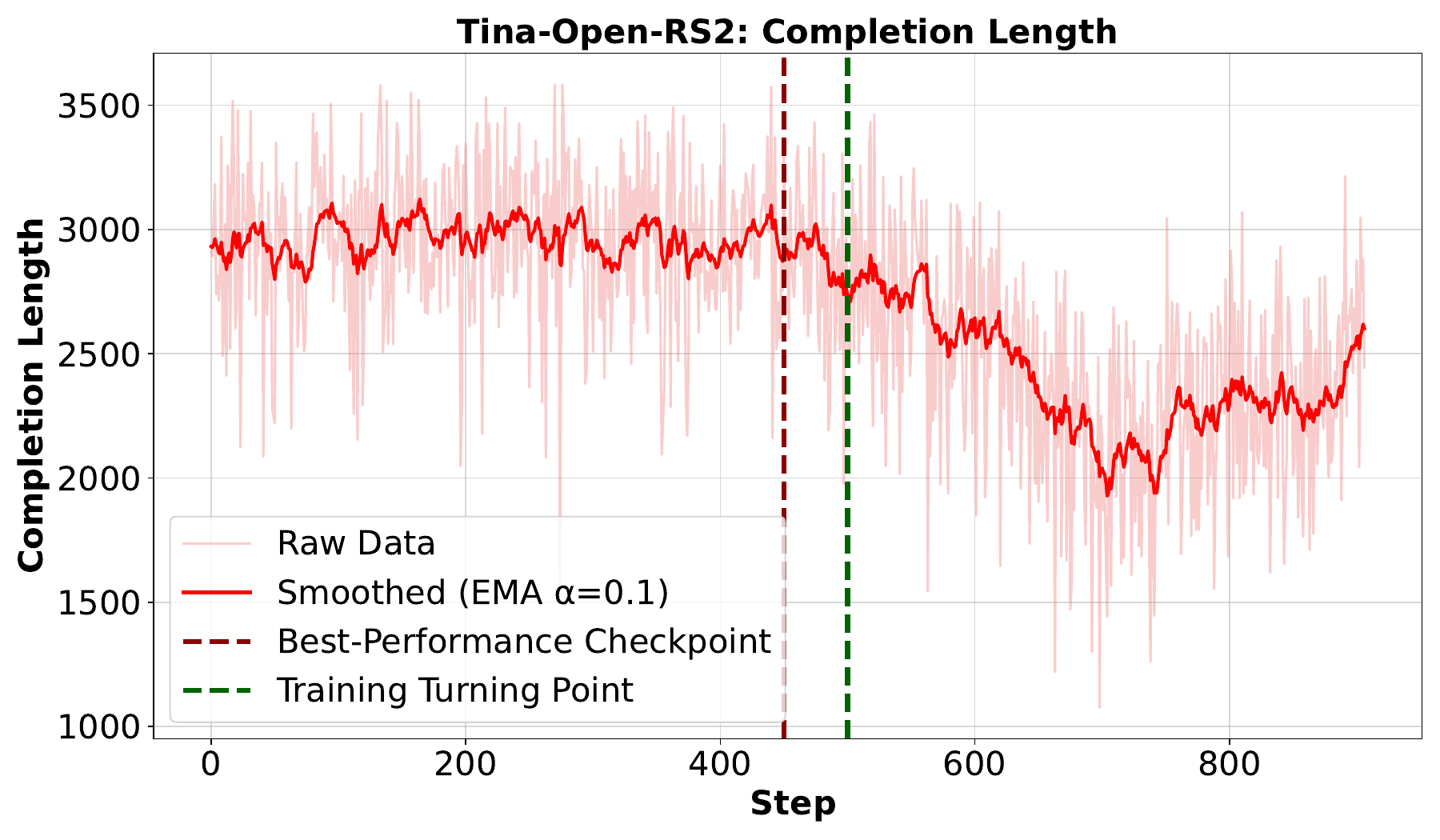}
    \caption{\textbf{Phase transition in LoRA-based RL.} The raw data is from the Weights \& Biases training logs and smoothed via exponential moving average (EMA) with factor $0.1$. The ``training turning point'' in the legend means the step where the format-like metrics (e.g., format reward, completion length) start to destabilize. Refer to Appendix~\ref{app:full_tina_phase_transit} for the full set of plots.}
    \label{fig:phase_transit}
\end{figure}

\textbf{Phase Transition in LoRA-based RL.}
Further insights into the LoRA-based RL mechanism arise from analyzing the training logs. That is, a distinct pattern emerges in  Figure~\ref{fig:phase_transit}, which displays accuracy rewards, format rewards, and completion lengths over training steps for various Tina model runs. We consistently observe a training phase transition or turning point evident in the format-related metrics (format reward, \textit{row 2}; completion length, \textit{row 3}) across most Tina models. Around this transition point (indicated by the green vertical dashed line), the format reward often peaks or destabilizes, while the completion length frequently reaches a minimum before potentially reversing its trend.
Notably, this relatively sharp transition observed in format and length metrics does not typically have a corresponding distinct turning point in the accuracy reward plots (\textit{row 1}). The accuracy reward often exhibits more gradual fluctuations or slower drift over the training duration, without a clear inflection aligned with the format transition.

Another crucial observation is the timing of optimal performance: the best-performing checkpoint, yielding the highest reasoning accuracy on held-out evaluations, consistently occurs just prior to or around this observed phase transition point in the format metrics (indicated by the red vertical dashed line). This decoupling between the dynamics of accuracy-based and format-based metrics suggests that the LoRA-based RL process rapidly optimizes the model's ability to adhere to the structural and stylistic elements rewarded by the format score and length constraints. The subsequent transition point may signify where this structural optimization saturates, becomes unstable, or perhaps begins to compromise generative quality in other ways (\emph{e.g.,} by overly constraining or expanding length). The fact that peak reasoning accuracy is achieved just before this format-driven transition implies that while learning the correct output format is essential and efficiently achieved via LoRA, pushing further on format-centric optimization alone does not necessarily yield better reasoning, and may even be detrimental. This reinforces our hypothesis that LoRA efficiently adapts the model by primarily learning the form required for effective reasoning.
\section{Conclusion}
\label{sec:conclusion}

We presented Tina to demonstrate that effective reasoning capabilities can be instilled in language models with efficiency and effectiveness. The principal contribution of Tina lies in democratizing access to RL-driven reasoning model development. By combining LoRA with RL on a 1.5B parameter base model, we achieved reasoning performance competitive with significantly larger models, accomplishing this within an estimated computational budget of only \$9. This outcome prompts reflection on the factors enabling such minimalist approaches, and on their possible future trajectories. Despite encouraging results, this work is subject to certain limitations:

\textbf{Base Model Scale}: Our experiments centered on a 1.5B parameter model. While showcasing cost-performance efficiency, the absolute reasoning ceiling achievable with this ``tiny'' model may naturally be lower for complex, multi-step reasoning problems than what larger models can offer.

\textbf{Reasoning Task Scope}: Our evaluation focused primarily on mathematical and formal logic reasoning benchmarks (AIME, AMC, MATH, GPQA, Minerva). The effectiveness and transferability of the learned reasoning skills to other domains, such as coding, warrants further investigation.

\textbf{Hyperparameter Optimization}: We intentionally minimized hyperparameter tuning costs by adopting established configurations. While this demonstrates a certain form of robustness to our methodology, there may be potential for further performance gains derived from additional tuning, perhaps tailored to the interplay between LoRA, the RL algorithm, and the target reasoning tasks.

\section{Acknowledgment}

We want to express our gratitude to the broader open-source community. This research was made possible by leveraging numerous publicly available resources, including training and evaluation framework, open datasets, accessible pre-trained language models, and the insights shared through technical reports. The computational resources required for the experiments described herein were provided by the Center for Advanced Research Computing (CARC) at the University of Southern California (USC). We are grateful for the support which enabled the training and evaluation of our models. J.A.\ was supported by the National Science Foundation Graduate Research Fellowship Program under Grant No.\ DGE-1842487. Any opinions, findings, and conclusions or recommendations expressed in this material are those of the authors and do not necessarily reflect the views of the National Science Foundation.

\clearpage
\bibliography{main}

\appendix
\newpage

\appendix

\section*{\hspace{-4mm} \centering Appendix}
\vspace{3mm}

\section{Cost Breakdown}
\label{sec:cost_breakdown}

This section provides further details on how training data amounts, computational cost, time cost, and performance metrics reported in this paper – particularly those presented in figures like Figures~\ref{fig:overall_comparison} and~\ref{fig:flop} – were determined and should be interpreted.

\textbf{Overall Comparison (Figure~\ref{fig:overall_comparison}).}
For the baseline models included in Figure~\ref{fig:overall_comparison}, the approximate training data amounts, computational costs (typically reported as GPU hours or total FLOPs), and training times are sourced from their respective technical reports or publications, leveraging the helpful summary provided in the Open-RS paper \citep{dang2025reinforcementlearningreasoningsmall}. Reasoning performance scores for all models, encompassing both baselines and our Tina models, stem from results presented in Tables~\ref{tab:baseline_eval} and~\ref{tab:tina_eval}.

Also, it is crucial to understand the scope of reported costs:
\begin{itemize}[leftmargin=7mm,itemsep=2mm, topsep=0em]
    \item Epoch vs. Best Checkpoint: Costs cited for Tina and baseline models reflect the resources needed to complete a full training epoch or a predefined training run, not necessarily the minimal cost to reach the single best-performing checkpoint within that run.
    \item Training vs. Evaluation: Reported costs cover training only, omitting the computational expense required for model evaluation across benchmarks since such information is missing from several baseline models.
\end{itemize}
Particularly, the \$9 USD in the abstract represents the estimated cost to train the Tina model up to its best-performing checkpoint and subsequently evaluate that specific checkpoint.
For context comparing potential full training runs, the cost to train a Tina model for a complete epoch is \$14 USD (training only). Including evaluation costs for such a full run would increase the total to approximately \$31 USD. We emphasize the \$9 as representing the efficient path to the best Tina model.

\textbf{FLOPs Estimation (Figure~\ref{fig:flop}).}
The approximate training FLOPs shown in Figure~\ref{fig:flop} serve as a hardware-agnostic measure of computational work. For both Tina and baseline models, these values were estimated based on reported training durations and hardware configurations sourced from technical reports or the Open-RS summary, using standard FLOPs calculation methodologies.

\clearpage
\section{Background behind Tina Training}
\label{app:tina_background}

\subsection{GRPO Formulation}
Recall the following formulation of GRPO: For each question $q$, GRPO samples a group $G = \{o_1, o_2, \dots, o_G\}$ of outputs from the old policy $\pi_{\theta_{\text{old}}}$ and optimizes the policy $\pi_\theta$ by maximizing the following objective:
\begin{align}
\E_{\substack{q \sim P(Q), \\ \{o_i\}_{i=1}^G \sim \pi_{\theta_{\text{old}}}(O|q)}} \nonumber \left[ \frac{1}{G} \sum_{i=1}^G \left( \min\left( \frac{\pi_\theta(o_i|q)}{\pi_{\theta_{\text{old}}}(o_i|q)} A_i, \clip\left(\frac{\pi_\theta(o_i|q)}{\pi_{\theta_{\text{old}}}(o_i|q)}, 1-\epsilon, 1+\epsilon\right) A_i \right) - \beta \, \mathbb{D}_{\text{KL}}(\pi_\theta \| \pi_{\text{ref}}) \right) \right].
\end{align}
Here $A_i$ denotes the advantage computed from a group of rewards $\{r_1, r_2, \dots, r_G\}$
\[
    A_i = \frac{r_i - {\text{mean}}(\{r_1, r_2, \dots, r_G\})}{{\text{std}}(\{r_1, r_2, \dots, r_G\})},
\]
and
\[ \mathbb{D}_{\text{KL}}(\pi_\theta \| \pi_{\text{ref}}) = \frac{\pi_{\text{ref}}(o_i|q)}{\pi_\theta(o_i|q)} - \log \frac{\pi_{\text{ref}}(o_i|q)}{\pi_\theta(o_i|q)} - 1. \]
Note that $\epsilon$ and $\beta$ are parameters controlling the clipping range and KL penalty, respectively.

\subsection{LoRA Formulation}
We follow the standard LoRA setup~\citep{hu2021loralowrankadaptationlarge}. Given a frozen pretrained weight matrix $W_0 \in \mathbb{R}^{d \times k}$ and trainable low-rank matrices $A \in \mathbb{R}^{d \times r}$ and $B \in \mathbb{R}^{r \times k}$ with $r \ll \min(d, k)$, the original forward pass $h(x) = W_0 x$ is modified as
\begin{align*}
    \hat{h}(x) = W_0 x + ABx. 
\end{align*}
We use the default LoRA implementation provided in the \texttt{PEFT}~\citep{peft} library.

\clearpage
\section{Additional Experimental Details}

\subsection{Hyperparameters}
\label{app:hyperparameter}

We show our default choice of hyperparameter in Table~\ref{tab:common_hyperparameter} for all the LoRA-based RL experiments. 

\begin{table}[h]
    \small  
    \centering
    \begin{tabular}{lccccc}
        \toprule
        Tina-STILL-3-1.5B-preview & \multicolumn{5}{c}{LoRA} \\
        Tina-DeepScaleR-1.5B-Preview & \multicolumn{5}{c}{LoRA} \\
        Tina-Open-RS\{X\}-\{Y\} & \multicolumn{5}{c}{LoRA} \\
        Tina-LIMR-\{Z\} & \multicolumn{5}{c}{LoRA} \\
        Tina-OpenR1 & \multicolumn{5}{c}{LoRA} \\
        Tina-OpenThoughts & \multicolumn{5}{c}{LoRA} \\
        \midrule
        LoRA Modules & \multicolumn{5}{c}{query, key, value, dense} \\
        LoRA Rank & \multicolumn{5}{c}{32} \\
        LoRA $\alpha$ & \multicolumn{5}{c}{128} \\
        LoRA Dropout & \multicolumn{5}{c}{0.05} \\
        \midrule
        Algorithm & \multicolumn{5}{c}{GRPO} \\ 
        Optimizer & \multicolumn{5}{c}{AdamW} \\
        Optimizer Momentum & \multicolumn{5}{c}{$\beta_1$, $\beta_2$ = 0.9, 0.999} \\
        Learning Rate & \multicolumn{5}{c}{1e-6}\\
        LR Scheduler & \multicolumn{5}{c}{Cosine with Min LR} \\
        Warmup Ratio & \multicolumn{5}{c}{0.1} \\
        Precision & \multicolumn{5}{c}{BF16-mixed} \\
        \midrule
        Gradient Accumulation Step & \multicolumn{5}{c}{4} \\
        Total Train Batch Size & \multicolumn{5}{c}{32} \\
        Epochs & \multicolumn{5}{c}{1} \\
        Hardware & \multicolumn{5}{c}{2 $\times$ NVIDIA L40S} \\
        \midrule
        Max Prompt Length & \multicolumn{5}{c}{512} \\
        Max Completion Length & \multicolumn{5}{c}{3584} \\
        Number of Generation & \multicolumn{5}{c}{4} \\
        Vllm GPU Memory Utilization & \multicolumn{5}{c}{0.4} \\
        Vllm Max Model Length & \multicolumn{5}{c}{4608} \\
        \bottomrule
    \end{tabular}
    \vspace{3mm}
    \caption{Common hyperparameter settings.}
    \label{tab:common_hyperparameter}
\end{table}

We also show the varied hyperparameter in Table~\ref{tab:varied_hyperparameter} for all the LoRA-based RL experiments. Particularly, all the reward types including Accuracy, Format, Length, Cosine, Tag Count, Reasoning Steps, Repetition Penalty, are defined and implemented by the OpenR1 code repository.\footnote{\href{https://github.com/huggingface/open-r1}{https://github.com/huggingface/open-r1}} 

\begin{sidewaystable}[h]
    \small  
    \centering
    \begin{tabular}{lcccccccc}
        \toprule
        Model & LoRA Rank & LoRA Alpha & LoRA Dropout & Algorithm & Learning Rate & Reward Type & Reward Weights \\
        \midrule
        Tina-STILL-3-1.5B-preview & - & - & - & - & - & Accuracy, Length & 2, 1 \\
        Tina-DeepScaleR-1.5B-Preview & - & - & - & - & - & Accuracy, Format & 2, 1 \\
        Tina-Open-RS3 & - & - & - & - & - & Cosine, Format & 2, 1 \\
        Tina-Open-RS3-DrGRPO & - & - & - & DrGRPO & - & Cosine, Format & 2, 1 \\
        Tina-Open-RS2 & - & - & - & - & - & Accuracy, Format & 2, 1 \\
        Tina-Open-RS1 & - & - & - & - & - & Accuracy, Format & 2, 1 \\
        Tina-LIMR & - & - & - & - & - & Accuracy, Format & 2, 1 \\
        Tina-LIMR-5e-6-lr & - & - & - & - & 5e-6 & Accuracy, Format & 2, 1 \\
        Tina-LIMR-5e-7-lr & - & - & - & - & 5e-7 & Accuracy, Format & 2, 1 \\
        Tina-LIMR-64-LoRA-rank & 64 & 256 & - & - & - & Accuracy, Format & 2, 1 \\
        Tina-LIMR-16-LoRA-rank & 16 & 64 & - & - & - & Accuracy, Format & 2, 1 \\
        Tina-LIMR-8-LoRA-rank & 8 & 32 & - & - & - & Accuracy, Format & 2, 1 \\
        Tina-LIMR-4-LoRA-rank & 4 & 16 & - & - & - & Accuracy, Format & 2, 1 \\
        Tina-OpenR1 & - & - & - & - & - &  \makecell[c]{Accuracy, Cosine,\\ Format, Length,\\ Tag Count,\\ Reasoning Steps,\\ Repetition Penalty} & \makecell[c]{1, 1, 1, 1,\\ 1, 1, 1} \\
        Tina-OpenThoughts & - & - & - & - & - &  \makecell[c]{Accuracy, Cosine,\\ Format, Length,\\ Tag Count,\\ Reasoning Steps,\\ Repetition Penalty} & \makecell[c]{1, 1, 1, 1,\\ 1, 1, 1} \\
        \bottomrule
    \end{tabular}
    \vspace{3mm}
    \caption{Varied hyperparameter settings where ``-'' means unchanged from the common settings in Table~\ref{tab:common_hyperparameter}.}
    \label{tab:varied_hyperparameter}
\end{sidewaystable}

\clearpage

\subsection{Evaluation Command}
\label{app:eval_command}

The following is the evaluation command we use to combine \texttt{lighteval} and \texttt{vLLM} for performance evaluation on reasoning tasks. The \texttt{MODEL\_PATH} should be replaced with either the local path or huggingface identifier to the model to be evaluated. \texttt{TASK} should be one of the six reasoning tasks including \texttt{aime24}, \texttt{aime25}, \texttt{amc23}, \texttt{math\_500}, \texttt{gpqa:diamond}, and \texttt{minerva}. \texttt{PATH\_TO\_OPEN\_R1\_EVALUATE\_SCRIPT} should be the path to the custom evaluate script provided by \texttt{OpenR1}.\footnote{\href{https://github.com/huggingface/open-r1/blob/4f5b21e21dec473af9729bce8e084deb16223ae4/src/open\_r1/evaluate.py}{https://github.com/huggingface/open-r1/blob/4f5b21e21dec473af9729bce8e084deb16223ae4/src/open\_r1/evaluate.py}}

\begin{verbatim}
MODEL_ARGS="pretrained=$MODEL_PATH,
            dtype=bfloat16,
            data_parallel_size=2,
            max_model_length=32768,
            gpu_memory_utilization=0.5,
            generation_parameters={max_new_tokens:32768,temperature:0.6,top_p:0.95}"

lighteval vllm $MODEL_ARGS "custom|$TASK|0|0"
    --custom-tasks $PATH_TO_OPEN_R1_EVALUATE_SCRIPT
    --use-chat-template
\end{verbatim}

\clearpage
\section{Full Tina Model Performance Evaluation}
\label{app:full_tina_eval}

In this section, we present all Tina models' detailed evaluation performance during post-training across six reasoning tasks including AIME24/25, AMC23, MATH500, GPQA and Minerva.

\begin{table}[h]
\centering
\resizebox{0.9\textwidth}{!}{
\begin{tabular}{c|cccccc|cc}
\toprule
\rowcolor{LightPink}
\textbf{\textsc{Checkpoint Steps (3740 Steps per Epoch)}} & \textbf{\textsc{AIME24}} & \textbf{\textsc{AIME25}} & \textbf{\textsc{AMC23}} & \textbf{\textsc{MATH500}} & \textbf{\textsc{GPQA}} & \textbf{\textsc{Minerva}} & \textbf{\textsc{Avg.}} \\
\midrule
500 & 30.00 & 13.33 & 75.00 & 83.60 & 35.86 & \textbf{32.35} & 45.02 \\
\midrule
1000 & \textbf{36.67} & 20.00 & 65.00 & \textbf{84.80} & 32.32 & 27.94 & 44.46 \\
\midrule
1500 & 26.67 & 20.00 & 70.00 & 83.80 & \textbf{37.37} & 26.84 & 44.11 \\
\midrule
2000 & \textbf{36.67} & \textbf{30.00} & \textbf{77.50} & 84.60 & 33.33 & 26.84 & \textbf{48.16} \\
\midrule
2500 & 33.33 & \textbf{30.00} & 70.00 & 83.00 & 35.35 & 27.57 & 46.54 \\
\midrule
3000 & 30.00 & 20.00 & 67.50 & 82.60 & 30.81 & 25.74 & 42.78 \\
\midrule
3500 & 30.00 & 26.67 & 67.50 & 82.20 & 32.32 & 26.10 & 44.13 \\
\bottomrule
\end{tabular}
}
\caption{Performance evaluation of \texttt{Tina-STILL-3-1.5B-preview}.}
\label{tab:tina_still_eval}
\end{table}

\begin{table}[h]
\centering
\resizebox{0.9\textwidth}{!}{
\begin{tabular}{c|cccccc|cc}
\toprule
\rowcolor{LightPink}
\textbf{\textsc{Checkpoint Steps (5039 Steps per Epoch)}} & \textbf{\textsc{AIME24}} & \textbf{\textsc{AIME25}} & \textbf{\textsc{AMC23}} & \textbf{\textsc{MATH500}} & \textbf{\textsc{GPQA}} & \textbf{\textsc{Minerva}} & \textbf{\textsc{Avg.}} \\
\midrule
500 & 30.00 & 23.33 & 67.50 & 82.40 & 39.39 & \textbf{31.25} & 45.65 \\
\midrule
1000 & \textbf{43.33} & \textbf{26.67} & 67.50 & \textbf{86.20} & 37.88 & 28.68 & \textbf{48.38} \\
\midrule
1500 & 30.00 & 20.00 & \textbf{80.00} & 84.80 & 32.83 & 29.41 & 46.17 \\
\midrule
2000 & 20.00 & \textbf{26.67} & 57.50 & 80.60 & 29.29 & 24.26 & 39.72 \\
\midrule
2500 & 13.33 & 16.67 & 52.50 & 75.00 & 31.31 & 18.01 & 34.47 \\
\midrule
3000 & 26.67 & 16.67 & 57.50 & 78.60 & 28.79 & 23.16 & 38.57 \\
\midrule
3500 & 23.33 & 23.33 & 62.50 & 80.40 & 31.82 & 24.26 & 40.94 \\
\midrule
4000 & 20.00 & 20.00 & 70.00 & 82.00 & \textbf{41.41} & 27.94 & 43.56 \\
\midrule
4500 & 23.33 & 20.00 & 72.50 & 80.80 & 34.85 & 26.47 & 42.99 \\
\midrule
5000 & 20.00 & \textbf{26.67} & 75.00 & 80.80 & 33.33 & 29.41 & 44.20 \\
\bottomrule
\end{tabular}
}
\caption{Performance evaluation of \texttt{Tina-DeepScaleR-1.5B-Preview}.}
\label{tab:tina_deepscaler_eval}
\end{table}

\begin{table}[h]
\centering
\resizebox{0.9\textwidth}{!}{
\begin{tabular}{c|cccccc|cc}
\toprule
\rowcolor{LightPink}
\textbf{\textsc{Checkpoint Steps (875 Steps per Epoch)}} & \textbf{\textsc{AIME24}} & \textbf{\textsc{AIME25}} & \textbf{\textsc{AMC23}} & \textbf{\textsc{MATH500}} & \textbf{\textsc{GPQA}} & \textbf{\textsc{Minerva}} & \textbf{\textsc{Avg.}} \\
\midrule
50 & 26.67 & 23.33 & 75.00 & 84.20 & 37.37 & 29.04 & 45.94 \\
\midrule
100 & 30.00 & \textbf{30.00} & 65.00 & 83.00 & 37.37 & 29.78 & 45.86 \\
\midrule
150 & 36.67 & 16.67 & 65.00 & 84.80 & 27.78 & 27.94 & 43.14 \\
\midrule
200 & 20.00 & 26.67 & 70.00 & 83.80 & 33.33 & 27.94 & 43.62 \\
\midrule
250 & 36.67 & 20.00 & 65.00 & 84.60 & 38.38 & 28.31 & 45.49 \\
\midrule
300 & 33.33 & 26.67 & 70.00 & 85.20 & 30.81 & 30.15 & 46.03 \\
\midrule
350 & \textbf{40.00} & 16.67 & 77.50 & 84.40 & 39.90 & 27.94 & 47.74 \\
\midrule
400 & 30.00 & 16.67 & 70.00 & 82.80 & 35.86 & 31.25 & 44.43 \\
\midrule
450 & 36.67 & 26.67 & 70.00 & 85.60 & 33.84 & \textbf{32.72} & 47.58\\
\midrule
500 & 36.67 & 23.33 & \textbf{82.50} & 85.20 & 37.37 & 31.62 & \textbf{49.45} \\
\midrule
550 & 26.67 & 16.67 & 80.00 & \textbf{86.00} & 35.35 & 29.78 & 45.75 \\
\midrule
600 & 30.00 & 26.67 & 70.00 & 84.60 & 37.88 & 29.78 & 46.49 \\
\midrule
650 & 20.00 & 23.33 & 80.00 & 85.00 & 33.33 & 27.94 & 44.93 \\
\midrule
700 & 33.33 & 13.33 & 72.50 & 85.00 & \textbf{40.40} & 31.99 & 46.09 \\
\midrule
750 & 33.33 & 23.33 & 75.00 & 83.60 & 31.31 & 27.57 & 45.69 \\
\midrule
800 & 30.00 & 23.33 & 65.00 & 84.20 & 38.38 & 29.04 & 44.99 \\
\midrule
850 & 26.67 & 26.67 & 75.00 & 83.80 & 31.82 & 27.94 & 45.32 \\
\bottomrule
\end{tabular}
}
\caption{Performance evaluation of \texttt{Tina-Open-RS3}.}
\label{tab:tina_openrs3_eval}
\end{table}

\begin{table}[h]
\centering
\resizebox{0.9\textwidth}{!}{
\begin{tabular}{c|cccccc|cc}
\toprule
\rowcolor{LightPink}
\textbf{\textsc{Checkpoint Steps (875 Steps per Epoch)}} & \textbf{\textsc{AIME24}} & \textbf{\textsc{AIME25}} & \textbf{\textsc{AMC23}} & \textbf{\textsc{MATH500}} & \textbf{\textsc{GPQA}} & \textbf{\textsc{Minerva}} & \textbf{\textsc{Avg.}} \\
\midrule
50 & 33.33 & 23.33 & \textbf{77.50} & 84.20 & 38.89 & 29.04 & 47.72 \\
\midrule
100 & 36.67 & 23.33 & 72.50 & 84.20 & 31.31 & 28.68 & 46.12 \\
\midrule
150 & 40.00 & 23.33 & 72.50 & 85.80 & 30.30 & 30.51 & 47.07 \\
\midrule
200 & 26.67 & 23.33 & 70.00 & 83.80 &  39.39 & 29.41 & 45.43 \\
\midrule
250 & \textbf{46.67} & 13.33 & 72.50 & 82.60 & 31.82 & 30.51 & 46.24 \\
\midrule
300 & 30.00 & \textbf{26.67} & 75.00 & 84.00 & 33.33 & 29.04 & 46.34 \\
\midrule
350 & 33.33 & 20.00 & 75.00 & 84.80 & 37.37 & 28.68 & 46.53 \\
\midrule
400 & 26.67 & 16.67 & 70.00 & 83.20 & 37.37 & 27.57 & 43.58 \\
\midrule
450 & 43.33 & \textbf{26.67} & \textbf{77.50} & \textbf{87.00} & 36.36 & \textbf{32.72} & \textbf{50.60} \\
\midrule
500 & 20.00 & 23.33 & 67.50 & 84.20 & 33.84 & 29.41 & 43.05 \\
\midrule
550 & 40.00 & 23.33 & 72.50 & 83.60 & \textbf{40.91} & 30.88 & 48.54 \\
\midrule
600 & 33.33 & 20.00 & 72.50 & 84.20 & 32.83 & 30.88 & 45.62 \\
\midrule
650 & 33.33 & 23.33 & 57.50 & 83.80 & 34.85 & 30.51 & 43.89 \\
\midrule
700 & 23.33 & \textbf{26.67} & 70.00 & 82.40 & 33.33 & 28.68 & 44.07 \\
\midrule
750 & 30.00 & 23.33 & 72.50 & 84.20 & 38.89 & 29.04 & 46.33 \\
\midrule
800 & 30.00 & \textbf{26.67} & 75.00 & 84.40 & 32.32 & 29.41 & 46.30 \\
\midrule
850 & 26.67 & 23.33 & 70.00 & 83.80 & 35.86 & 28.68 & 44.72 \\
\bottomrule
\end{tabular}
}
\caption{Performance evaluation of \texttt{Tina-Open-RS2}.}
\label{tab:tina_openrs2_eval}
\end{table}

\begin{table}[h]
\centering
\resizebox{0.9\textwidth}{!}{
\begin{tabular}{c|cccccc|cc}
\toprule
\rowcolor{LightPink}
\textbf{\textsc{Checkpoint Steps (2327 Steps per Epoch)}} & \textbf{\textsc{AIME24}} & \textbf{\textsc{AIME25}} & \textbf{\textsc{AMC23}} & \textbf{\textsc{MATH500}} & \textbf{\textsc{GPQA}} & \textbf{\textsc{Minerva}} & \textbf{\textsc{Avg.}} \\
\midrule
400 & 33.33 & 20.00 & 75.00 & 83.80 & 31.82 & 29.78 & 45.62 \\
\midrule
600 & 30.00 & \textbf{30.00} & 77.50 & 84.20 & 34.34 & \textbf{31.62} & 47.94 \\
\midrule
800 & \textbf{43.33} & 20.00 & 80.00 & 84.00 & 35.35 & 28.68 & \textbf{48.56} \\
\midrule
1000 & 33.33 & 20.00 & \textbf{82.50} & \textbf{84.40} & 35.86 & 29.78 & 47.64 \\
\midrule
1200 & 36.67 & 20.00 & 67.50 & \textbf{84.40} & \textbf{37.88} & 30.15 & 46.10 \\
\midrule
1400 & 30.00 & 20.00 & 67.50 & 83.40 & 31.82 & 29.78 & 43.75 \\
\midrule
1600 & 23.33 & 13.33 & 65.00 & 83.40 & 35.86 & 26.84 & 41.29 \\
\midrule
1800 & 26.67 & 20.00 & 75.00 & 84.20 & 34.34 & 27.57 & 44.63 \\
\midrule
2000 & 30.00 & 26.67 & 72.50 & 83.00 & 36.36 & 27.94 & 46.08 \\
\midrule
2200 & 30.00 & 23.33 & 70.00 & 81.40 & 30.81 & 26.47 & 43.67 \\
\midrule
2400 & 30.00 & 23.33 & 67.50 & 81.80 & 30.30 & 27.57 & 43.42 \\
\bottomrule
\end{tabular}
}
\caption{Performance evaluation of \texttt{Tina-Open-RS1}.}
\label{tab:tina_openrs1_eval}
\end{table}

\begin{table}[h]
\centering
\resizebox{0.9\textwidth}{!}{
\begin{tabular}{c|cccccc|cc}
\toprule
\rowcolor{LightPink}
\textbf{\textsc{Checkpoint Steps (174 Steps per Epoch)}} & \textbf{\textsc{AIME24}} & \textbf{\textsc{AIME25}} & \textbf{\textsc{AMC23}} & \textbf{\textsc{MATH500}} & \textbf{\textsc{GPQA}} & \textbf{\textsc{Minerva}} & \textbf{\textsc{Avg.}} \\
\midrule
50 & 20.00 & 26.67 & 67.50 & \textbf{85.40} & \textbf{37.88} & 30.51 & 44.66 \\
\midrule
100 & \textbf{46.67} & 20.00 & \textbf{75.00} & 83.80 & 34.85 & 30.51 & \textbf{48.47} \\
\midrule
150 & 26.67 & 20.00 & 72.50 & 84.00 & 37.37 & 30.15 & 45.12 \\
\midrule
200 & 33.33 & \textbf{30.00} & 62.50 & 83.40 & 29.80 & \textbf{30.88} & 44.99 \\
\bottomrule
\end{tabular}
}
\caption{Performance evaluation of \texttt{Tina-LIMR}.}
\label{tab:tina_limr_eval}
\end{table}

\begin{table}[h]
\centering
\resizebox{0.9\textwidth}{!}{
\begin{tabular}{c|cccccc|cc}
\toprule
\rowcolor{LightPink}
\textbf{\textsc{Checkpoint Steps (11716 Steps per Epoch)}} & \textbf{\textsc{AIME24}} & \textbf{\textsc{AIME25}} & \textbf{\textsc{AMC23}} & \textbf{\textsc{MATH500}} & \textbf{\textsc{GPQA}} & \textbf{\textsc{Minerva}} & \textbf{\textsc{Avg.}} \\
\midrule
500 & 30.00 & 20.00 & \textbf{77.50} & 85.20 & 33.84 & 30.15 & 46.12 \\
\midrule
1000 & 30.00 & 23.33 & 72.50 & 85.60 & 33.84 & 26.67 & 45.32 \\
\midrule
1500 & \textbf{36.67} & 26.67 & 75.00 & \textbf{86.80} & \textbf{39.90} & 30.51 & \textbf{49.26} \\
\midrule
2000 & 26.67 & 23.33 & 67.50 & 83.20 & 29.80 & \textbf{31.62} & 43.69 \\
\midrule
2500 & 30.00 & 23.33 & 72.50 & 83.80 & 33.84 & 26.84 & 45.05 \\
\midrule
3000 & 20.00 & \textbf{30.00} & 67.50 & 84.60 & 34.34 & 28.31 & 44.13 \\
\midrule
3500 & \textbf{36.67} & 23.33 & 67.50 & 83.60 & 31.31 & 25.74 & 44.69 \\
\bottomrule
\end{tabular}
}
\caption{Performance evaluation of \texttt{Tina-OpenR1}.}
\label{tab:tina_openr1_eval}
\end{table}

\begin{table}[h]
\centering
\resizebox{0.9\textwidth}{!}{
\begin{tabular}{c|cccccc|cc}
\toprule
\rowcolor{LightPink}
\textbf{\textsc{Checkpoint Steps (8259 Steps per Epoch)}} & \textbf{\textsc{AIME24}} & \textbf{\textsc{AIME25}} & \textbf{\textsc{AMC23}} & \textbf{\textsc{MATH500}} & \textbf{\textsc{GPQA}} & \textbf{\textsc{Minerva}} & \textbf{\textsc{Avg.}} \\
\midrule
500 & 33.30 & 16.67 & 77.50 & 84.20 & 35.86 & 30.15 & 46.28 \\
\midrule
1000 & 33.33 & 23.33 & \textbf{80.00} & 85.20 & 24.75 & 32.72 & 46.56 \\
\midrule
1500 & 30.00 & 23.33 & 70.00 & \textbf{86.00} & 37.88 & 29.04 & 46.04 \\
\midrule
2000 & 30.00 & 23.33 & 70.00 & 84.20 & 33.33 & 28.31 & 44.86 \\
\midrule
2500 & \textbf{36.67} & 26.67 & 72.50 & 84.80 & \textbf{41.41} & \textbf{33.09} & \textbf{49.19} \\
\midrule
3000 & 26.67 & 23.33 & 75.00 & 83.60 & 34.34 & 32.72 & 45.94 \\
\midrule
3500 & 20.00 & 16.67 & 60.00 & 84.20 & 32.32 & 26.10 & 39.88 \\
\midrule
4000 & 33.33 & 23.33 & 72.50 & 83.60 & 38.38 & 27.94 & 46.51 \\
\midrule
4500 & 30.00 & 20.00 & 65.00 & 85.00 & 33.84 & 26.84 & 43.45 \\
\midrule
5000 & 20.00 & \textbf{33.33} & 65.00 & 84.80 & 40.91 & 30.88 & 45.82 \\
\bottomrule
\end{tabular}
}
\caption{Performance evaluation of \texttt{Tina-OpenThoughts}.}
\label{tab:tina_openthoughts_eval}
\end{table}

\begin{table}[h]
\centering
\resizebox{0.9\textwidth}{!}{
\begin{tabular}{c|cccccc|cc}
\toprule
\rowcolor{LightPink}
\textbf{\textsc{Checkpoint Steps (875 Steps per Epoch)}} & \textbf{\textsc{AIME24}} & \textbf{\textsc{AIME25}} & \textbf{\textsc{AMC23}} & \textbf{\textsc{MATH500}} & \textbf{\textsc{GPQA}} & \textbf{\textsc{Minerva}} & \textbf{\textsc{Avg.}} \\
\midrule
50 & 33.33 & 16.67 & 75.00 & 83.80 & 37.37 & 26.84 & 45.50 \\
\midrule
100 & 16.67 & 20.00 & 70.00 & 83.20 & 33.33 & 26.47 & 41.61 \\
\midrule
150 & \textbf{43.33} & 23.33 & \textbf{80.00} & 85.00 & 35.35 & 30.15 & \textbf{49.53} \\
\midrule
200 & 30.00 & 23.33 & 70.00 & 84.00 & \textbf{39.90} & 28.68 & 45.99 \\
\midrule
250 & 33.33 & \textbf{30.00} & 65.00 & 83.80 & 34.34 & 28.31 & 45.80 \\
\midrule
300 & 36.67 & 16.67 & 67.50 & 84.40 & 37.88 & 29.78 & 45.48 \\
\midrule
350 & 26.67 & \textbf{30.00} & 75.00 & 84.00 & 37.88 & 29.78 & 47.22 \\
\midrule
400 & 36.67 & 23.33 & 72.50 & 84.40 & 32.83 & 27.57 & 46.22 \\
\midrule
450 & 36.67 & 16.67 & 72.50 & \textbf{85.60} & 29.29 & 27.57 & 44.72 \\
\midrule
500 & 30.00 & 20.00 & 72.50 & \textbf{85.60} & 37.37 & 29.41 & 45.81 \\
\midrule
550 & 30.00 & 23.33 & 77.50 & 84.80 & 36.87 & \textbf{31.62} & 47.35 \\
\midrule
600 & 33.33 & 26.67 & 72.50 & 83.80 & 30.30 & 28.31 & 45.82 \\
\midrule
650 & 26.67 & 20.00 & 77.50 & 82.40 & 37.88 & 27.94 & 45.40 \\
\midrule
700 & 36.67 & 20.00 & \textbf{80.00} & 83.80 & 35.35 & 31.25 & 47.85 \\
\midrule
750 & 30.00 & 26.67 & 75.00 & 84.20 & 38.89 & 27.57 & 47.06 \\
\midrule
800 & 20.00 & \textbf{30.00} & 75.00 & 82.40 & 35.86 & 28.31 & 45.26 \\
\midrule
850 & 23.33 & 20.00 & 72.50 & 85.40 & 36.36 & 30.15 & 44.62 \\
\bottomrule
\end{tabular}
}
\caption{Performance evaluation of \texttt{Tina-Open-RS3-DrGRPO}.}
\label{tab:tina_openrs3_drgrpo_eval}
\end{table}

\begin{table}[h]
\centering
\resizebox{0.9\textwidth}{!}{
\begin{tabular}{c|cccccc|cc}
\toprule
\rowcolor{LightPink}
\textbf{\textsc{Checkpoint Steps (174 Steps per Epoch)}} & \textbf{\textsc{AIME24}} & \textbf{\textsc{AIME25}} & \textbf{\textsc{AMC23}} & \textbf{\textsc{MATH500}} & \textbf{\textsc{GPQA}} & \textbf{\textsc{Minerva}} & \textbf{\textsc{Avg.}} \\
\midrule
50 & 20.00 & 26.67 & 67.50 & \textbf{85.40} & \textbf{37.88} & 30.51 & 44.66 \\
\midrule
100 & \textbf{46.67} & 20.00 & \textbf{75.00} & 83.80 & 34.85 & 30.51 & \textbf{48.47} \\
\midrule
150 & 26.67 & 20.00 & 72.50 & 84.00 & 37.37 & 30.15 & 45.12 \\
\midrule
200 & 33.33 & \textbf{30.00} & 62.50 & 83.40 & 29.80 & \textbf{30.88} & 44.99 \\
\bottomrule
\end{tabular}
}
\caption{Performance evaluation of \texttt{Tina-LIMR-5e-6-lr} with learning rate \texttt{5e-6}.}
\label{tab:tina_limr_lr_5e6_eval}
\end{table}

\begin{table}[h]
\centering
\resizebox{0.9\textwidth}{!}{
\begin{tabular}{c|cccccc|cc}
\toprule
\rowcolor{LightPink}
\textbf{\textsc{Checkpoint Steps (174 Steps per Epoch)}} & \textbf{\textsc{AIME24}} & \textbf{\textsc{AIME25}} & \textbf{\textsc{AMC23}} & \textbf{\textsc{MATH500}} & \textbf{\textsc{GPQA}} & \textbf{\textsc{Minerva}} & \textbf{\textsc{Avg.}} \\
\midrule
50 & 40.00 & 13.33 & 72.50 & 83.00 & 34.34 & 29.04 & 45.37 \\
\midrule
100 & \textbf{43.33} & 16.67 & \textbf{77.50} & 84.60 & 34.85 & 30.51 & \textbf{47.91} \\
\midrule
150 & 30.00 & \textbf{23.33} & 72.50 & \textbf{86.20} & \textbf{37.37} & 30.51 & 46.65 \\
\midrule
200 & 33.33 & 13.33 & 70.00 & 83.20 & 29.29 & \textbf{31.25} & 43.40 \\
\bottomrule
\end{tabular}
}
\caption{Performance evaluation of \texttt{Tina-LIMR-5e-7-lr} with learning rate \texttt{5e-7}.}
\label{tab:tina_limr_lr_5e7_eval}
\end{table}

\begin{table}[h]
\centering
\resizebox{0.9\textwidth}{!}{
\begin{tabular}{c|cccccc|cc}
\toprule
\rowcolor{LightPink}
\textbf{\textsc{Checkpoint Steps (174 Steps per Epoch)}} & \textbf{\textsc{AIME24}} & \textbf{\textsc{AIME25}} & \textbf{\textsc{AMC23}} & \textbf{\textsc{MATH500}} & \textbf{\textsc{GPQA}} & \textbf{\textsc{Minerva}} & \textbf{\textsc{Avg.}} \\
\midrule
50 & 20.00 & \textbf{30.00} & \textbf{77.50} & 84.20 & \textbf{38.38} & \textbf{31.62} & \textbf{46.95} \\
\midrule
100 & 30.00 & 23.33 & 72.50 & \textbf{84.60} & 32.32 & 29.78 & 45.42 \\
\midrule
150 & \textbf{36.67} & 20.00 & 70.00 & 83.40 & 31.82 & 30.88 & 45.46 \\
\midrule
200 & 33.33 & 20.00 & 72.50 & 85.00 & 29.80 & 29.41 & 45.01 \\
\bottomrule
\end{tabular}
}
\caption{Performance evaluation of \texttt{Tina-LIMR-64-LoRA-rank} with LoRA rank \texttt{64} and alpha \texttt{512}.}
\label{tab:tina_limr_rank_64_eval}
\end{table}

\clearpage

\begin{table}[h]
\centering
\resizebox{0.9\textwidth}{!}{
\begin{tabular}{c|cccccc|cc}
\toprule
\rowcolor{LightPink}
\textbf{\textsc{Checkpoint Steps (174 Steps per Epoch)}} & \textbf{\textsc{AIME24}} & \textbf{\textsc{AIME25}} & \textbf{\textsc{AMC23}} & \textbf{\textsc{MATH500}} & \textbf{\textsc{GPQA}} & \textbf{\textsc{Minerva}} & \textbf{\textsc{Avg.}} \\
\midrule
50 & 33.33 & 23.33 & 62.50 & \textbf{84.20} & \textbf{38.89} & \textbf{31.25} & 45.58 \\
\midrule
100 & \textbf{43.33} & \textbf{33.33} & 70.00 & 83.20 & 35.35 & 28.31 & \textbf{48.92} \\
\midrule
150 & 26.67 & 16.67 & 72.50 & 83.40 & 35.35 & 29.04 & 43.94 \\
\midrule
200 & 36.67 & 20.00 & \textbf{75.00} & 83.00 & 39.39 & 30.51 & 47.43 \\
\bottomrule
\end{tabular}
}
\caption{Performance evaluation of \texttt{Tina-LIMR-16-LoRA-rank} with LoRA rank \texttt{16} and alpha \texttt{64}.}
\label{tab:tina_limr_rank_16_eval}
\end{table}

\begin{table}[h]
\centering
\resizebox{0.9\textwidth}{!}{
\begin{tabular}{c|cccccc|cc}
\toprule
\rowcolor{LightPink}
\textbf{\textsc{Checkpoint Steps (174 Steps per Epoch)}} & \textbf{\textsc{AIME24}} & \textbf{\textsc{AIME25}} & \textbf{\textsc{AMC23}} & \textbf{\textsc{MATH500}} & \textbf{\textsc{GPQA}} & \textbf{\textsc{Minerva}} & \textbf{\textsc{Avg.}} \\
\midrule
50 & 30.00 & \textbf{26.67} & \textbf{82.50} & 83.80 & 33.84 & 30.51 & \textbf{47.89} \\
\midrule
100 & 26.67 & 16.67 & 72.50 & 84.00 & 36.87 & 29.78 & 44.42 \\
\midrule
150 & \textbf{53.33} & 20.00 & 60.00 & 83.20 & \textbf{37.37} & \textbf{30.88} & 47.46 \\
\midrule
200 & 23.33 & 20.00 & 72.50 & \textbf{85.40} & 32.83 & 28.68 & 43.86 \\
\bottomrule
\end{tabular}
}
\caption{Performance evaluation of \texttt{Tina-LIMR-8-LoRA-rank} with LoRA rank \texttt{8} and alpha \texttt{32}.}
\label{tab:tina_limr_rank_8_eval}
\end{table}

\begin{table}[h]
\centering
\resizebox{0.9\textwidth}{!}{
\begin{tabular}{c|cccccc|cc}
\toprule
\rowcolor{LightPink}
\textbf{\textsc{Checkpoint Steps (174 Steps per Epoch)}} & \textbf{\textsc{AIME24}} & \textbf{\textsc{AIME25}} & \textbf{\textsc{AMC23}} & \textbf{\textsc{MATH500}} & \textbf{\textsc{GPQA}} & \textbf{\textsc{Minerva}} & \textbf{\textsc{Avg.}} \\
\midrule
50 & 30.00 & 23.33 & 65.00 & 85.00 & 35.35 & \textbf{29.78} & 44.74 \\
\midrule
100 & 26.67 & \textbf{26.67} & 72.50 & 82.80 & 34.85 & 29.04 & 45.42 \\
\midrule
150 & \textbf{36.67} & 20.00 & \textbf{85.00} & 83.80 & 31.82 & 29.0 & \textbf{47.72} \\
\midrule
200 & 33.33 & 23.33 & 77.50 & \textbf{85.40} & \textbf{35.86} & 28.31 & 47.29 \\
\bottomrule
\end{tabular}
}
\caption{Performance evaluation of \texttt{Tina-LIMR-4-LoRA-rank} with LoRA rank \texttt{4} and alpha \texttt{16}.}
\label{tab:tina_limr_rank_4_eval}
\end{table}

\clearpage
\section{Full Tina Model Training Phase Transition}
\label{app:full_tina_phase_transit}

\begin{figure}[h!]
    \centering
    \includegraphics[width=.45\linewidth]{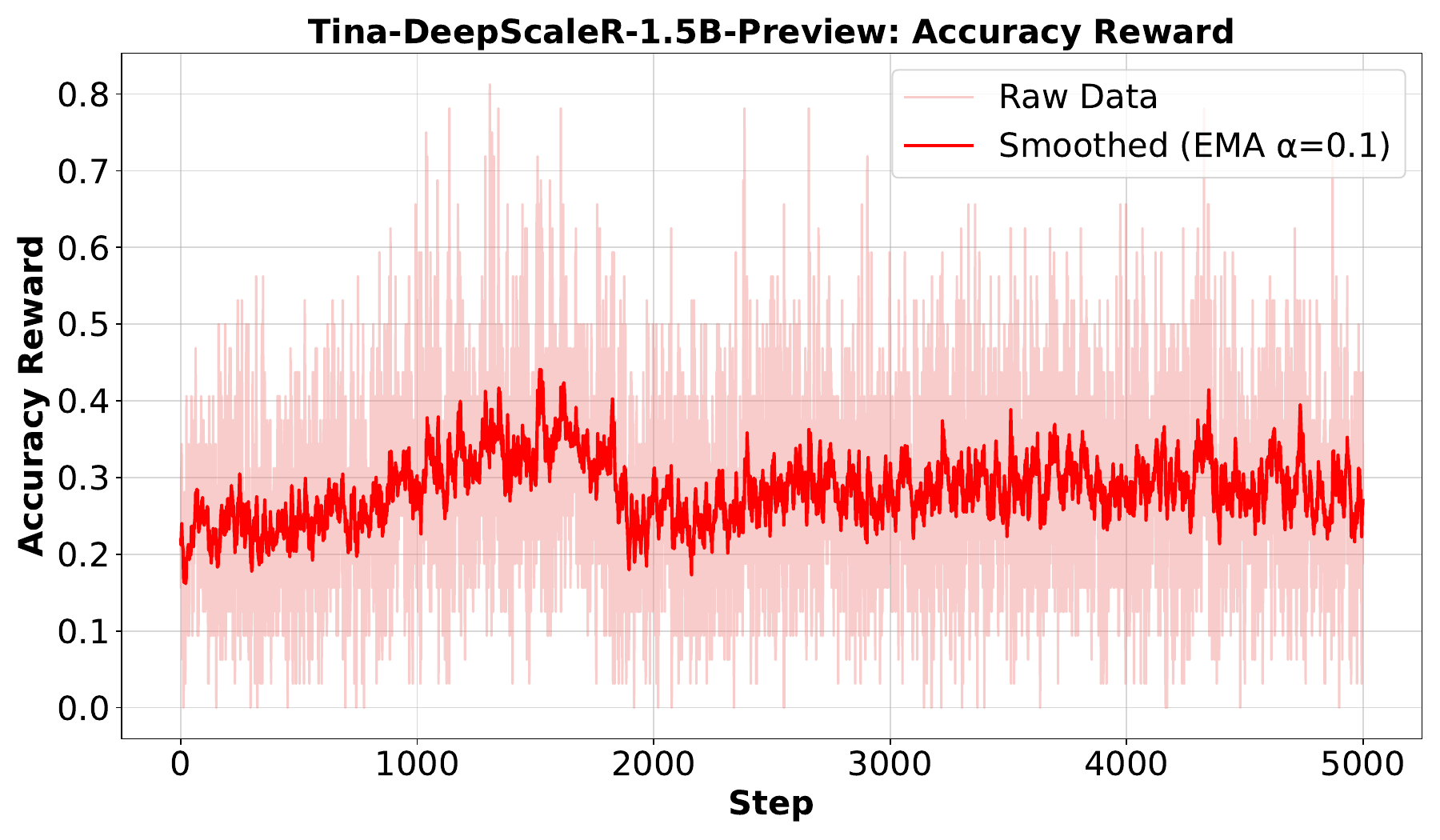}
    \includegraphics[width=.45\linewidth]{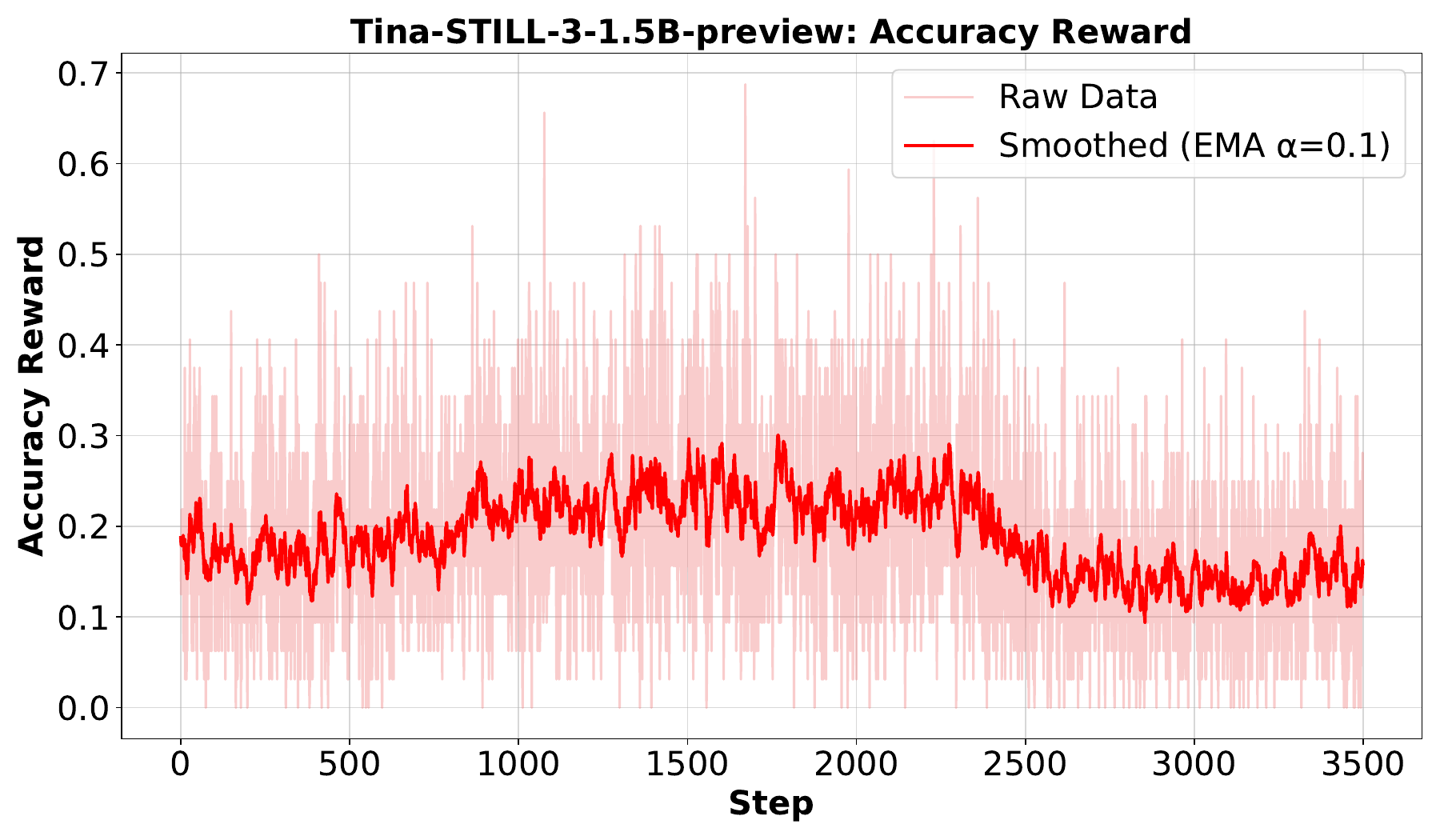}
    \includegraphics[width=.45\linewidth]{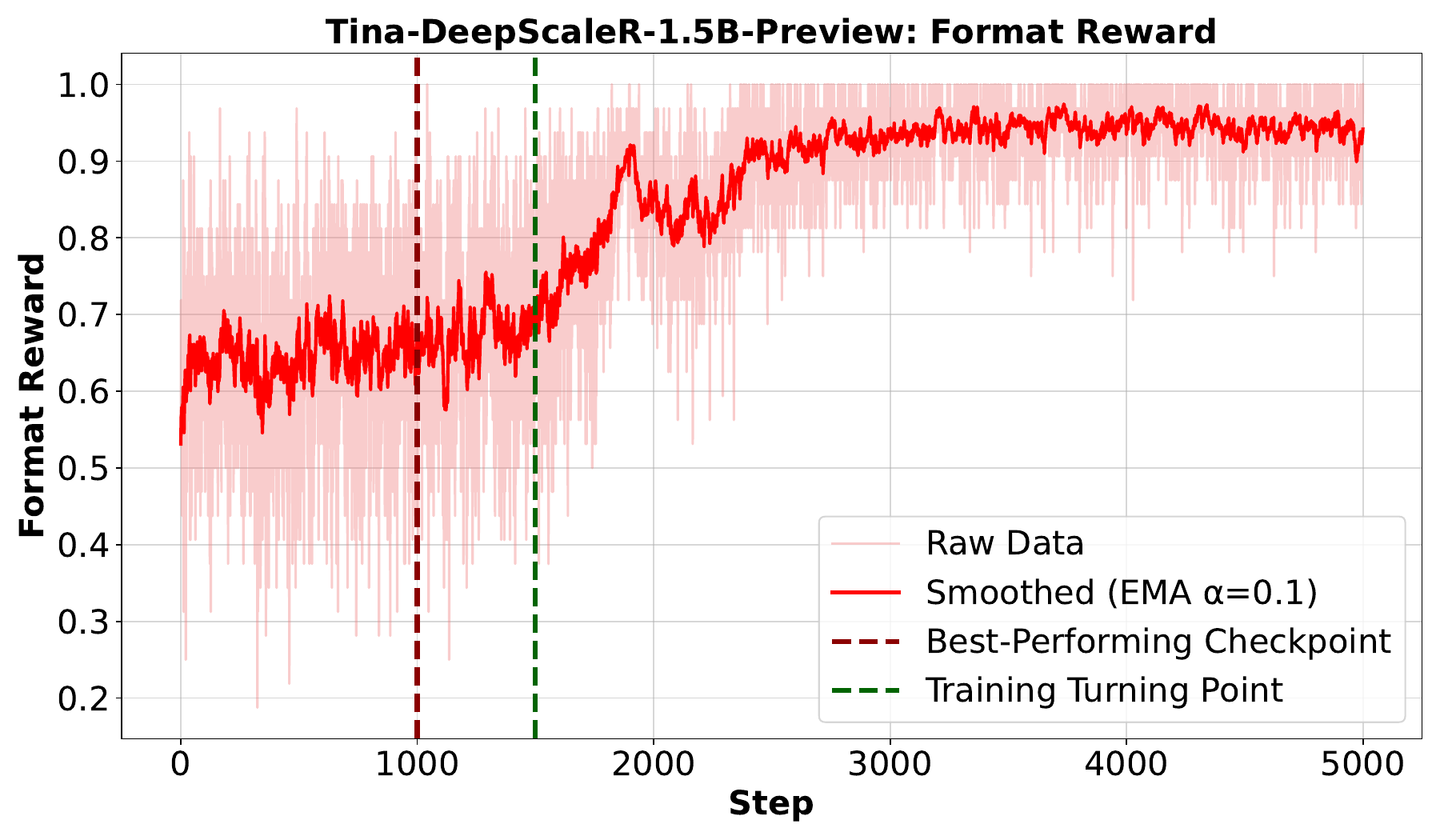}
    \includegraphics[width=.45\linewidth]{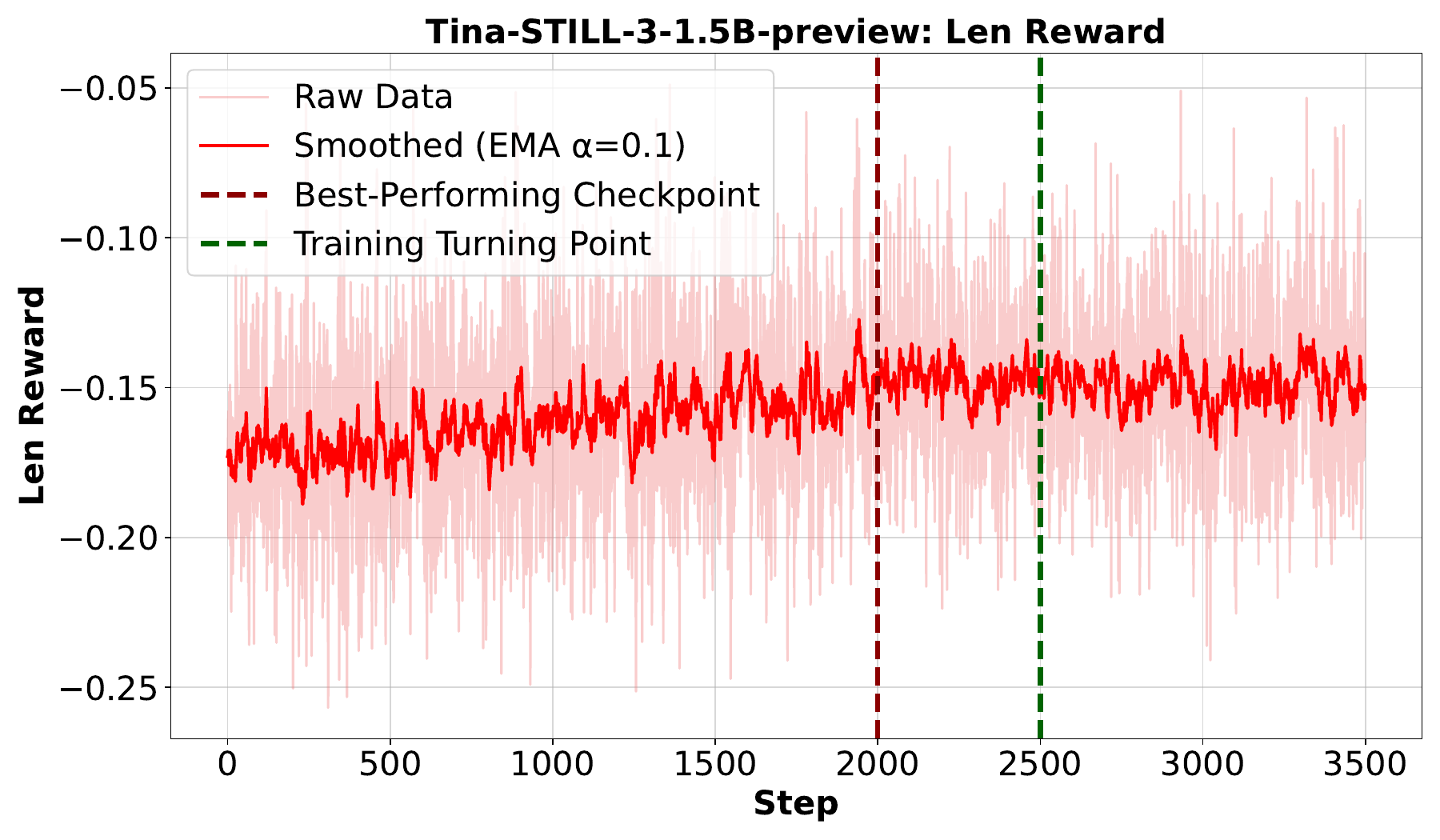}
    \includegraphics[width=.45\linewidth]{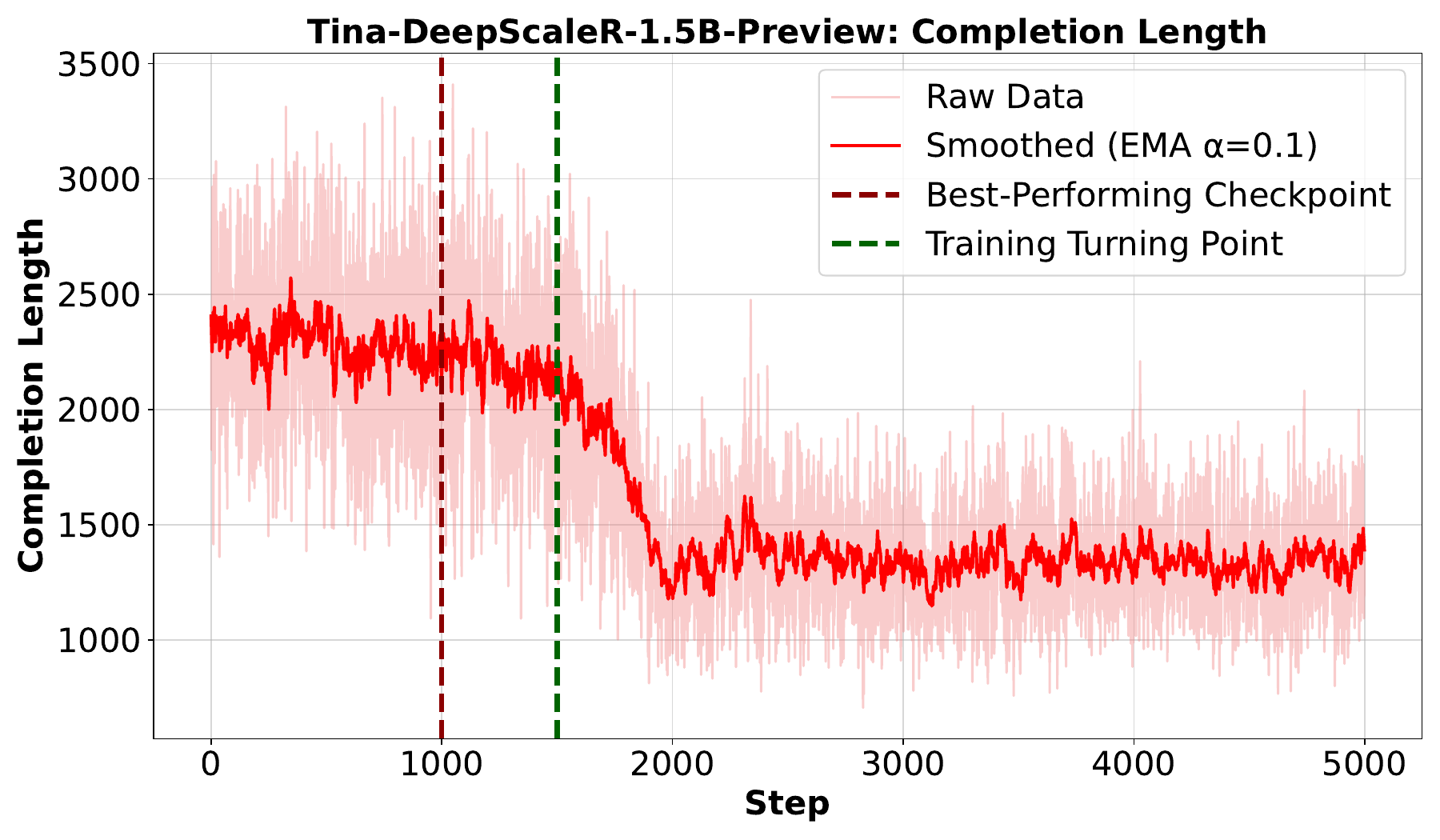}
    \includegraphics[width=.45\linewidth]{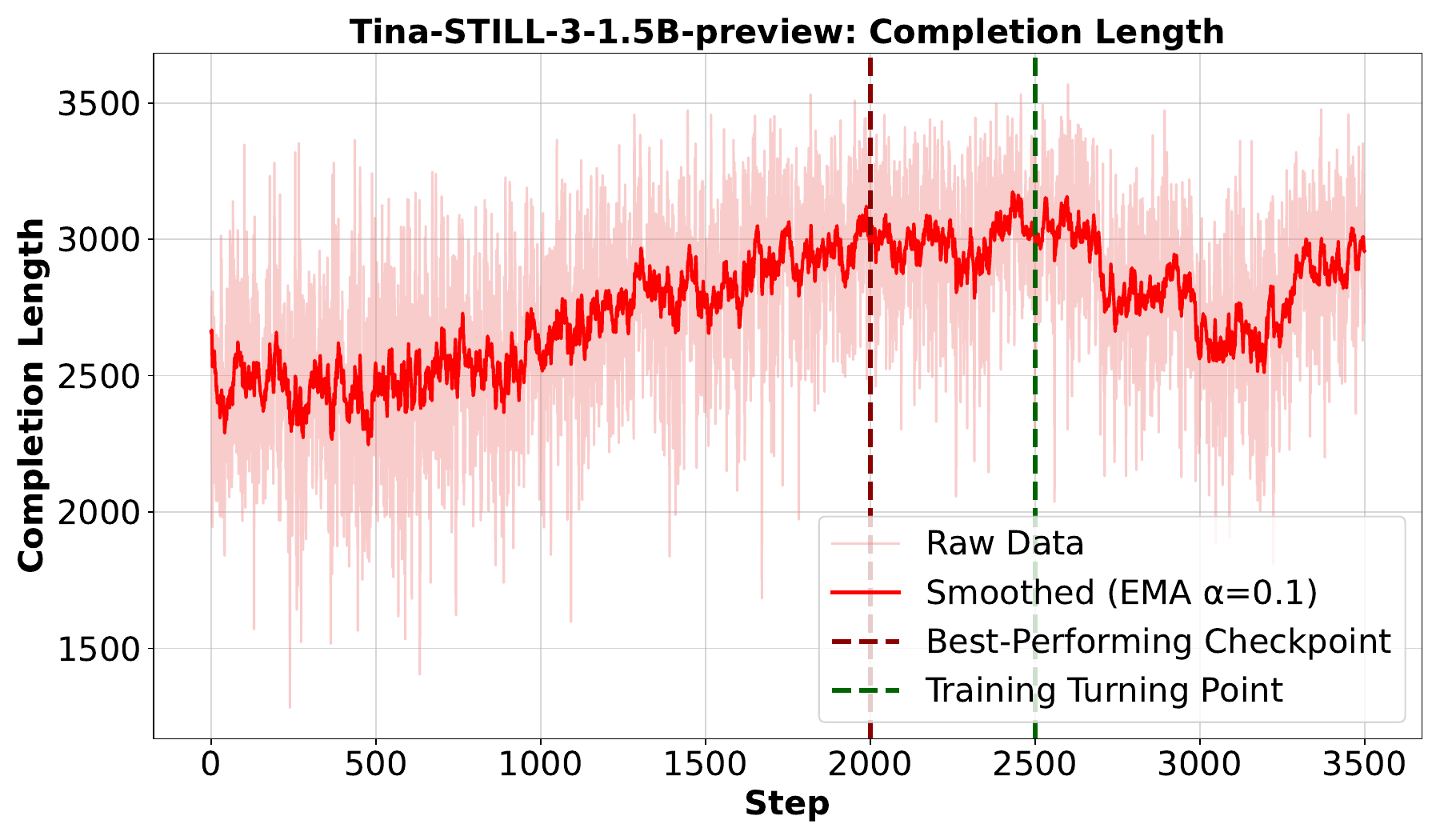}
    \caption{\textbf{Phase transition in \texttt{Tina-DeepScaleR-1.5B-Preview} and \texttt{Tina-STILL-3-1.5B-preview}.} The raw data is from the Weights \& Biases training logs and smoothed via exponential moving average (EMA) with factor $0.1$.}
    \label{fig:full_phase_transit_1}
\end{figure}

\begin{figure}[h!]
    \centering
    \includegraphics[width=.45\linewidth]{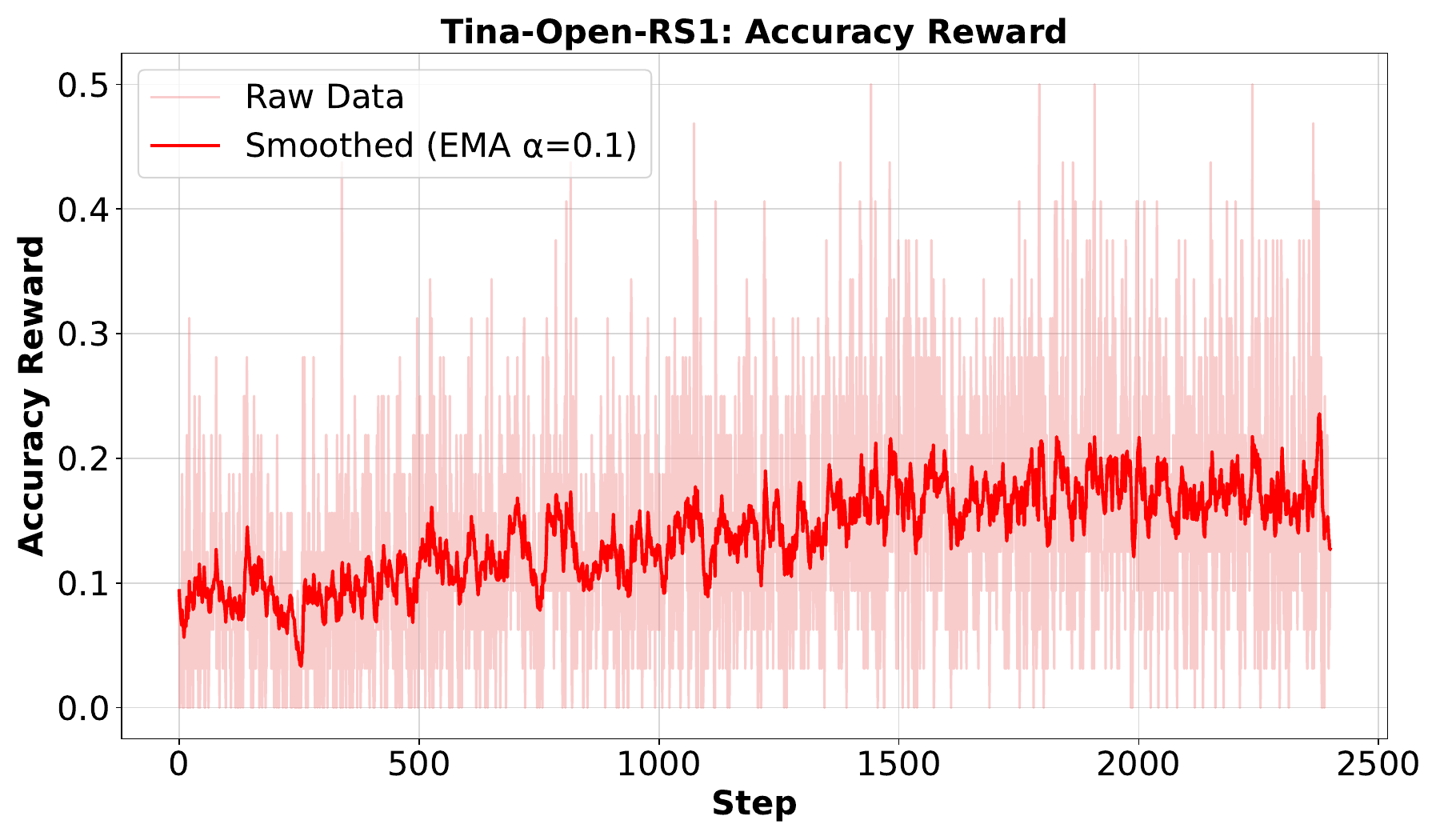}
    \includegraphics[width=.45\linewidth]{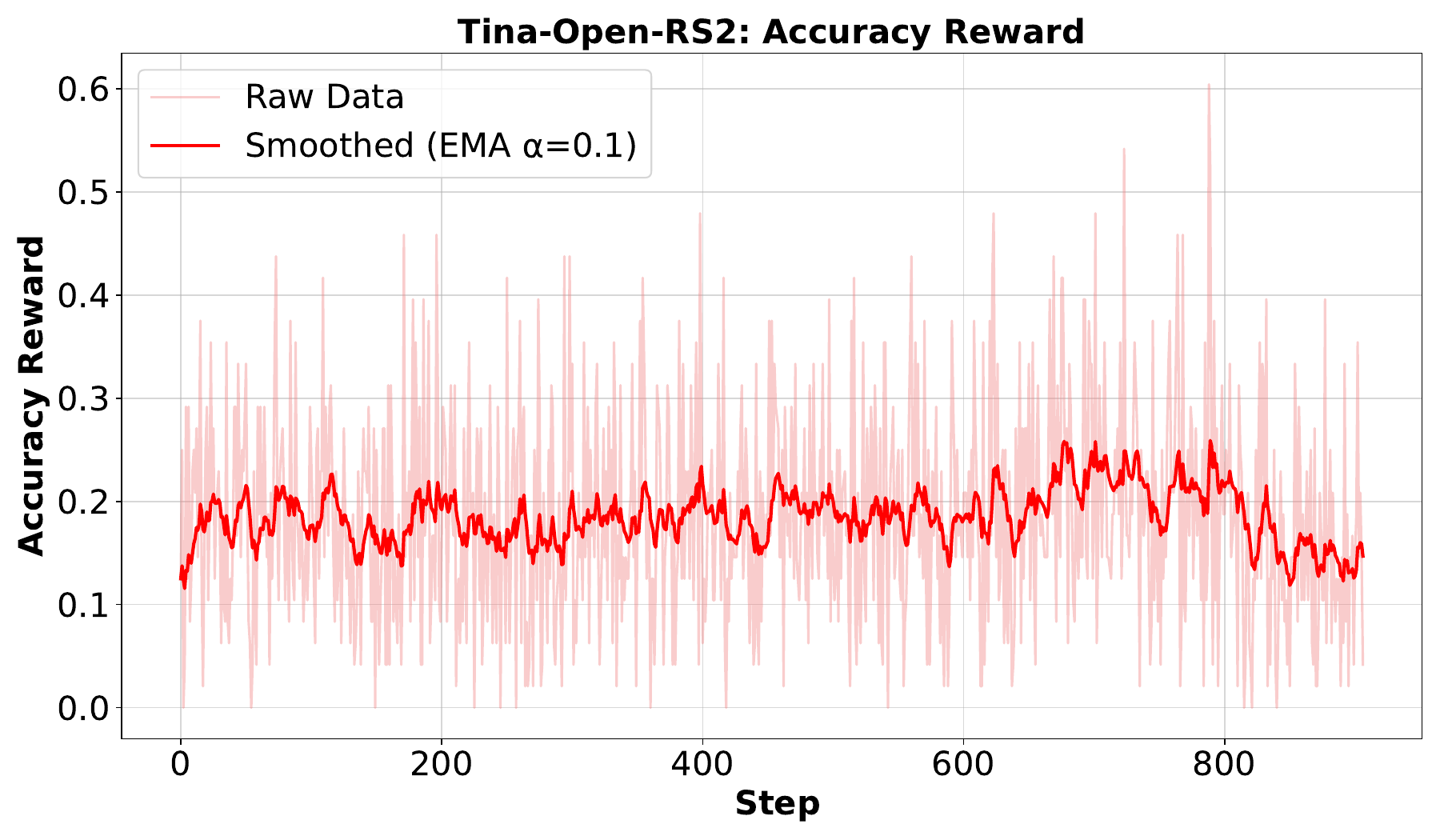}
    \includegraphics[width=.45\linewidth]{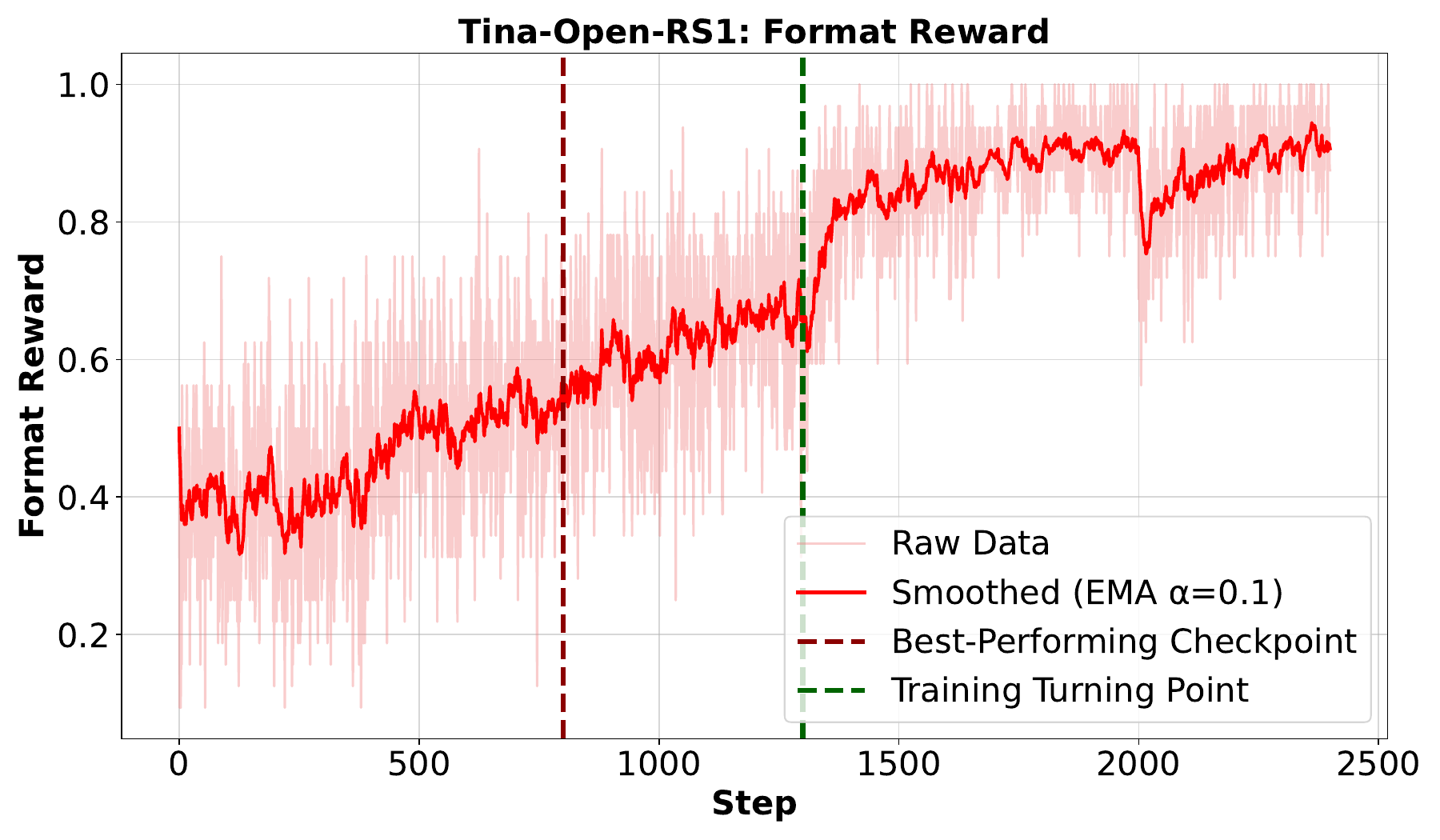}
    \includegraphics[width=.45\linewidth]{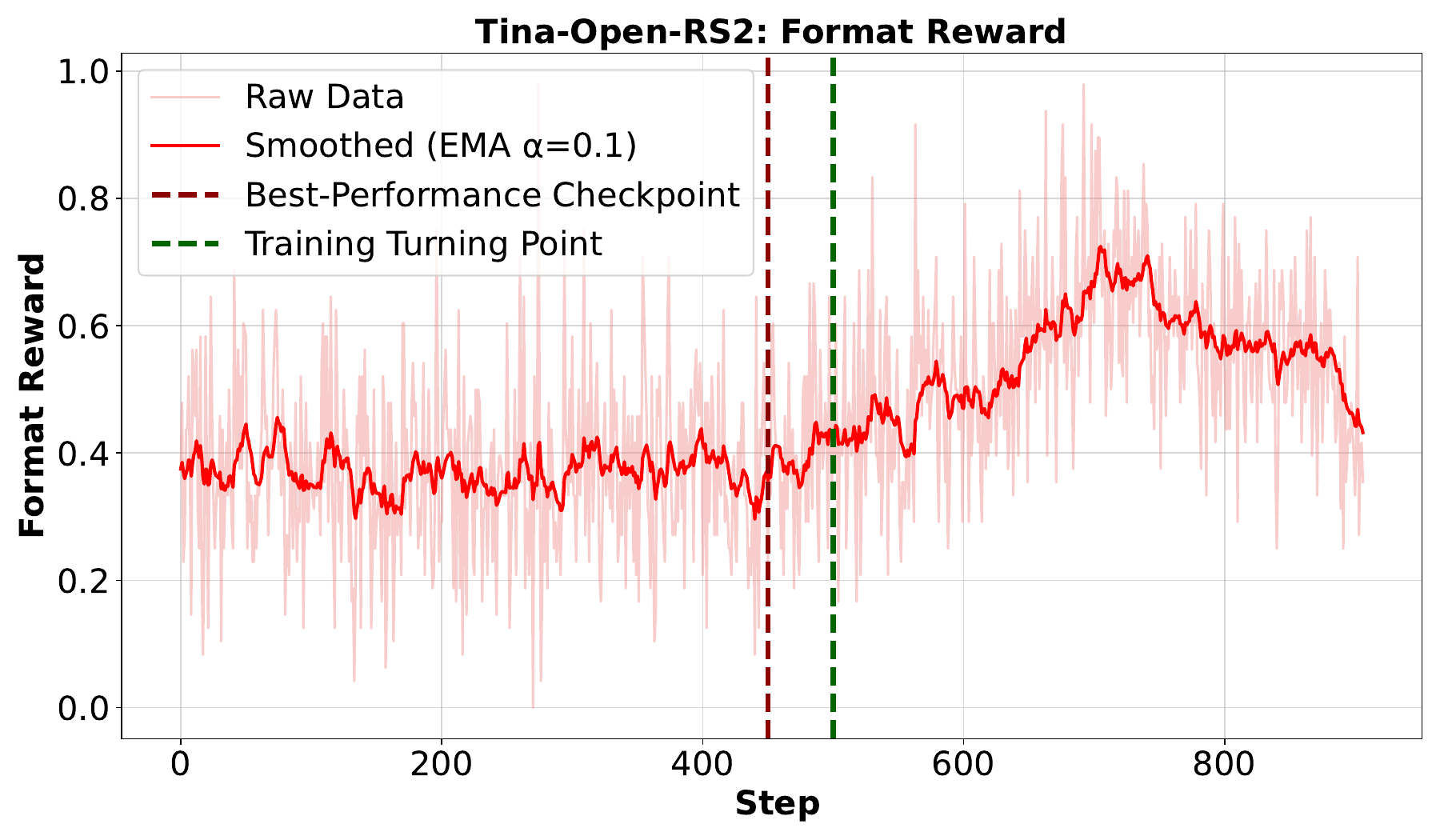}
    \includegraphics[width=.45\linewidth]{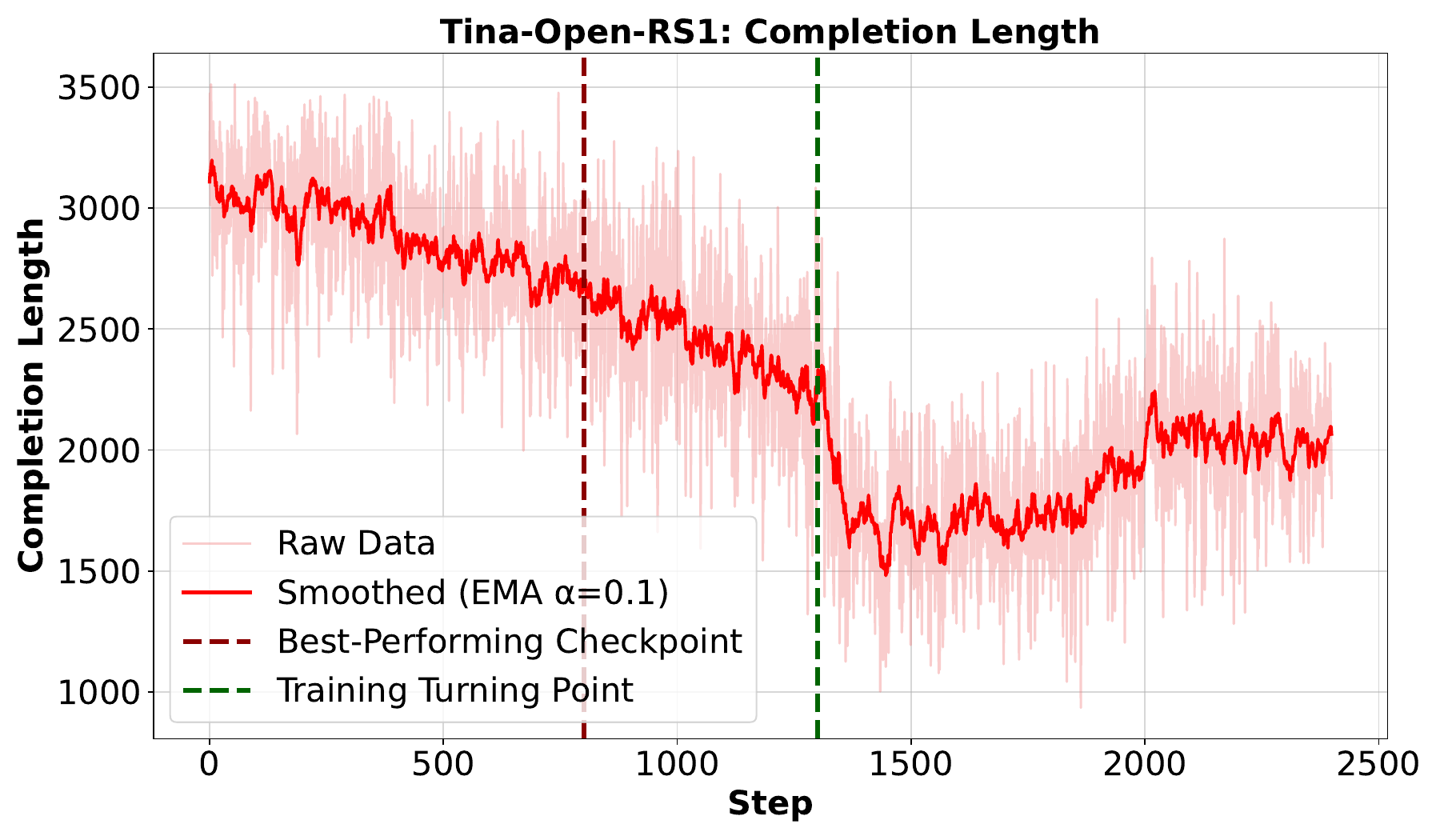}
    \includegraphics[width=.45\linewidth]{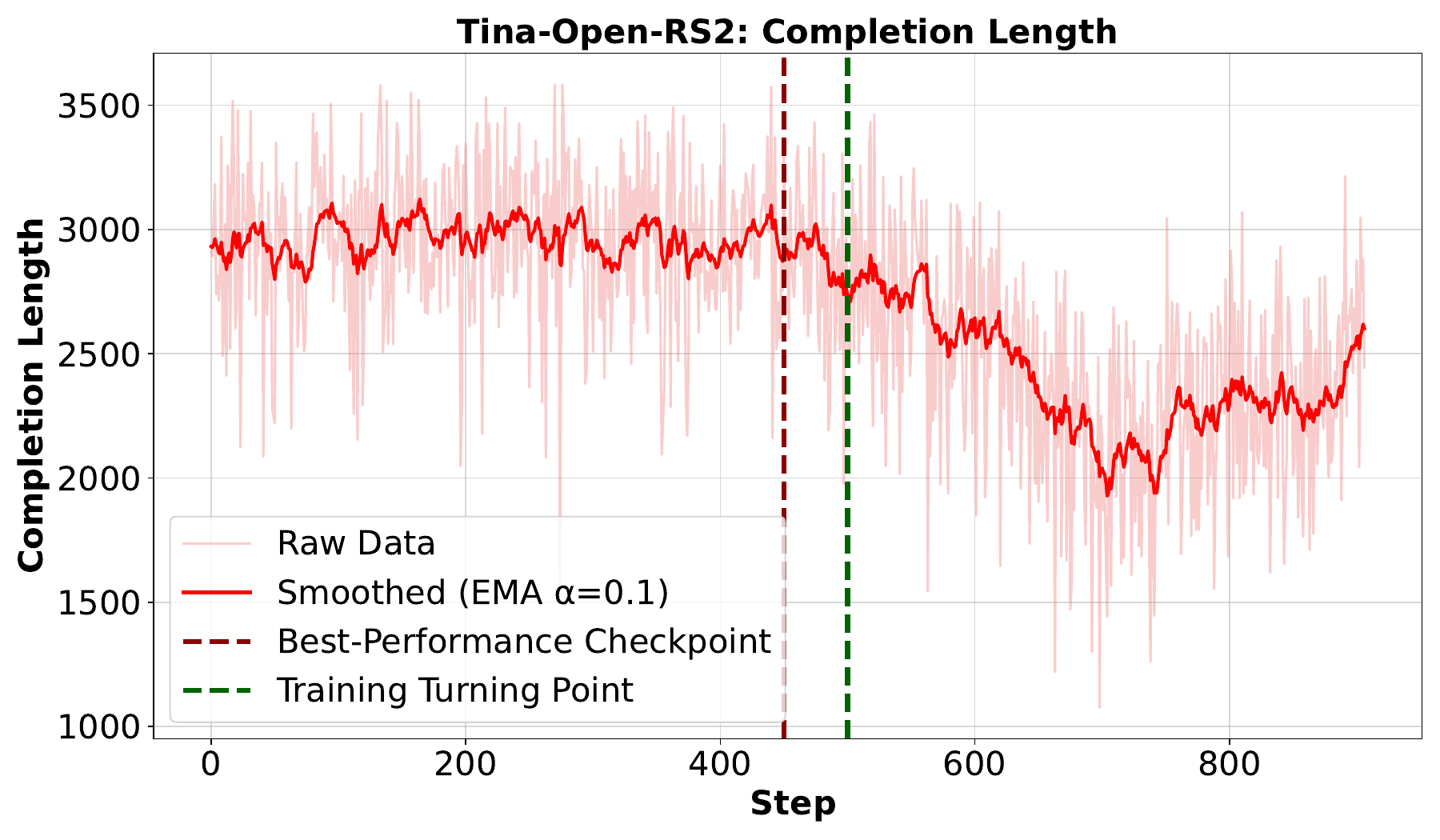}
    \caption{\textbf{Phase transition in \texttt{Tina-Open-RS1} and \texttt{Tina-Open-RS2}.} The raw data is from the Weights \& Biases training logs and smoothed via exponential moving average (EMA) with factor $0.1$.}
    \label{fig:full_phase_transit_2}
\end{figure}

\begin{figure}[h!]
    \centering
    \includegraphics[width=.45\linewidth]{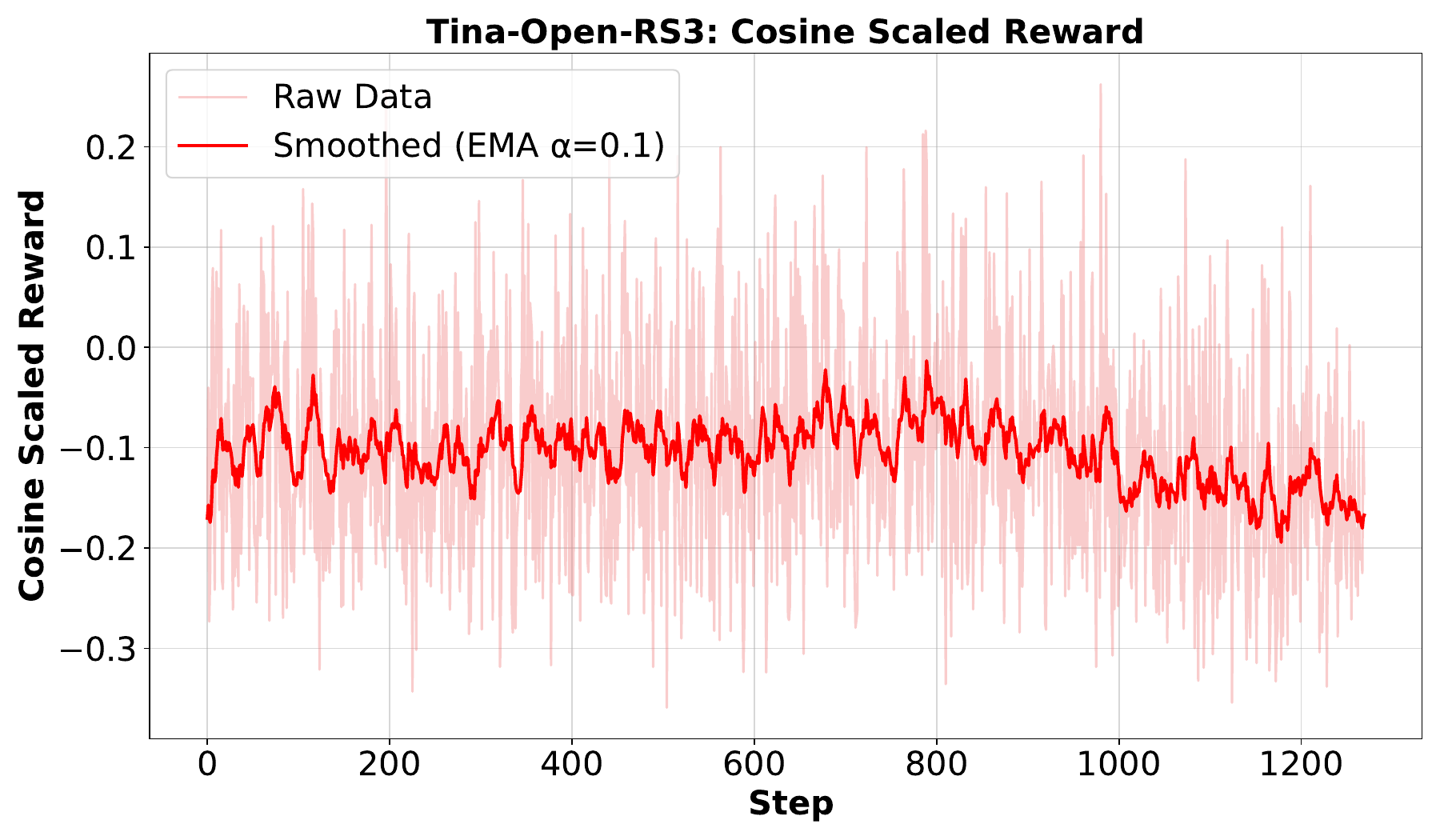}
    \includegraphics[width=.45\linewidth]{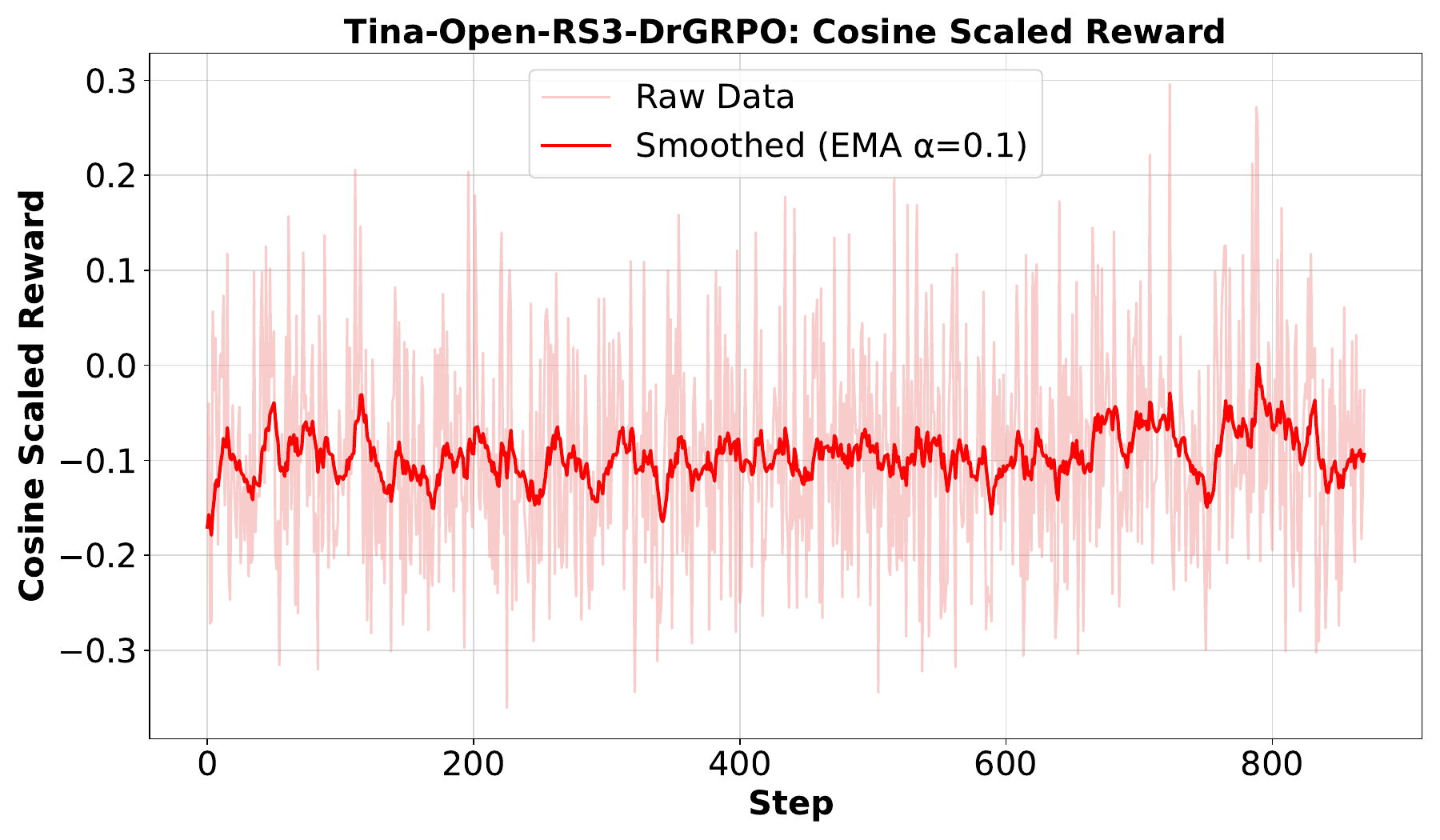}
    \includegraphics[width=.45\linewidth]{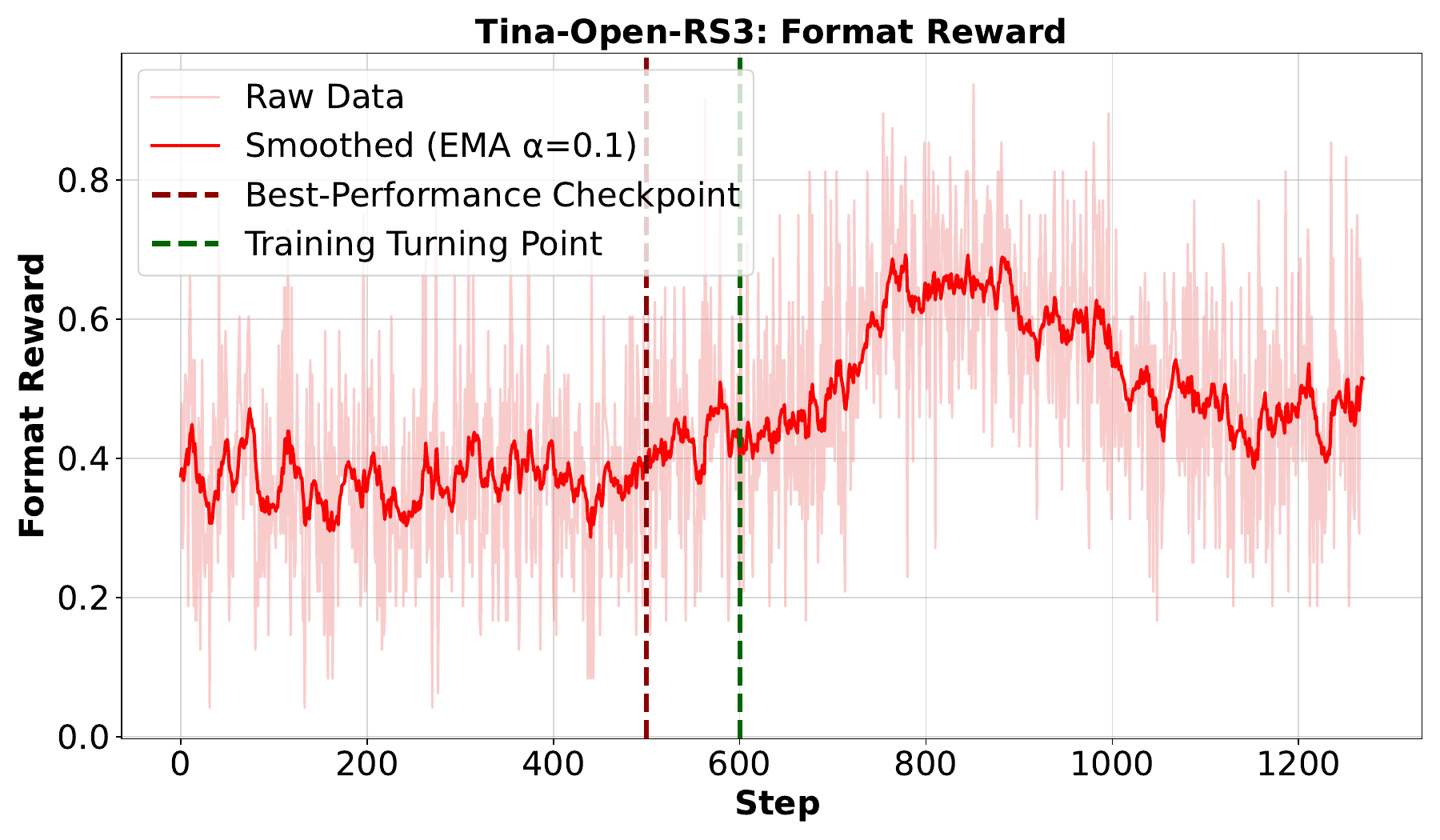}
    \includegraphics[width=.45\linewidth]{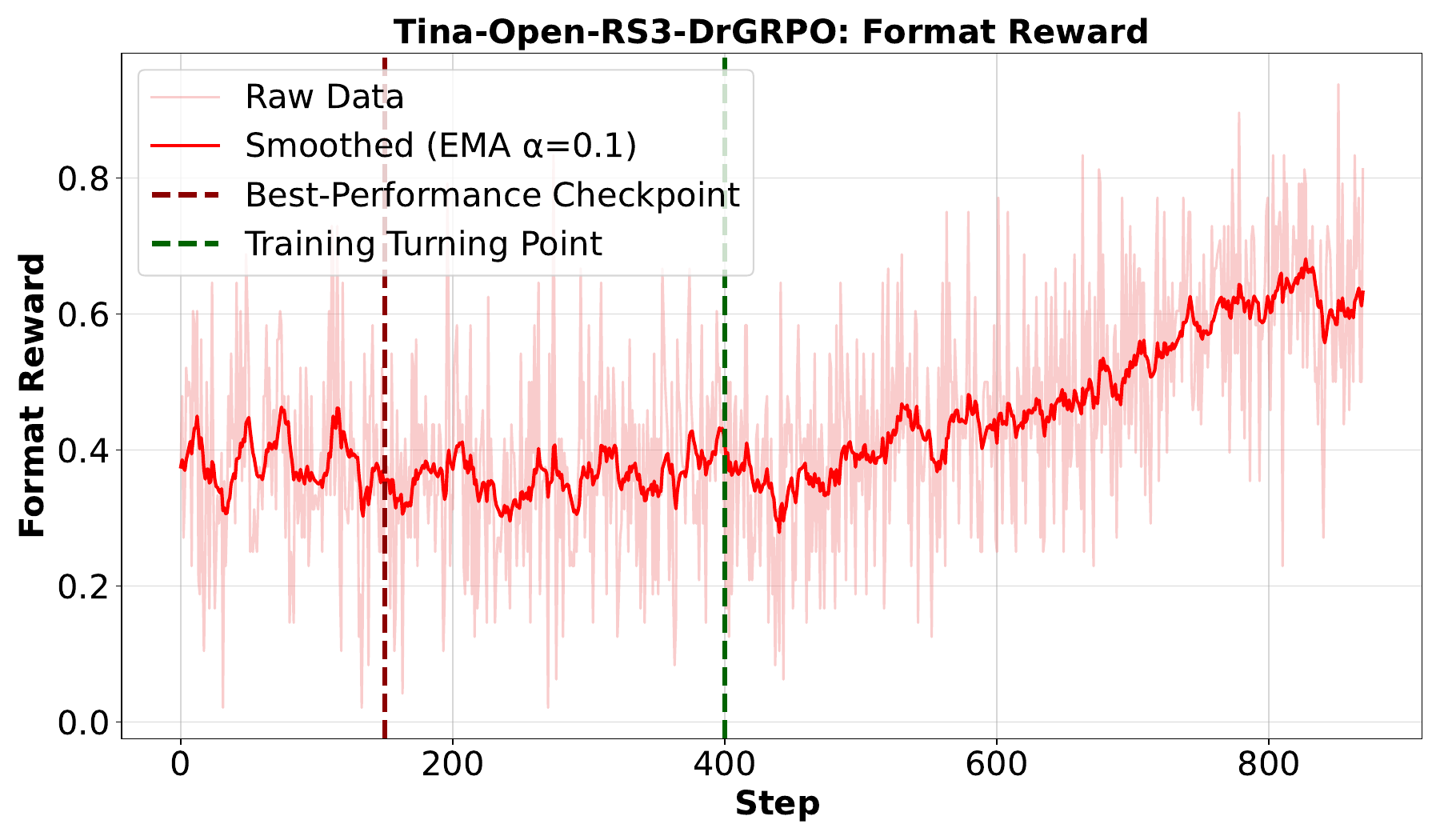}
    \includegraphics[width=.45\linewidth]{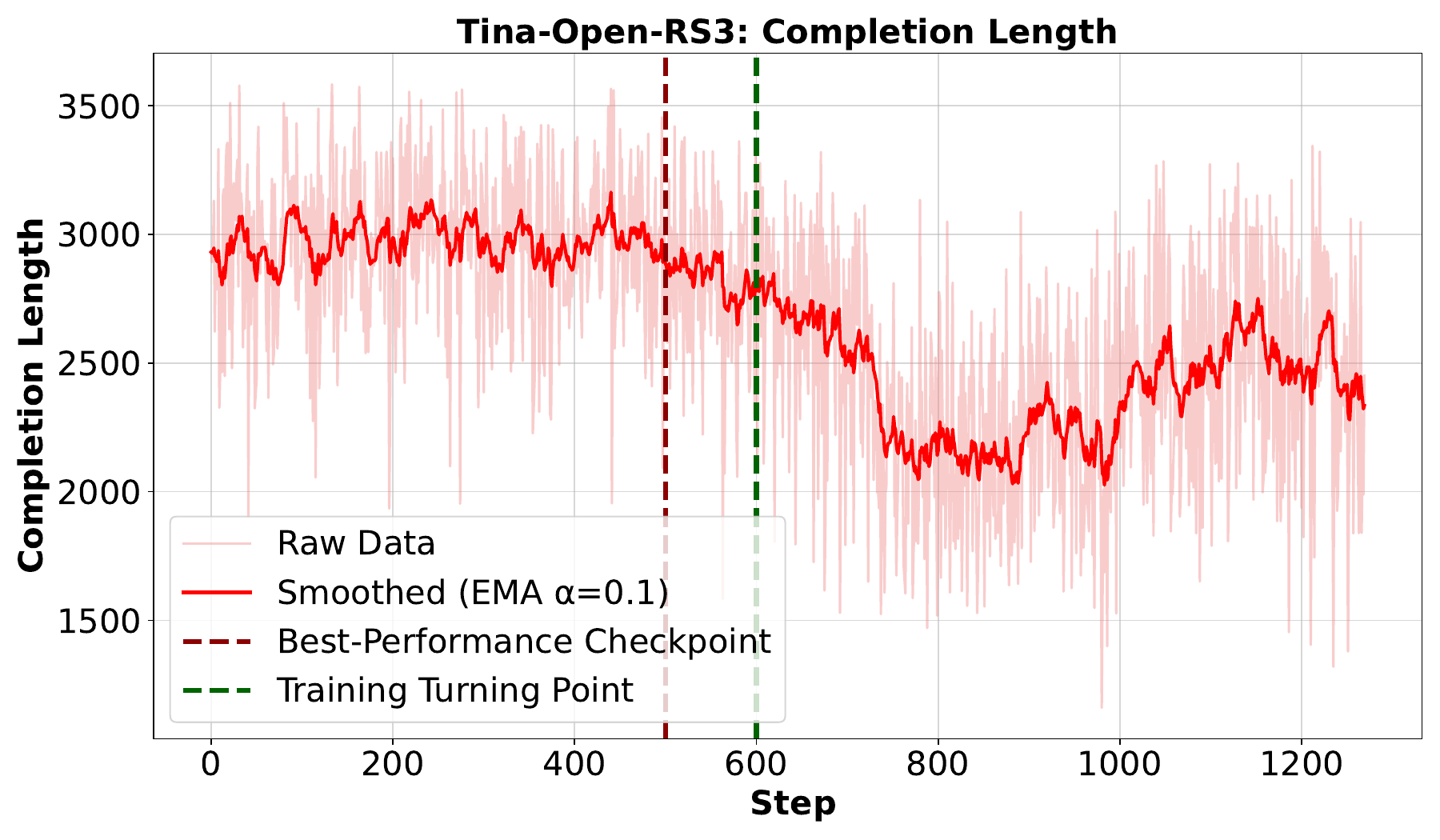}
    \includegraphics[width=.45\linewidth]{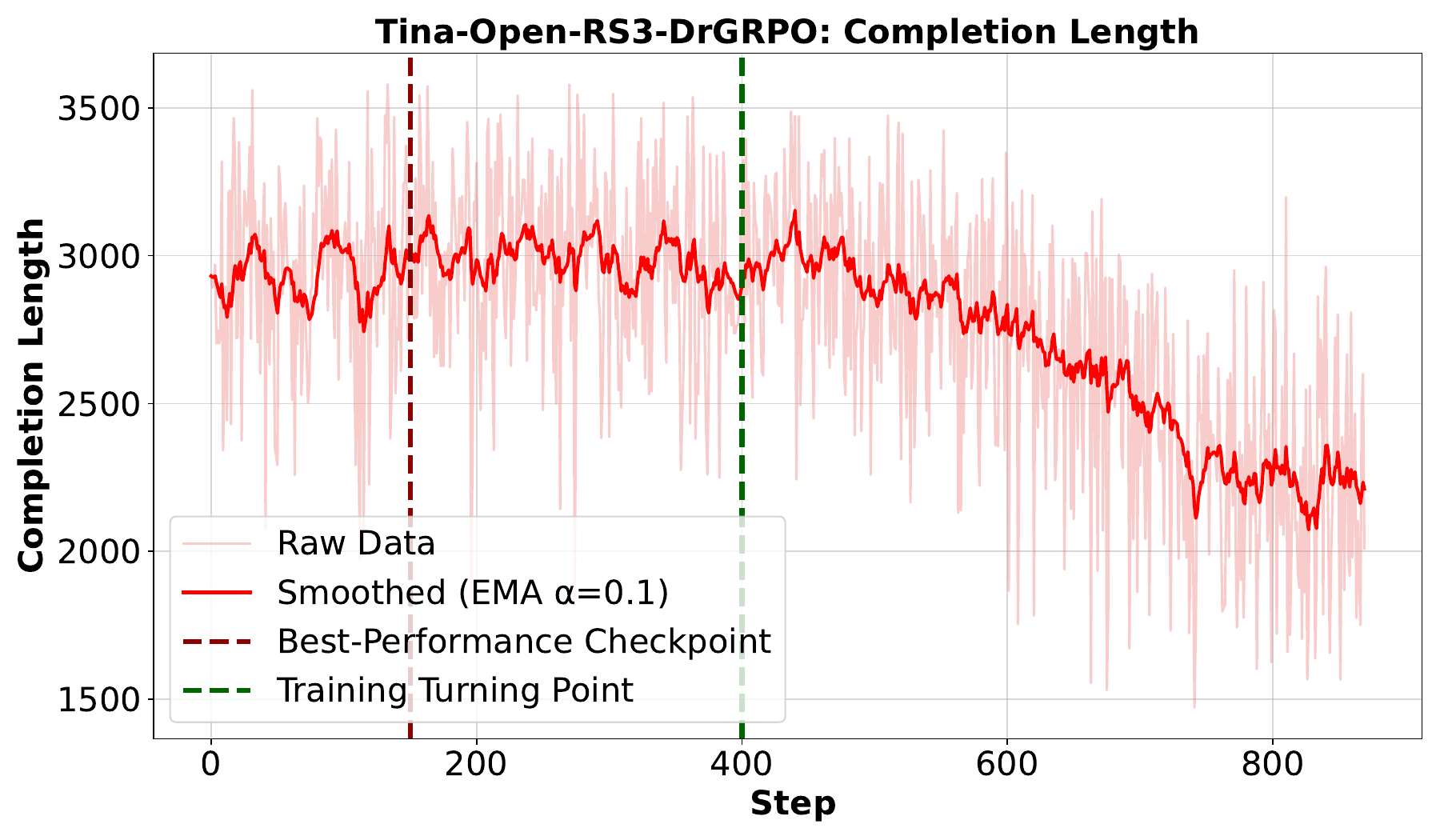}
    \caption{\textbf{Phase transition in \texttt{Tina-Open-RS3} and \texttt{Tina-Open-RS3-GRPO}.} The raw data is from the Weights \& Biases training logs and smoothed via exponential moving average (EMA) with factor $0.1$.}
    \label{fig:full_phase_transit_3}
\end{figure}

\begin{figure}[h!]
    \centering
    \includegraphics[width=.45\linewidth]{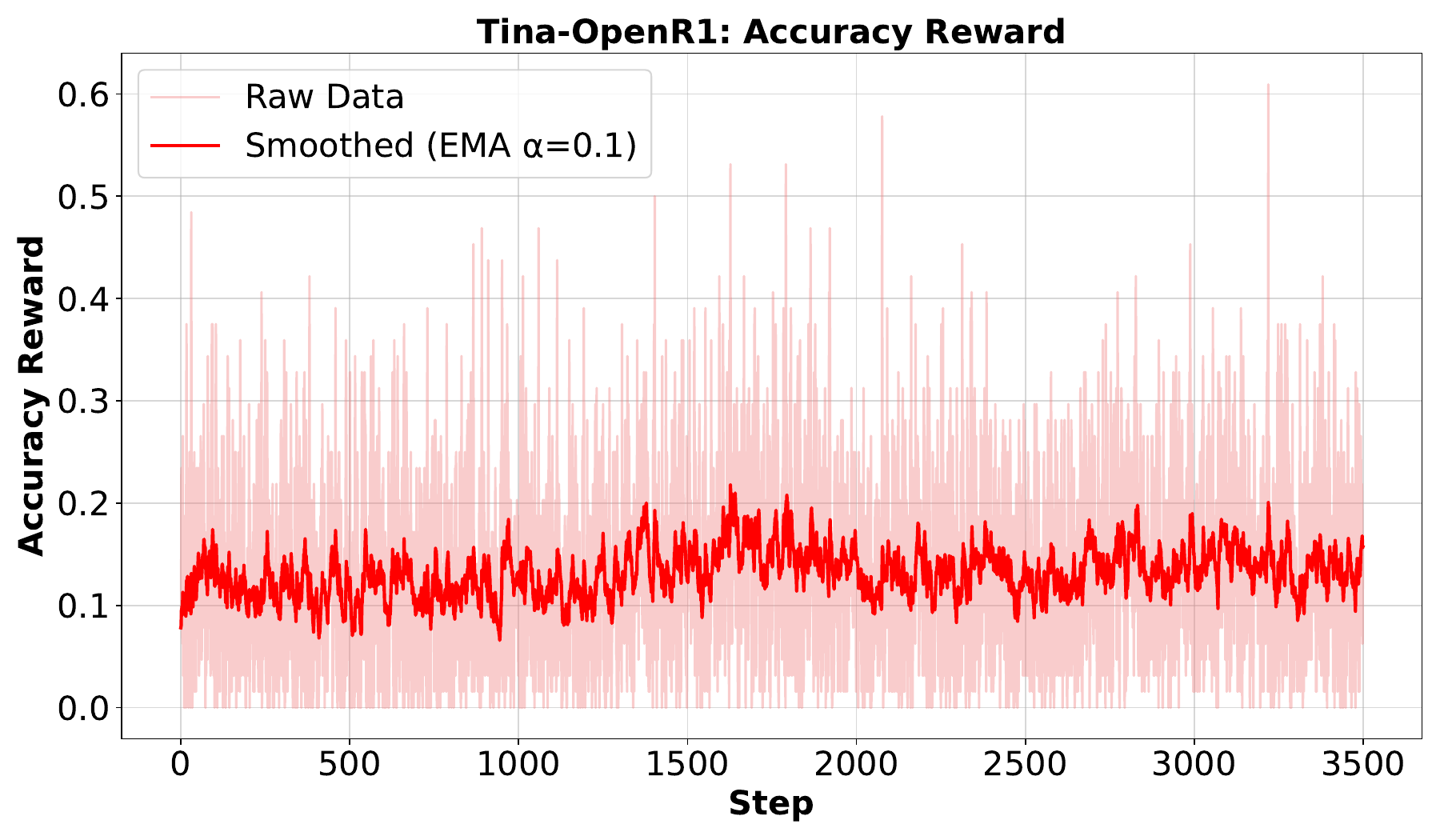}
    \includegraphics[width=.45\linewidth]{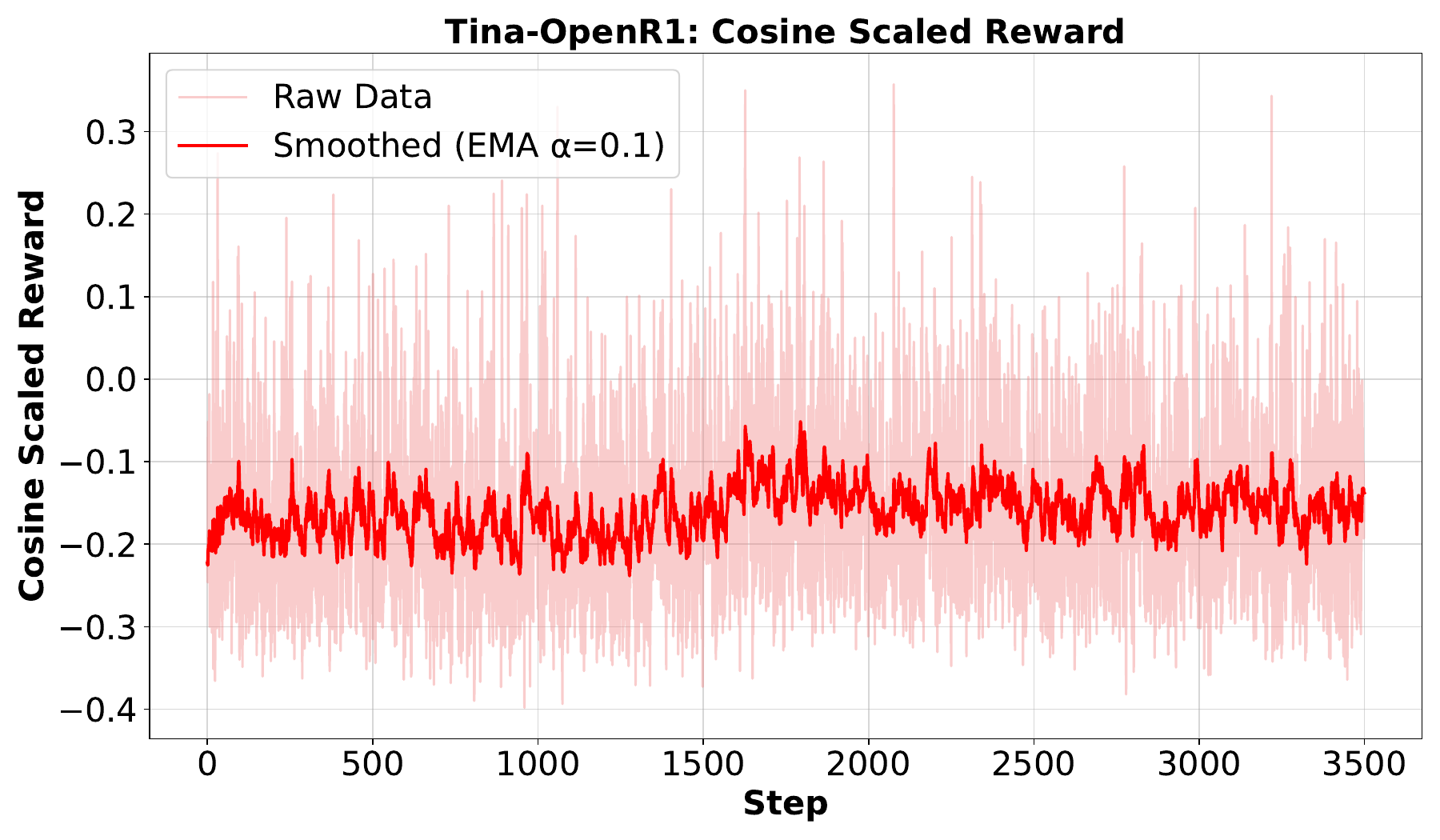}
    \includegraphics[width=.45\linewidth]{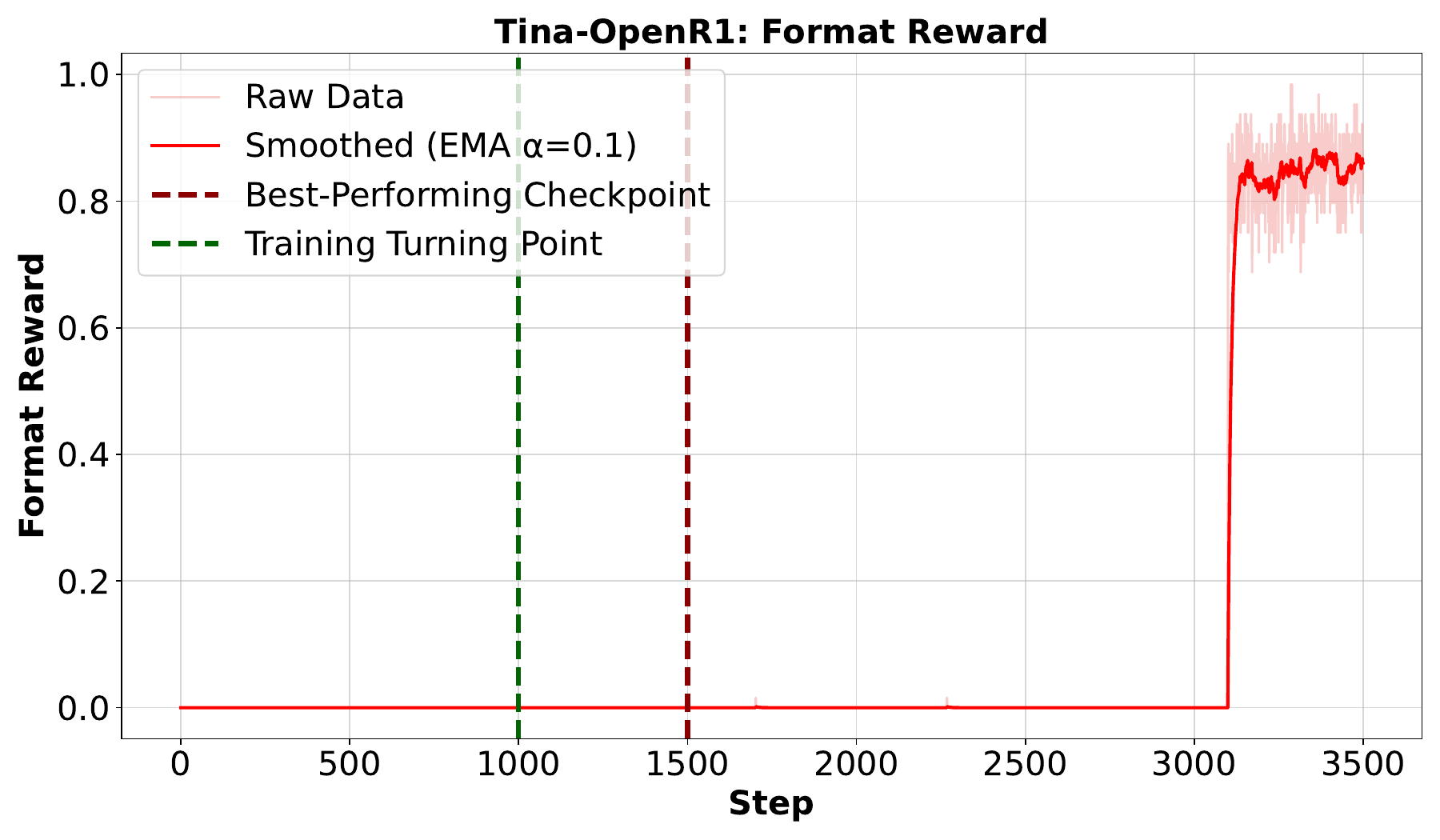}
    \includegraphics[width=.45\linewidth]{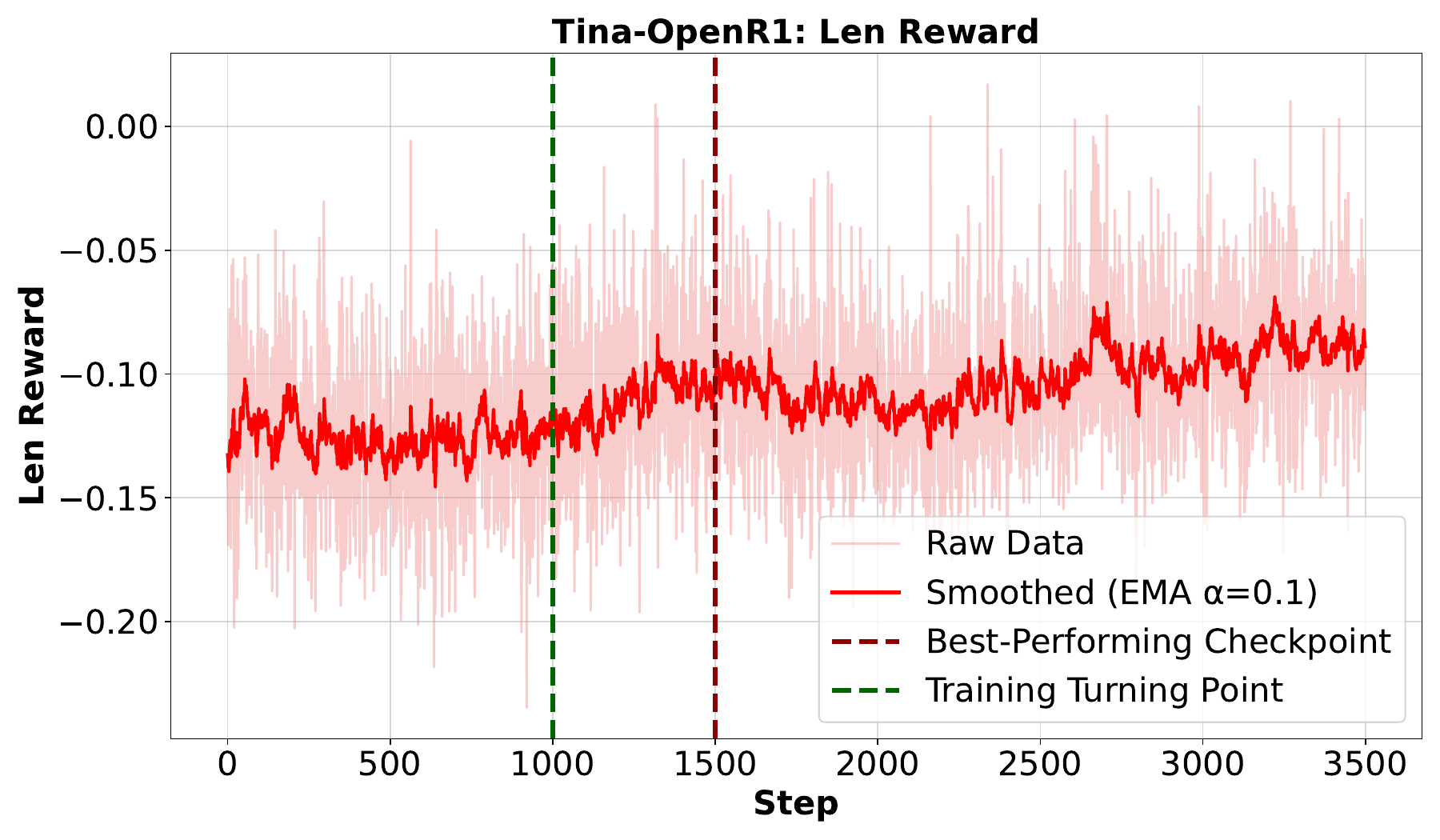}
    \includegraphics[width=.45\linewidth]{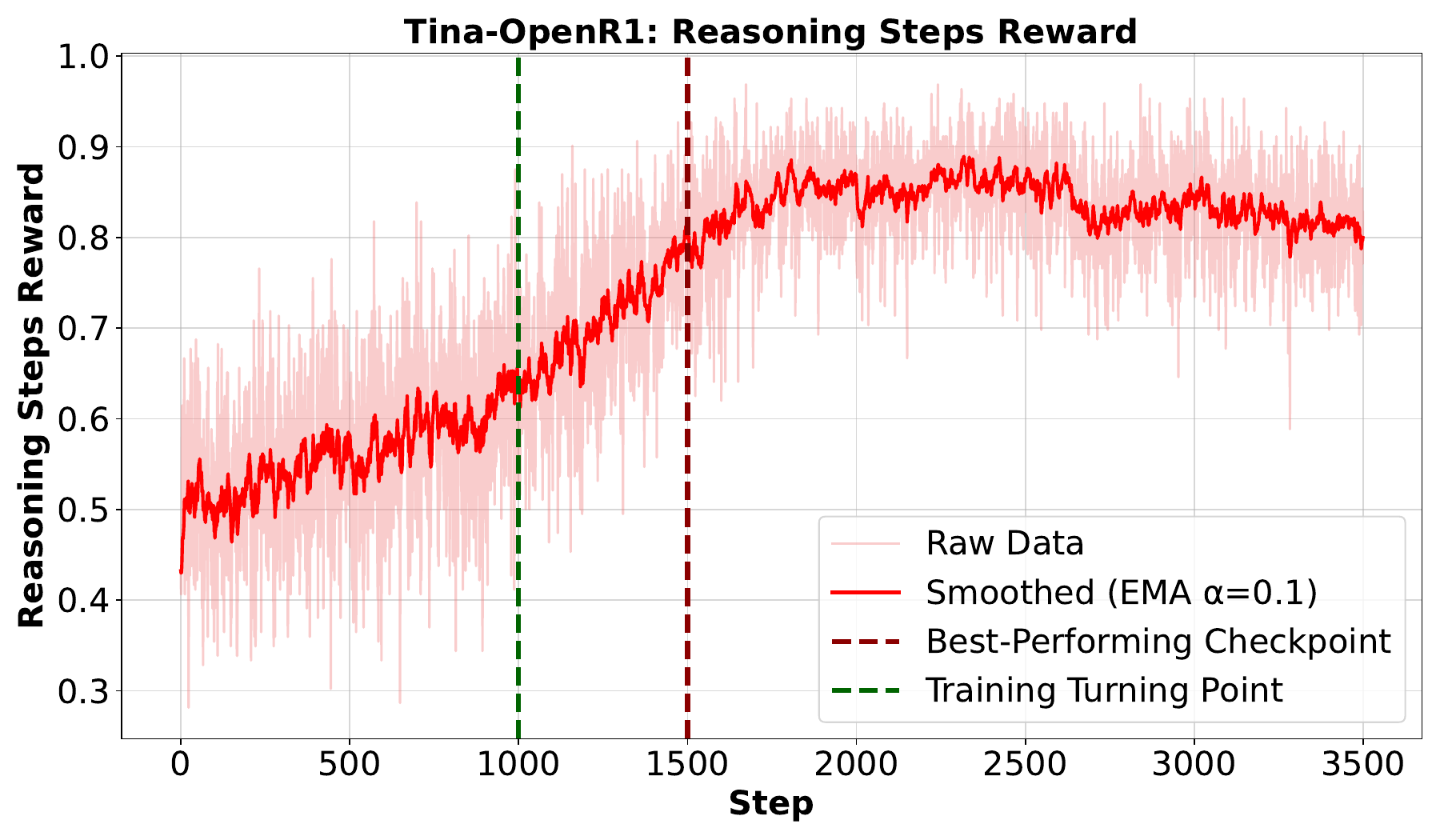}
    \includegraphics[width=.45\linewidth]{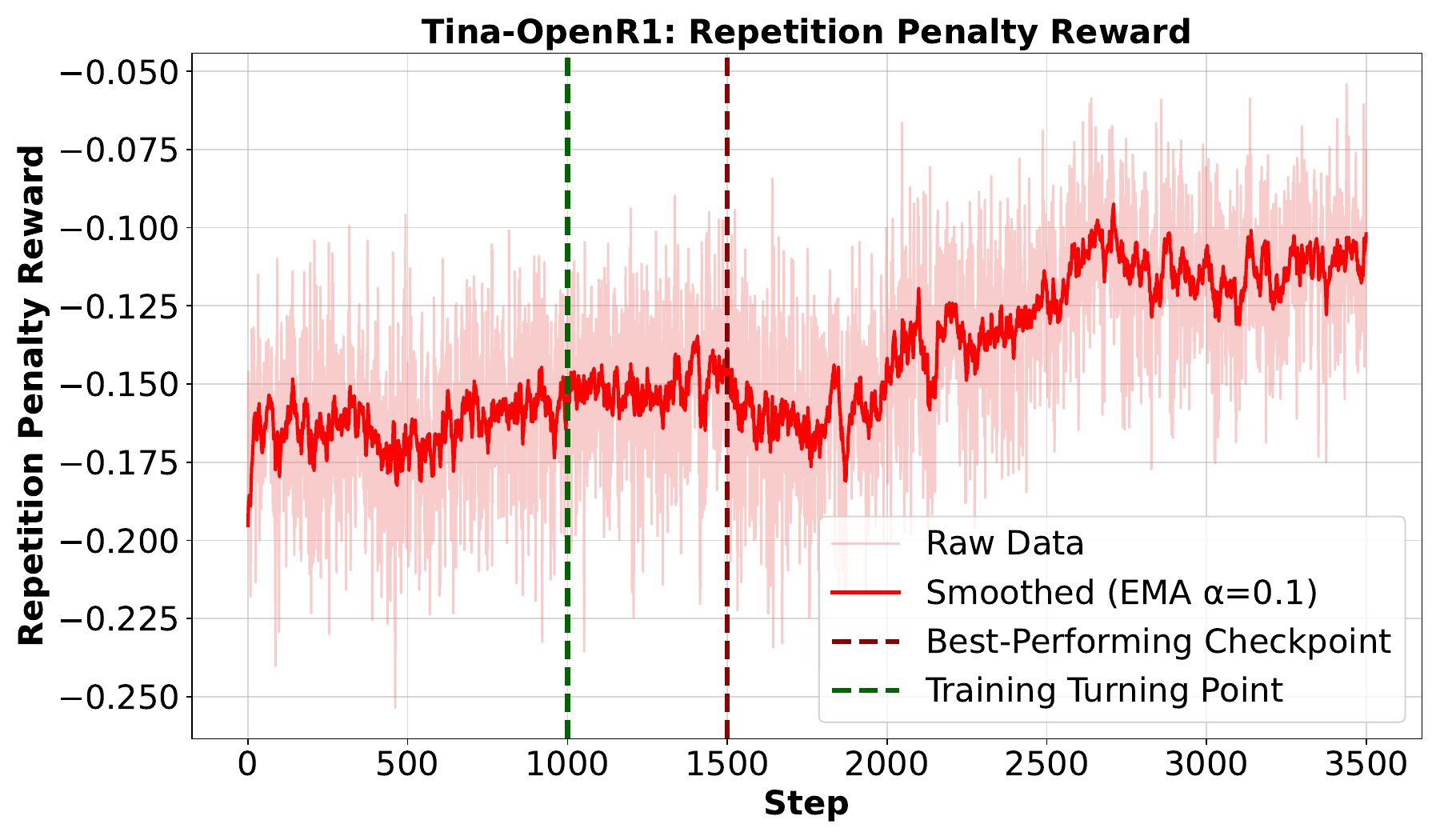}
    \includegraphics[width=.45\linewidth]{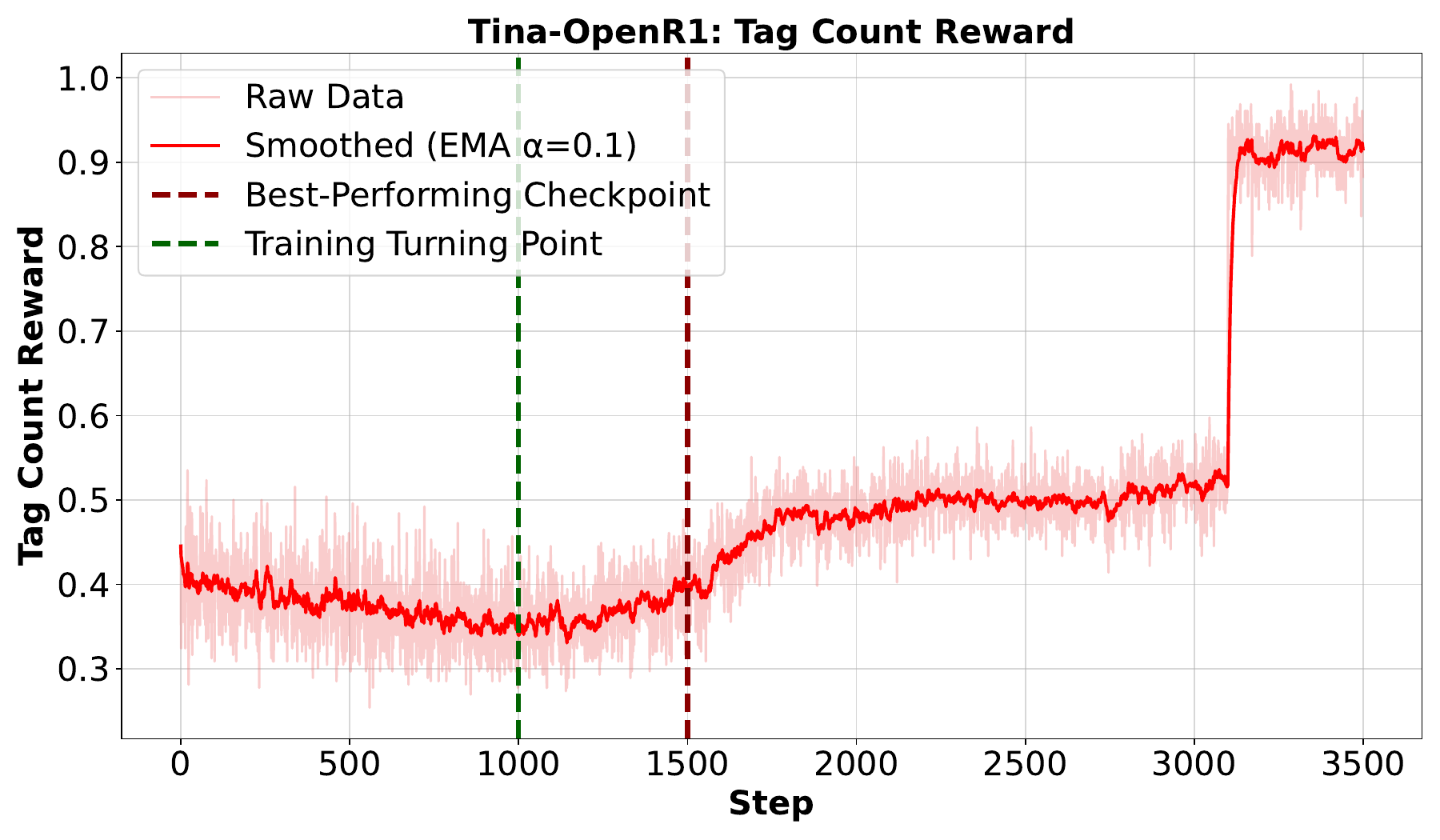}
    \includegraphics[width=.45\linewidth]{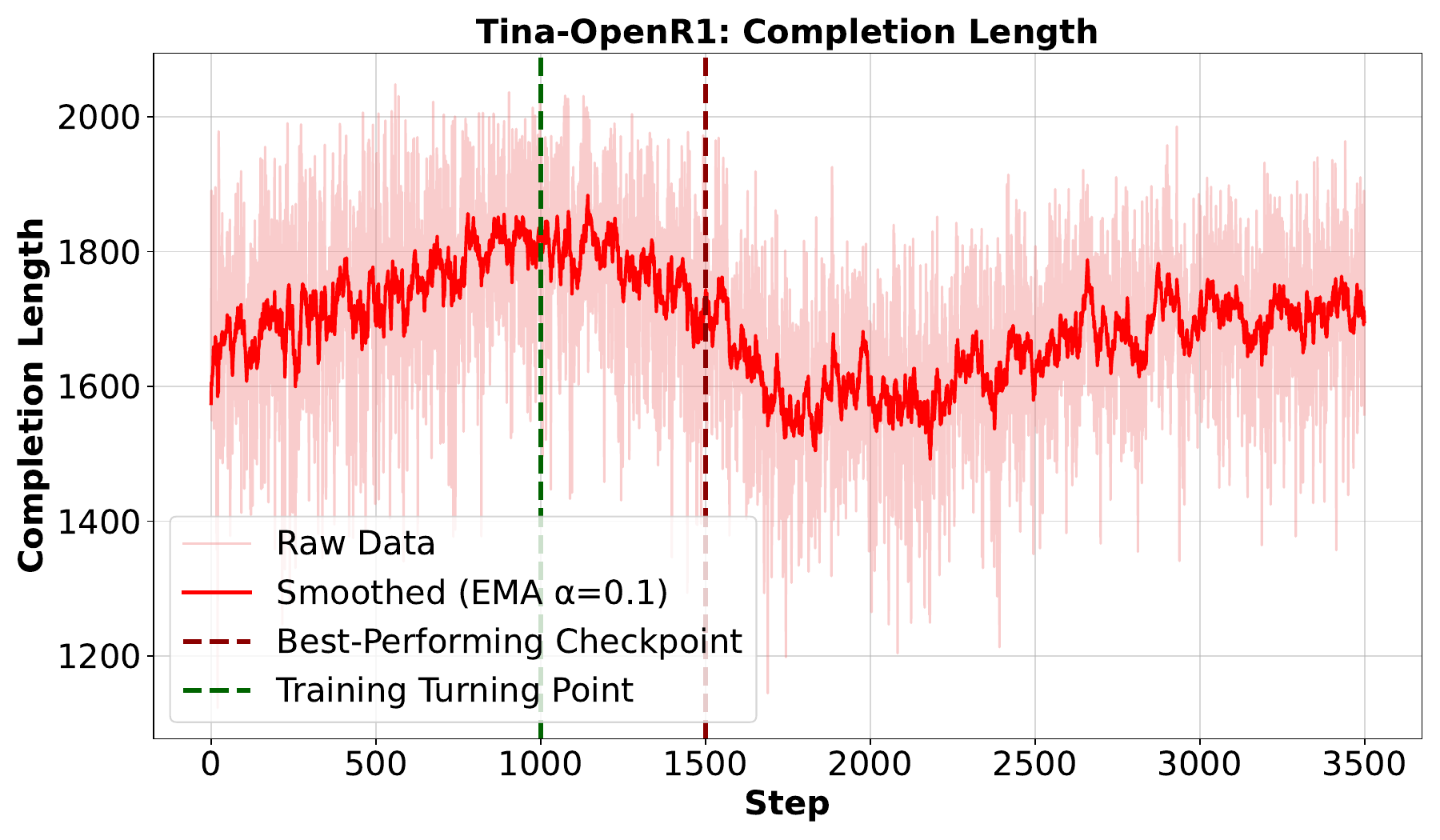}
    \caption{\textbf{Phase transition in \texttt{Tina-OpenR1}.} The raw data is from the Weights \& Biases training logs and smoothed via exponential moving average (EMA) with factor $0.1$.}
    \label{fig:full_phase_transit_4}
\end{figure}

\begin{figure}[h!]
    \centering

    \includegraphics[width=.45\linewidth]{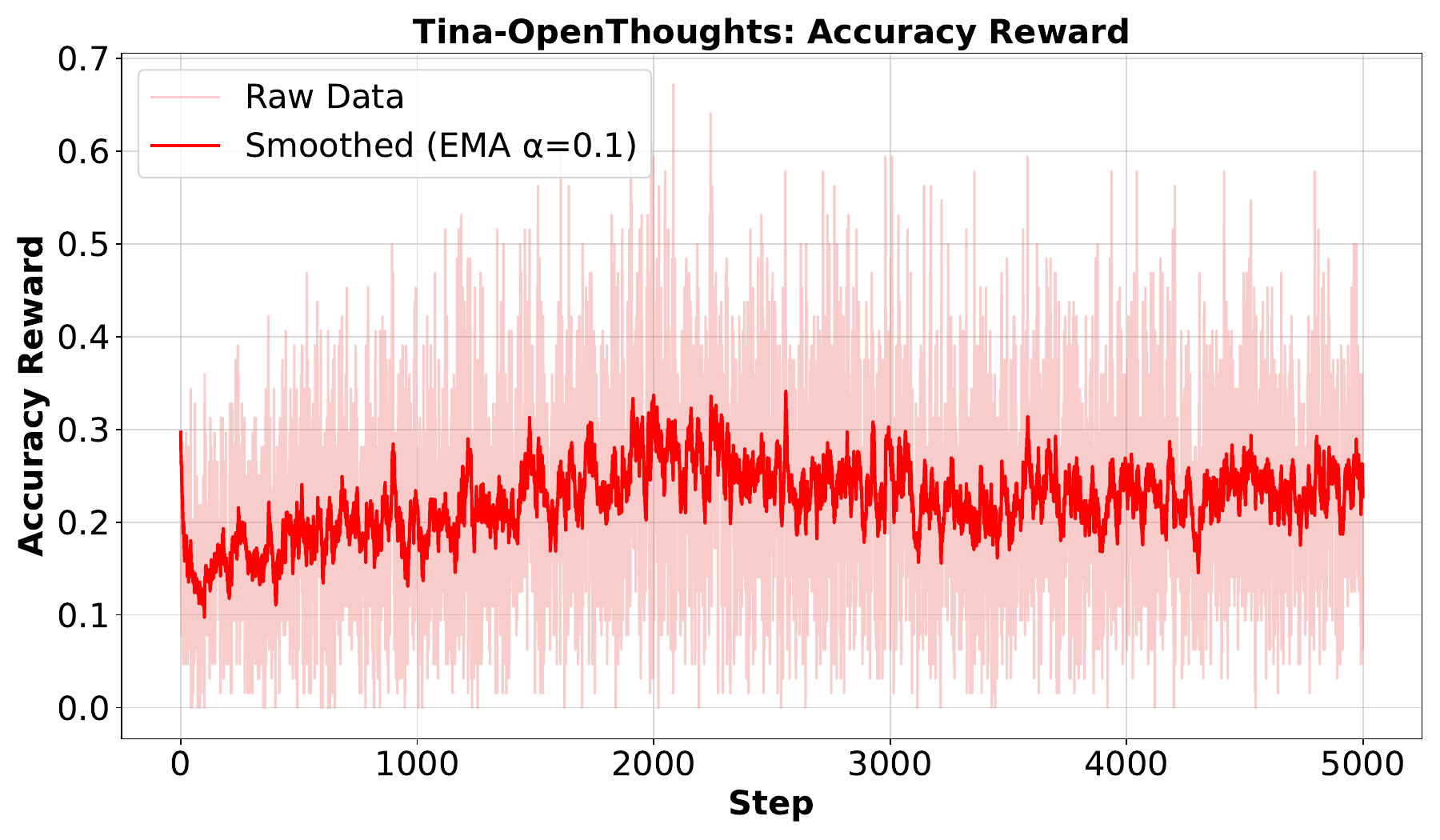}
    \includegraphics[width=.45\linewidth]{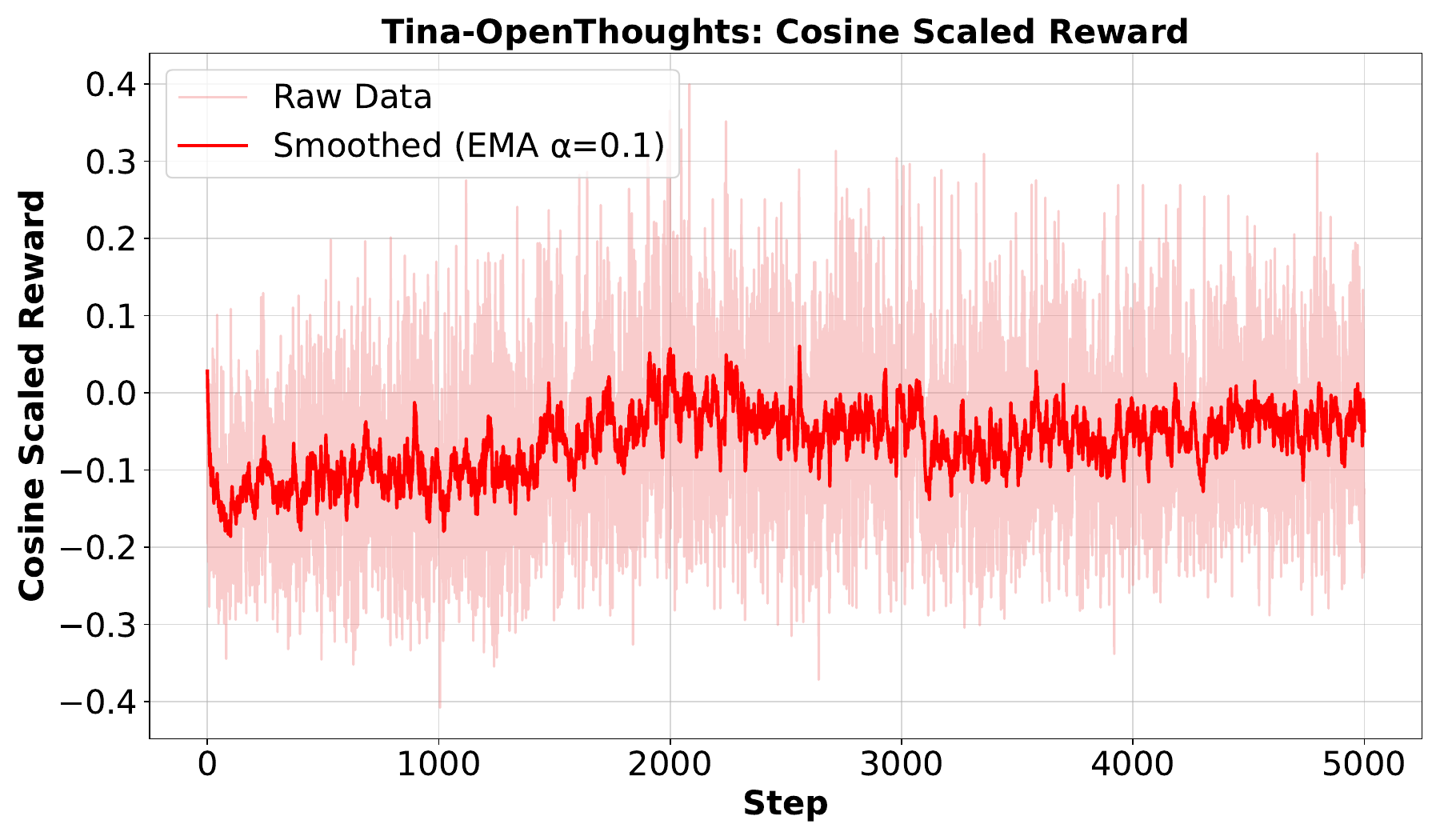}
    \includegraphics[width=.45\linewidth]{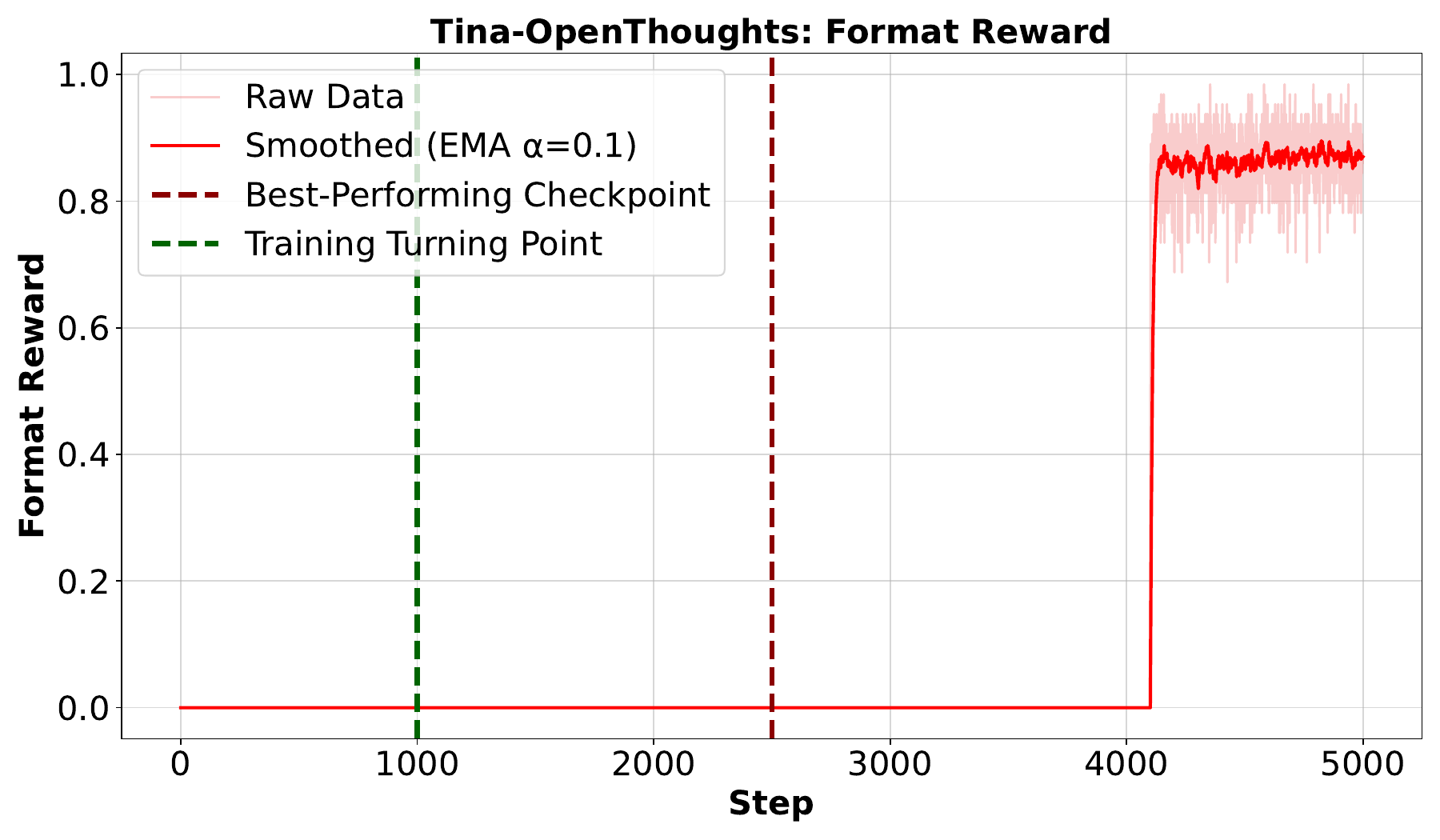}
    \includegraphics[width=.45\linewidth]{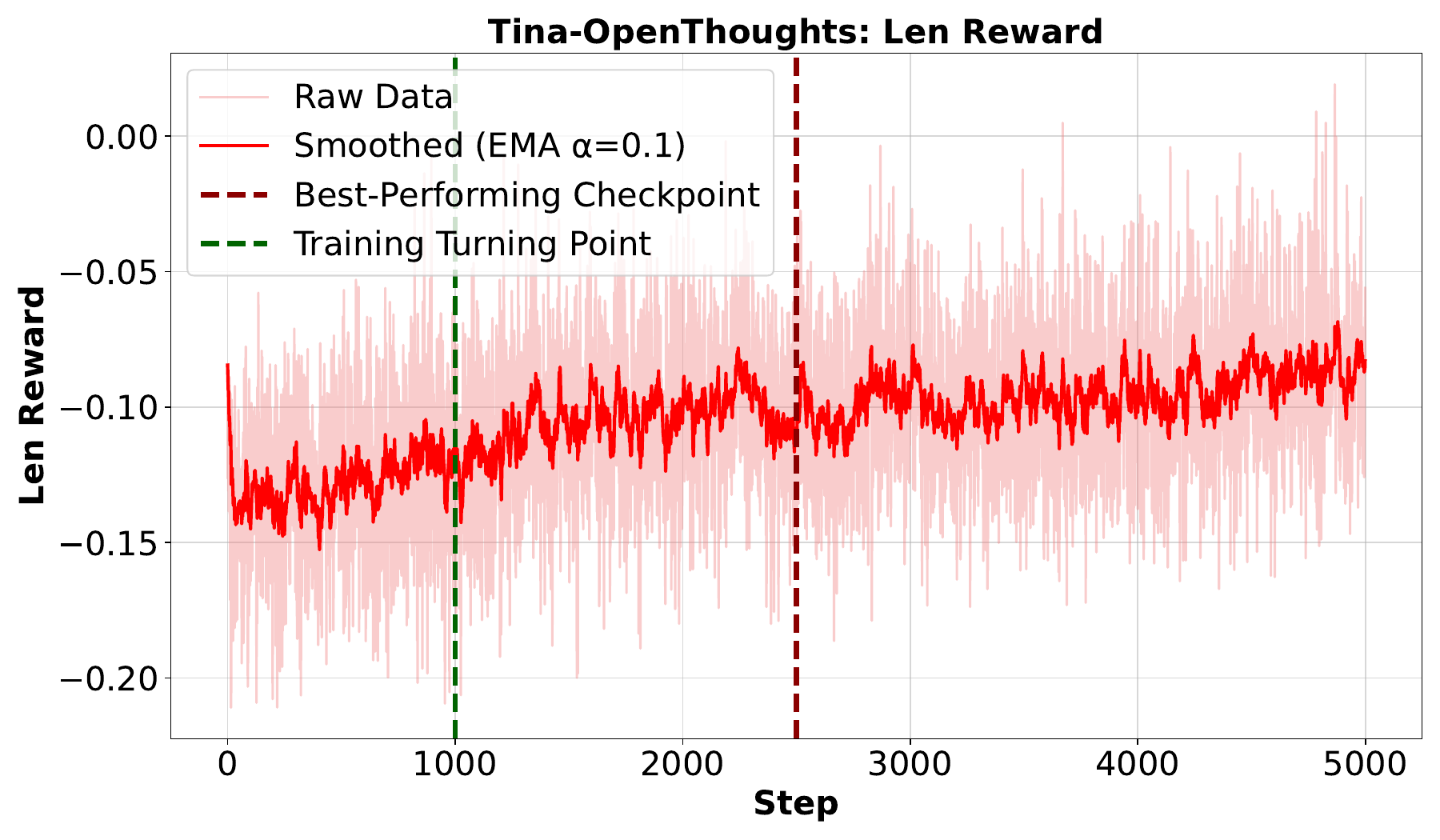}
    \includegraphics[width=.45\linewidth]{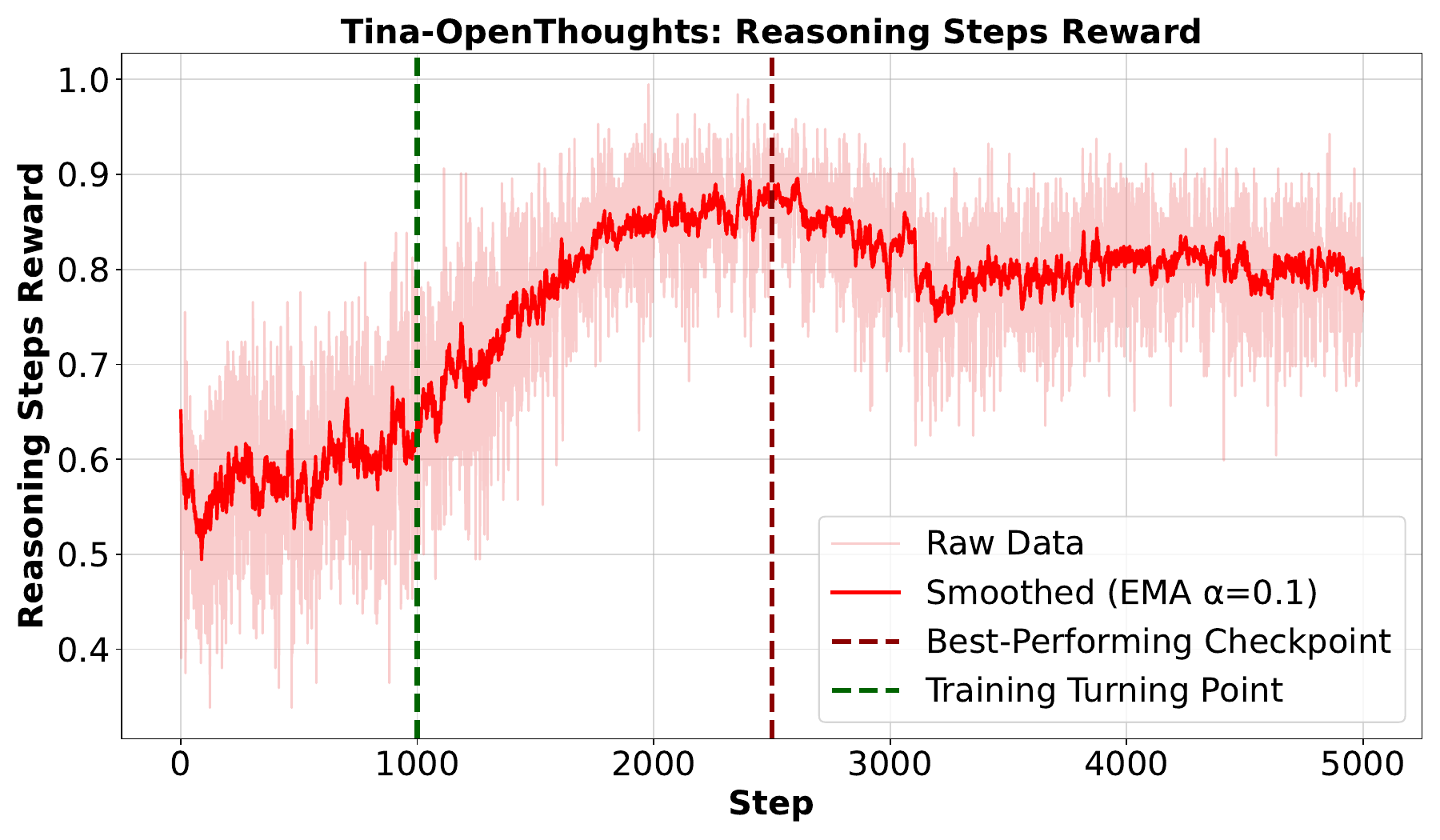}
    \includegraphics[width=.45\linewidth]{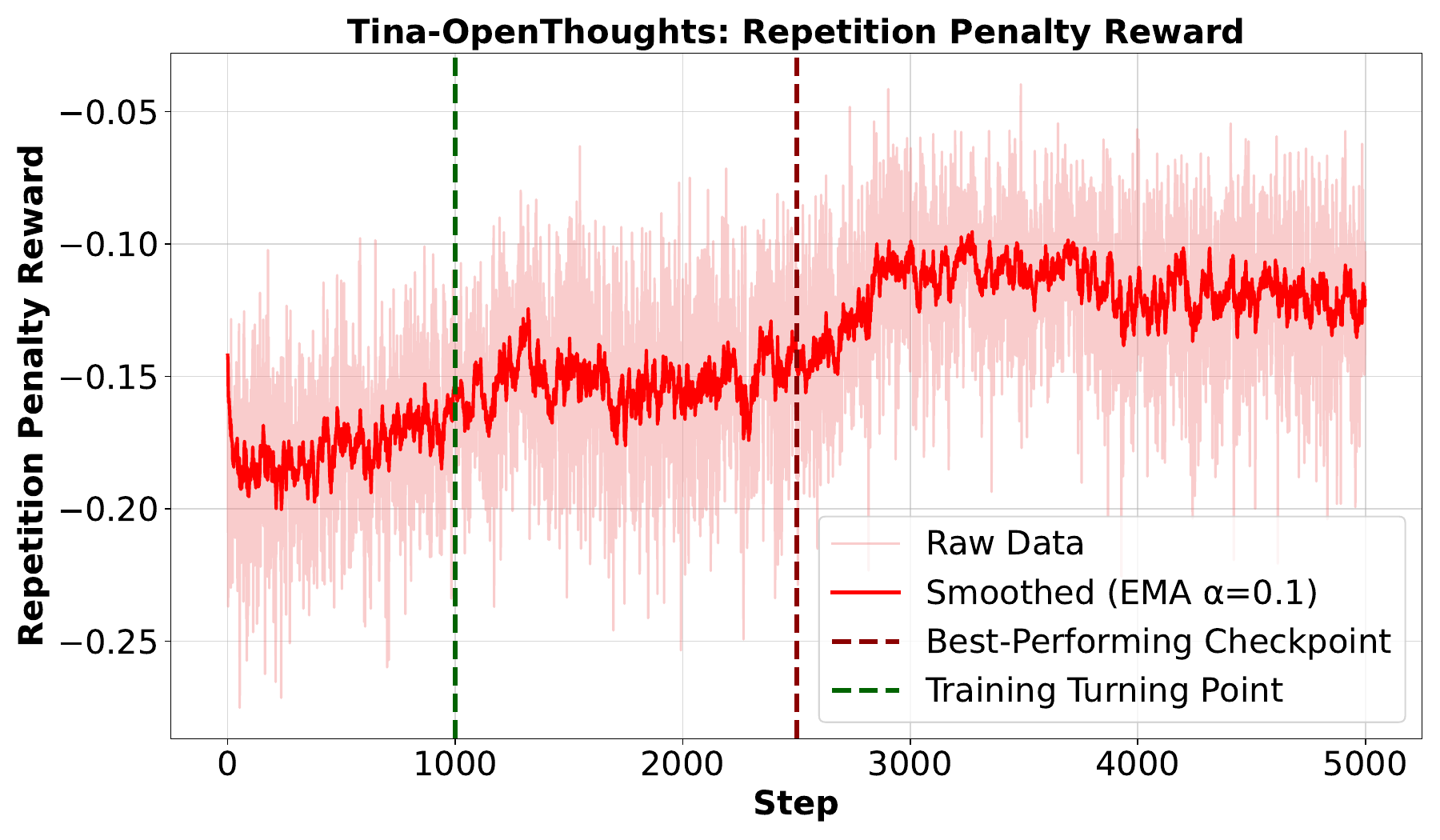}
    \includegraphics[width=.45\linewidth]{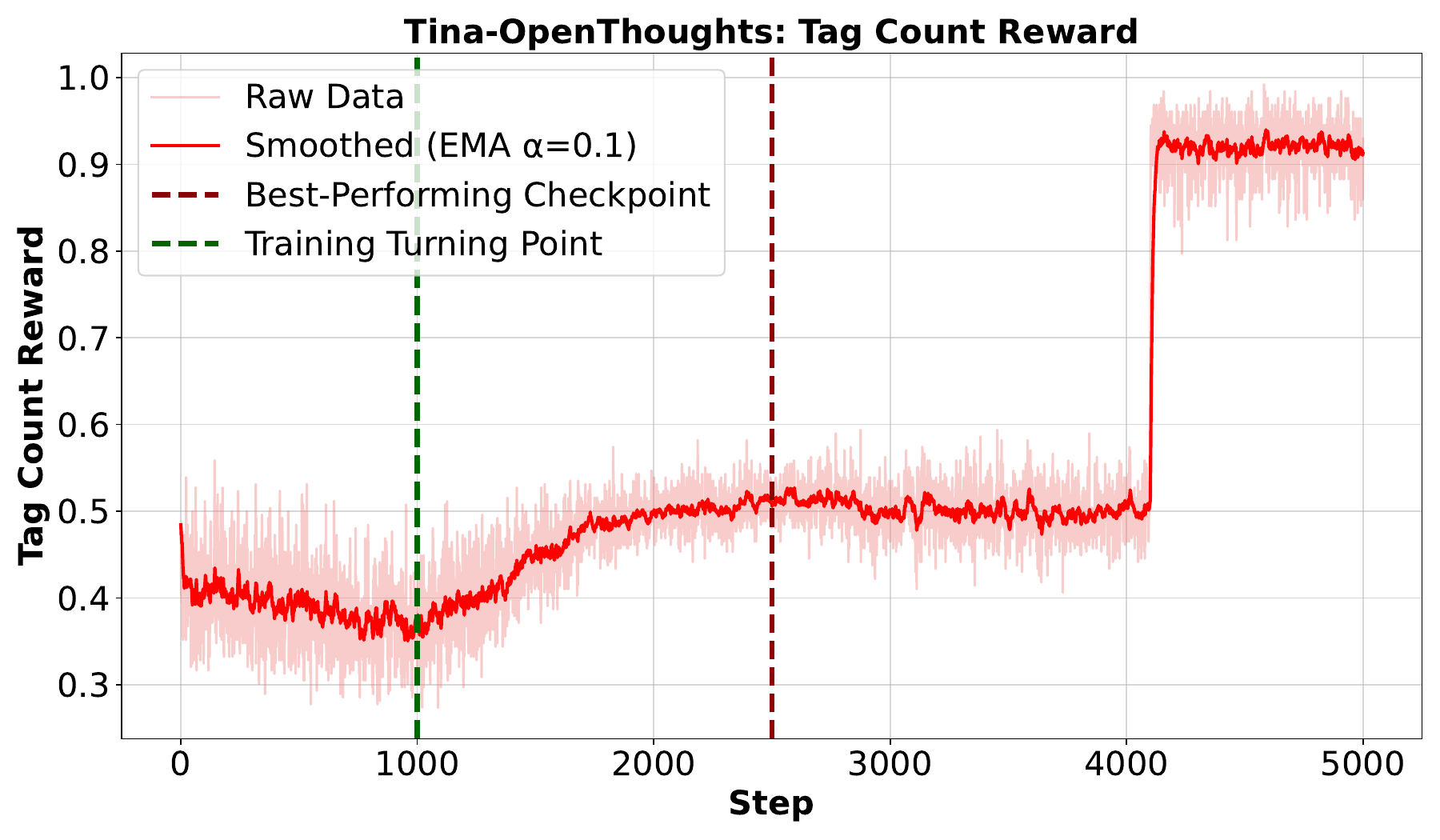}
    \includegraphics[width=.45\linewidth]{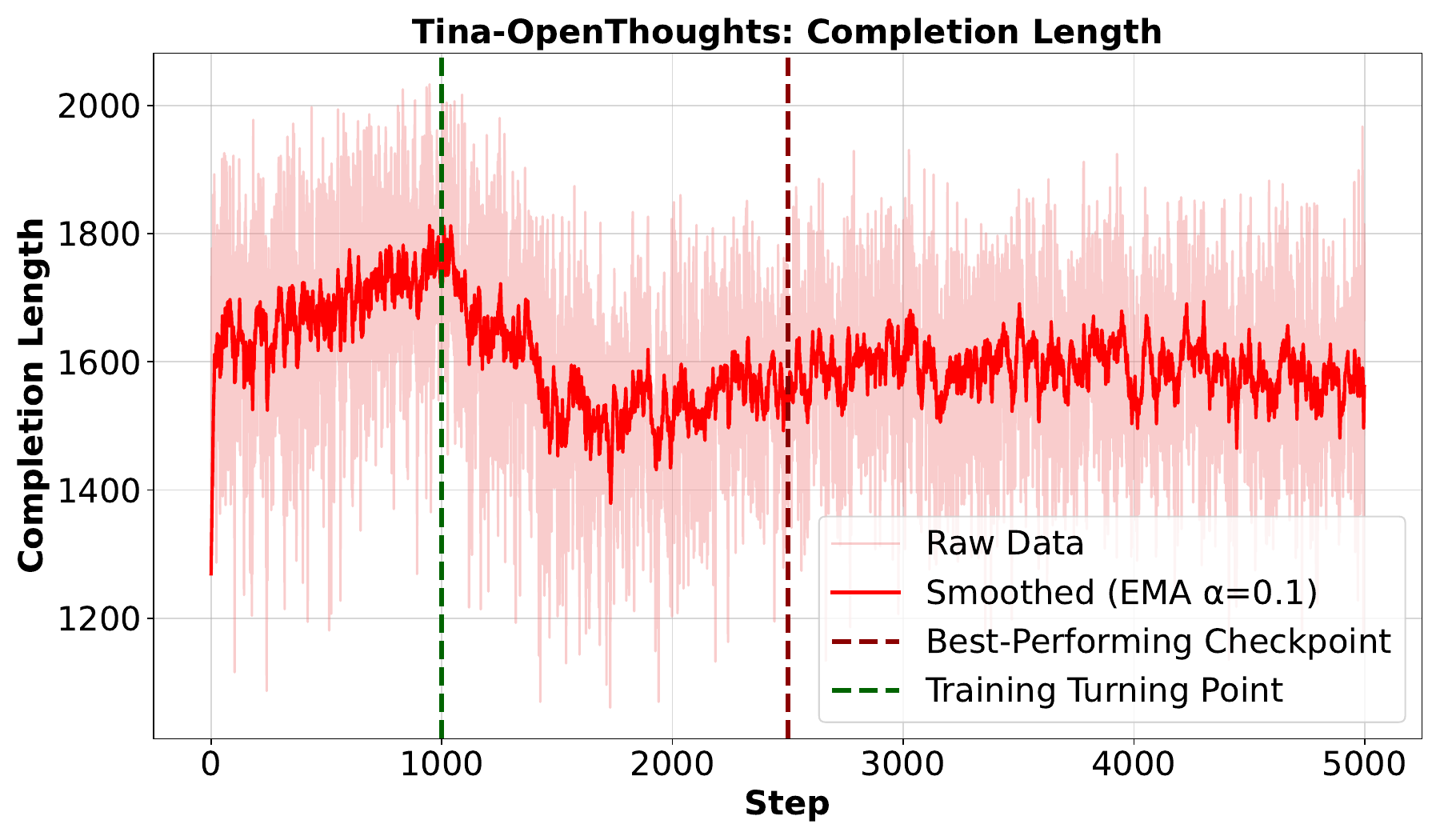}
    \caption{\textbf{Phase transition in \texttt{Tina-OpenThoughts}.} The raw data is from the Weights \& Biases training logs and smoothed via exponential moving average (EMA) with factor $0.1$.}
    \label{fig:full_phase_transit_5}
\end{figure}

\begin{figure}[h!]
    \centering
    \includegraphics[width=.32\linewidth]{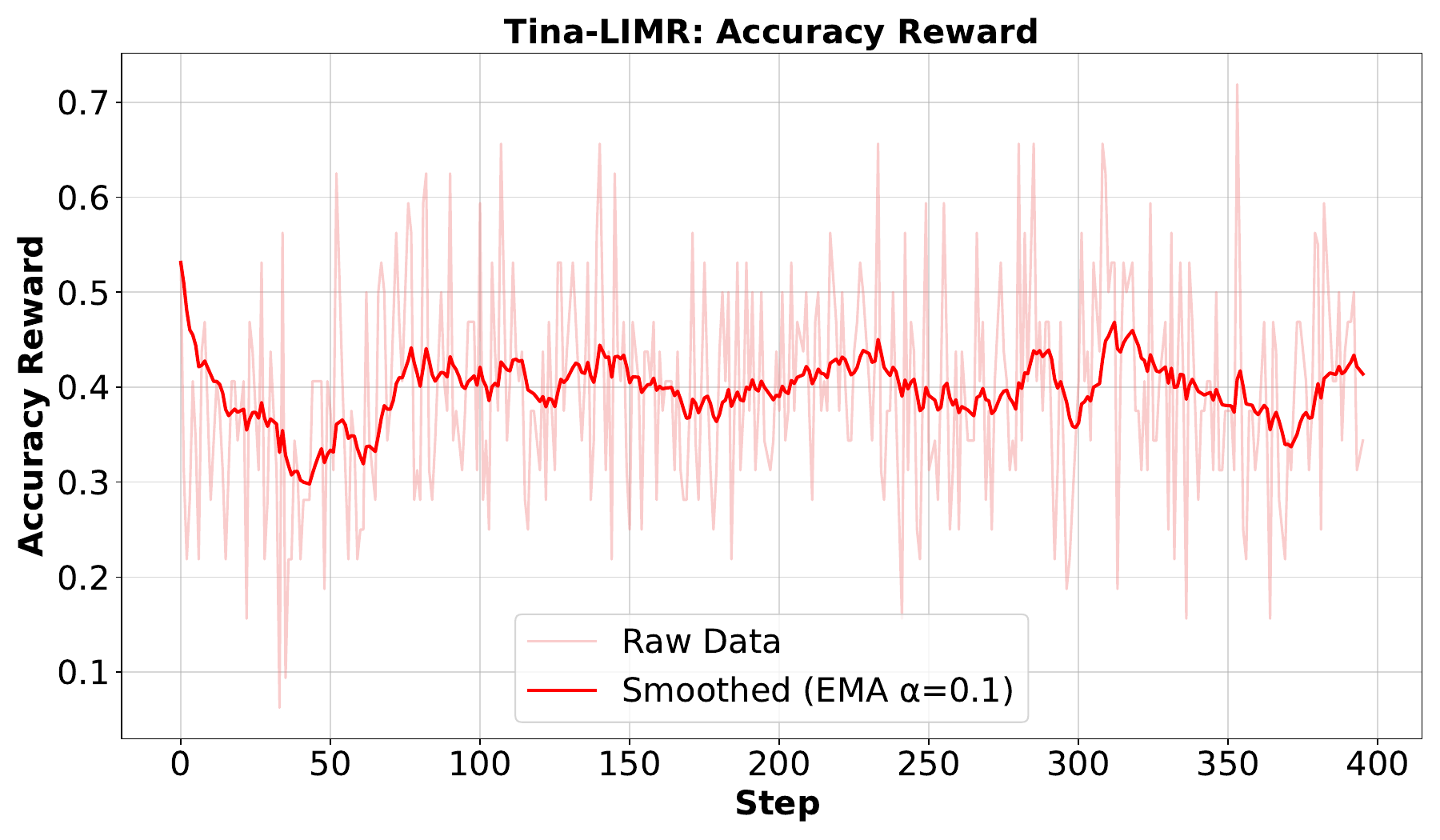}
    \includegraphics[width=.32\linewidth]{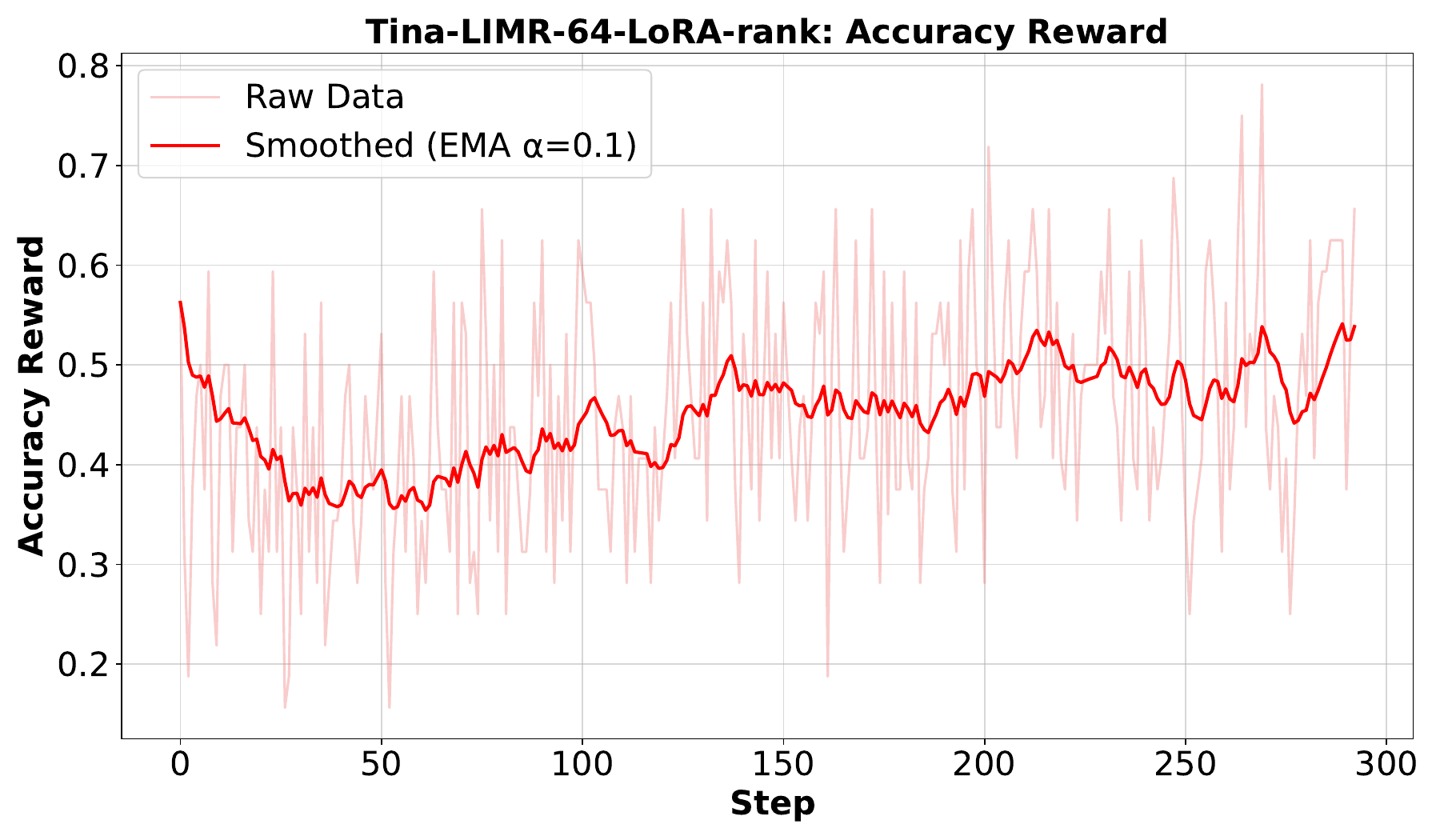}
    \includegraphics[width=.32\linewidth]{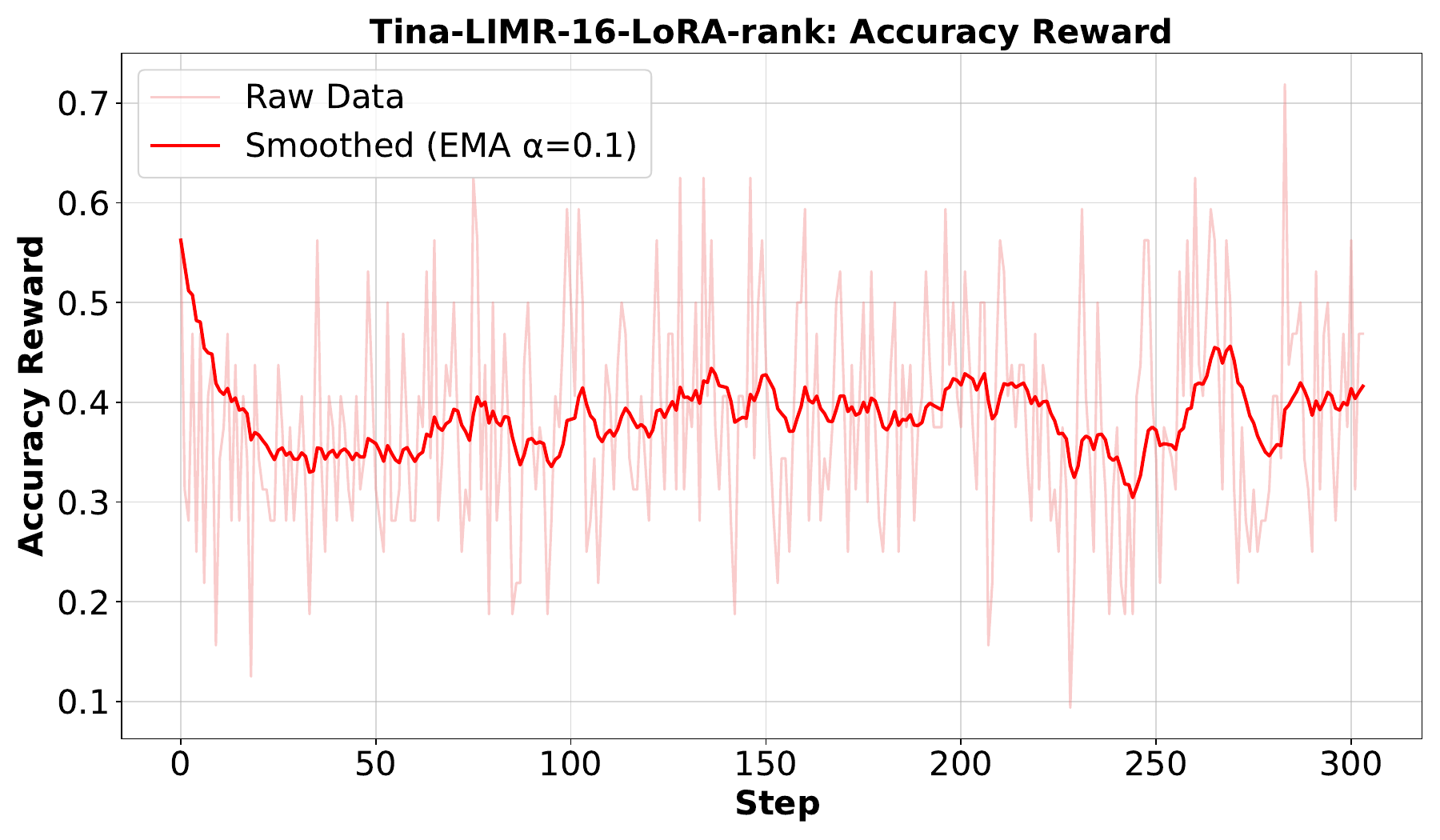}
    \includegraphics[width=.32\linewidth]{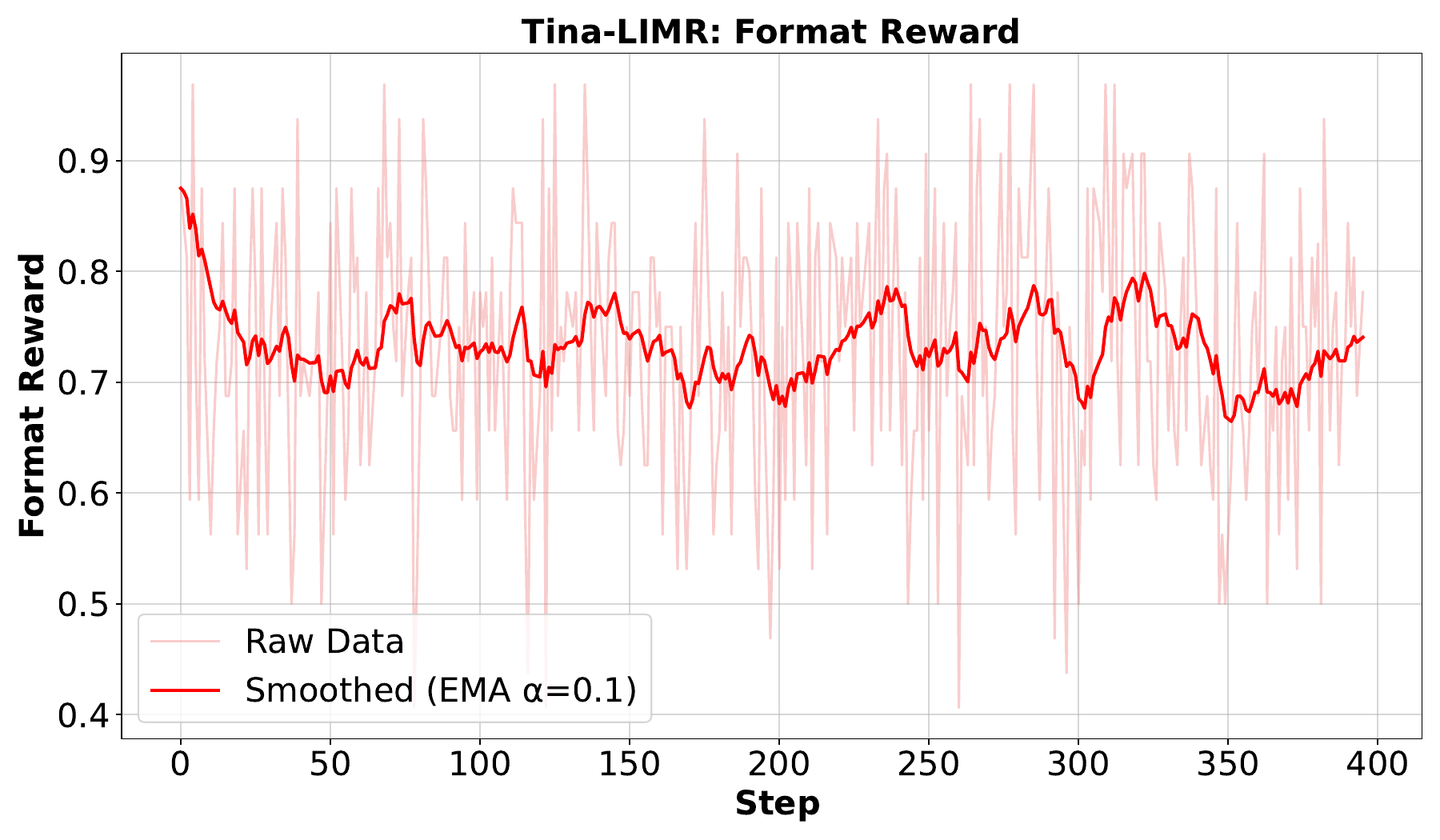}
    \includegraphics[width=.32\linewidth]{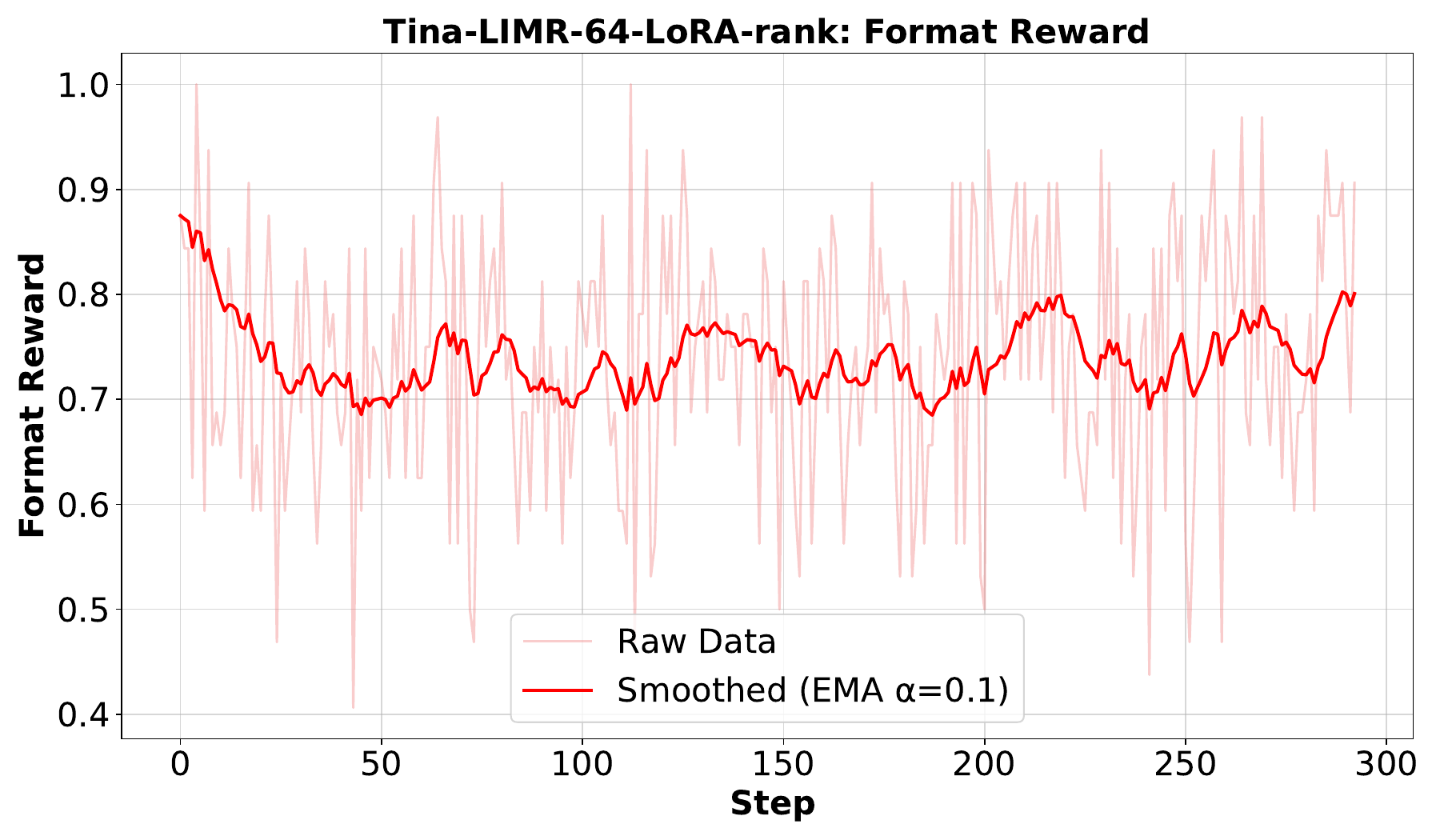}
    \includegraphics[width=.32\linewidth]{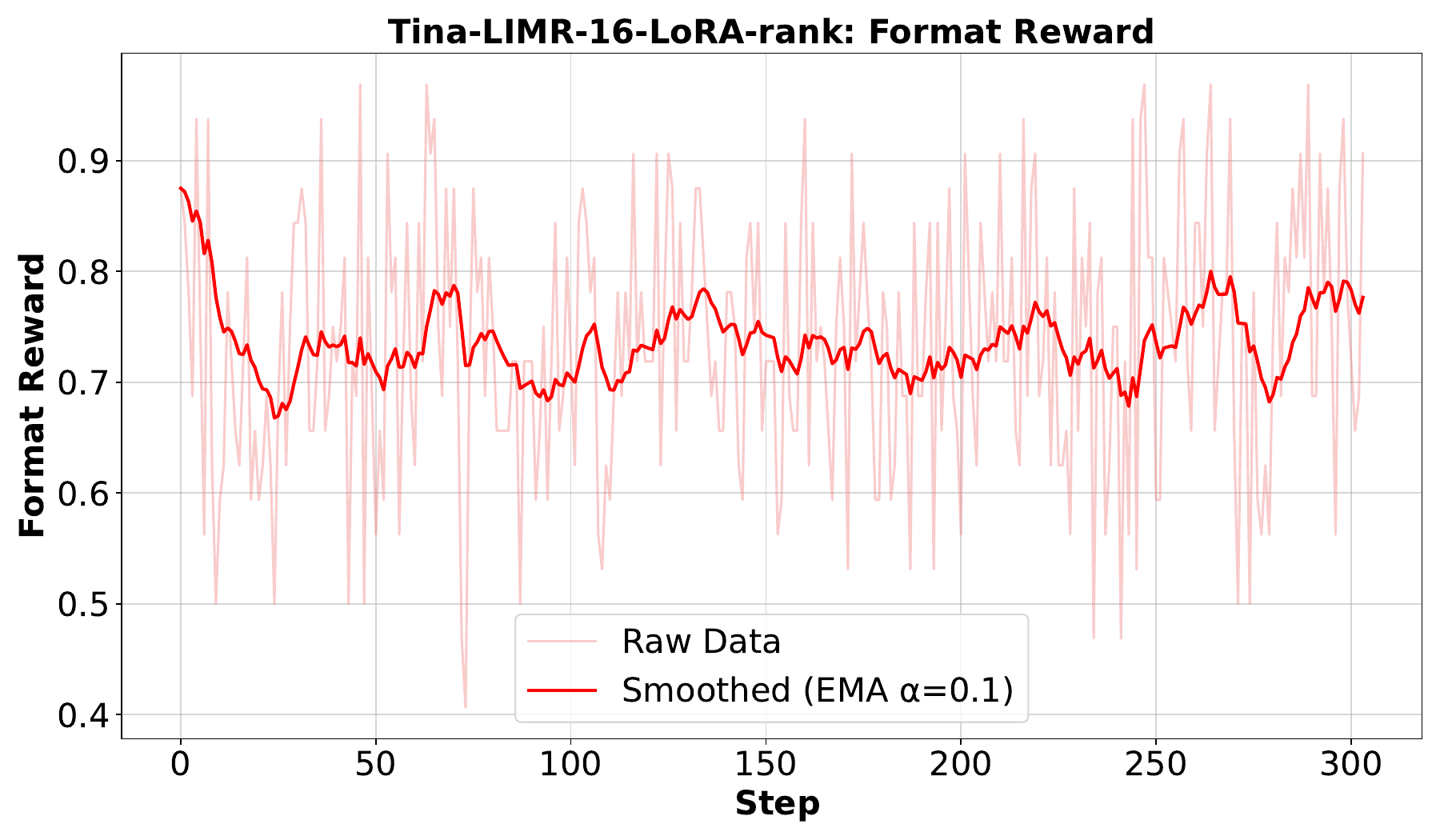}
    \includegraphics[width=.32\linewidth]{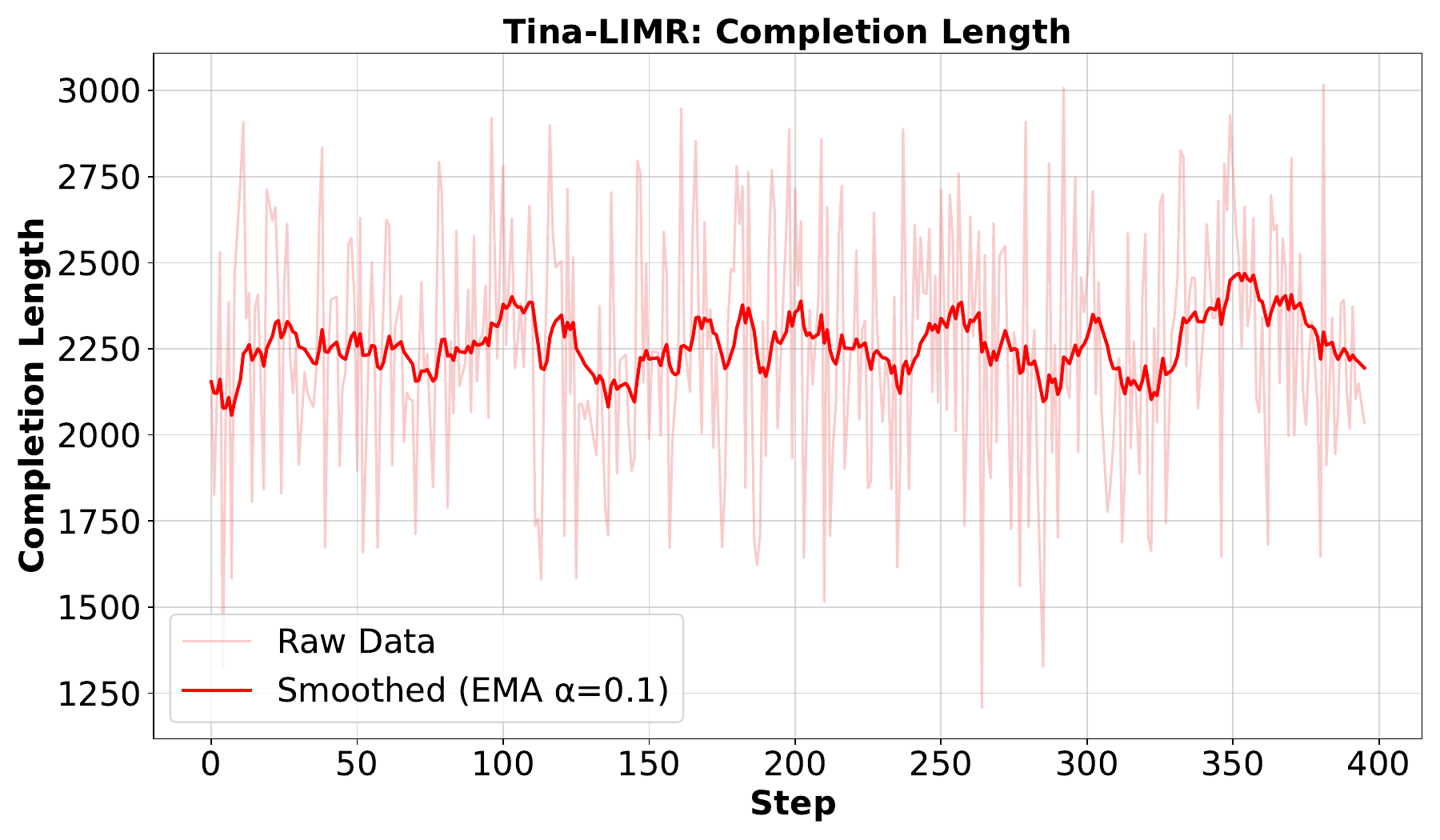}
    \includegraphics[width=.32\linewidth]{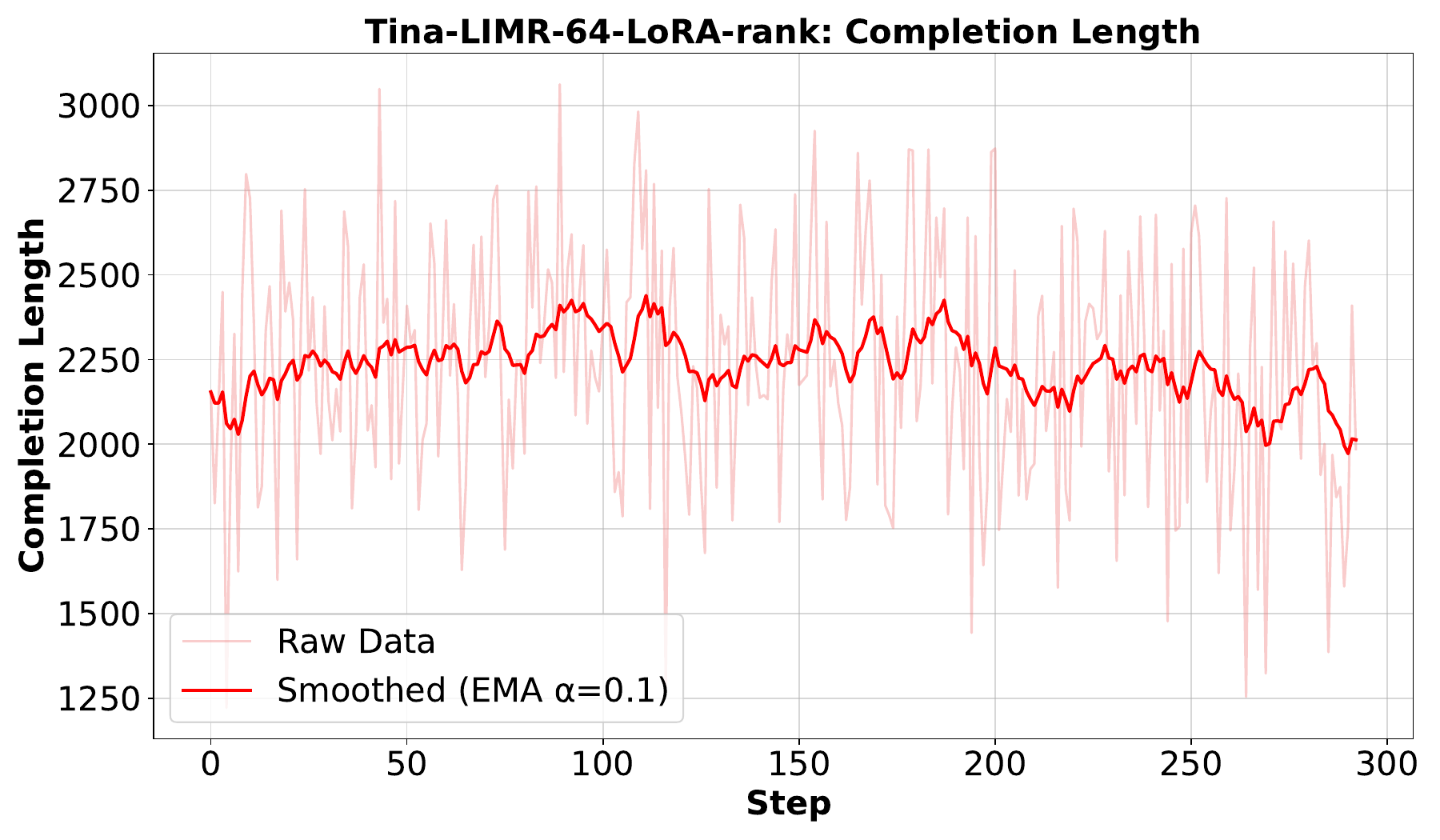}
    \includegraphics[width=.32\linewidth]{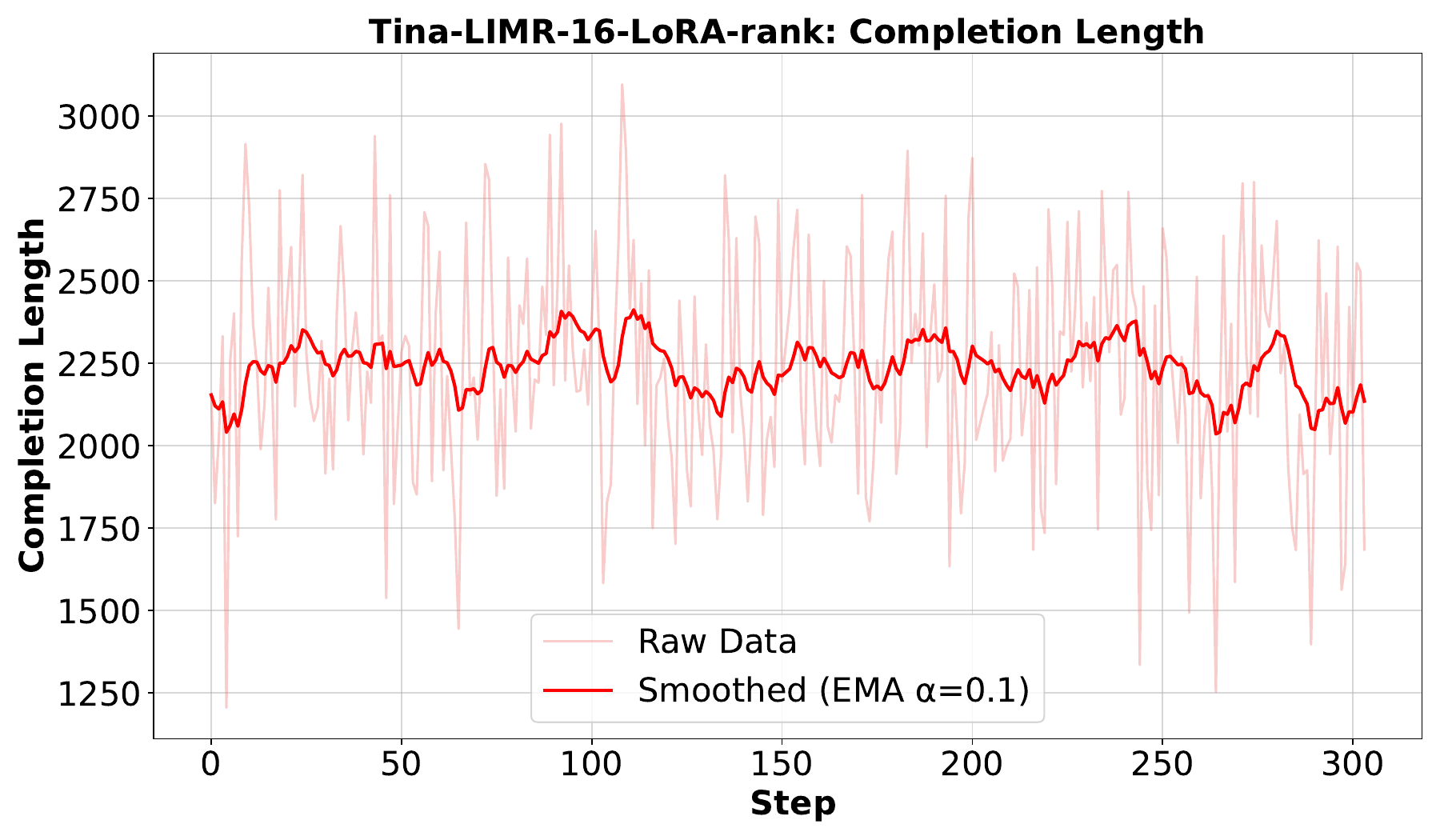}
    \caption{\textbf{Phase transition in \texttt{Tina-LIMR}, \texttt{Tina-LIMR-64-LoRA-rank} and \texttt{Tina-LIMR-16-LoRA-rank}.} The raw data is from the Weights \& Biases training logs and smoothed via exponential moving average (EMA) with factor $0.1$.}
    \label{fig:full_phase_transit_6}
\end{figure}

\begin{figure}[h!]
    \centering
    \includegraphics[width=.45\linewidth]{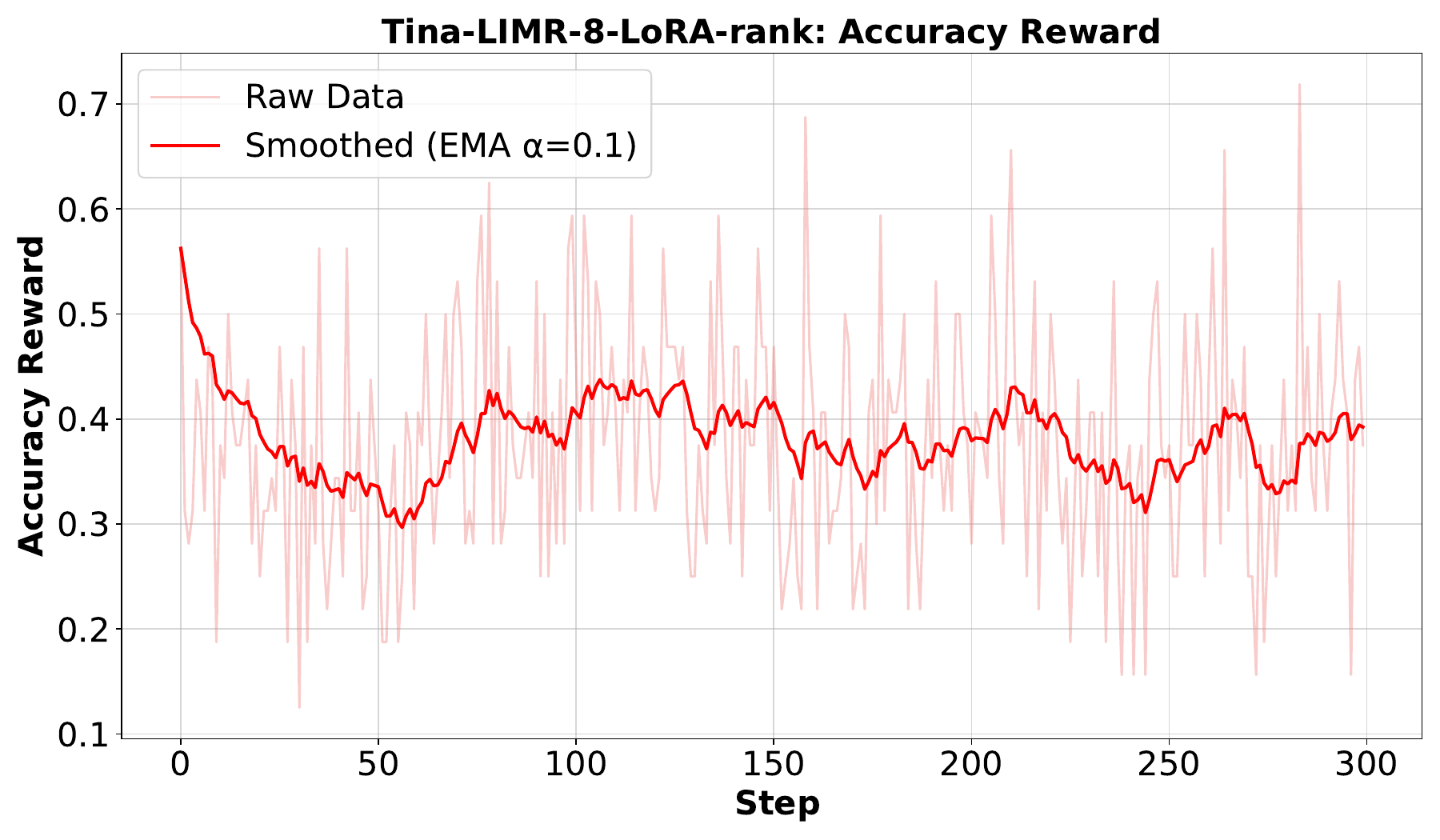}
    \includegraphics[width=.45\linewidth]{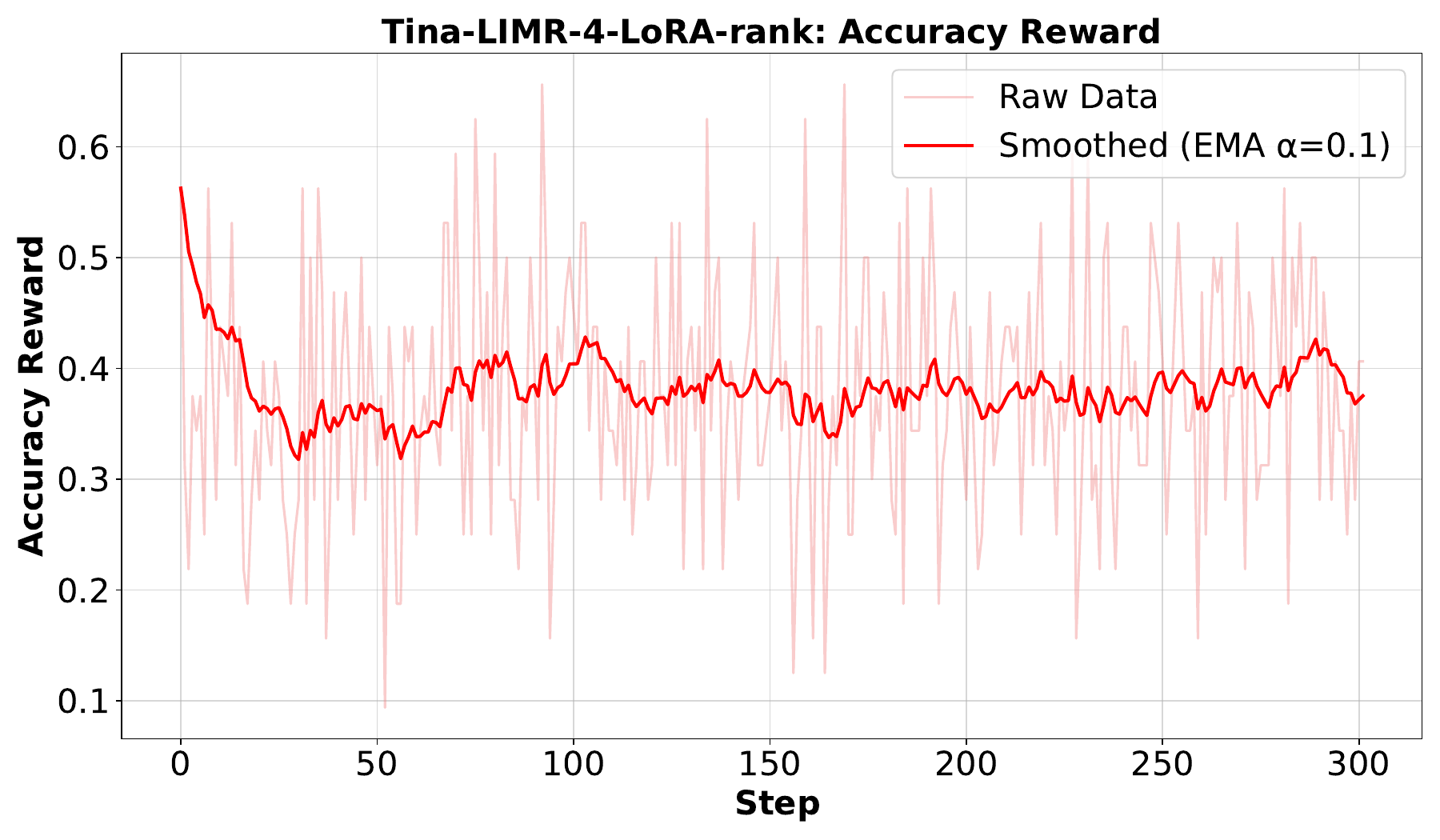}
    \includegraphics[width=.45\linewidth]{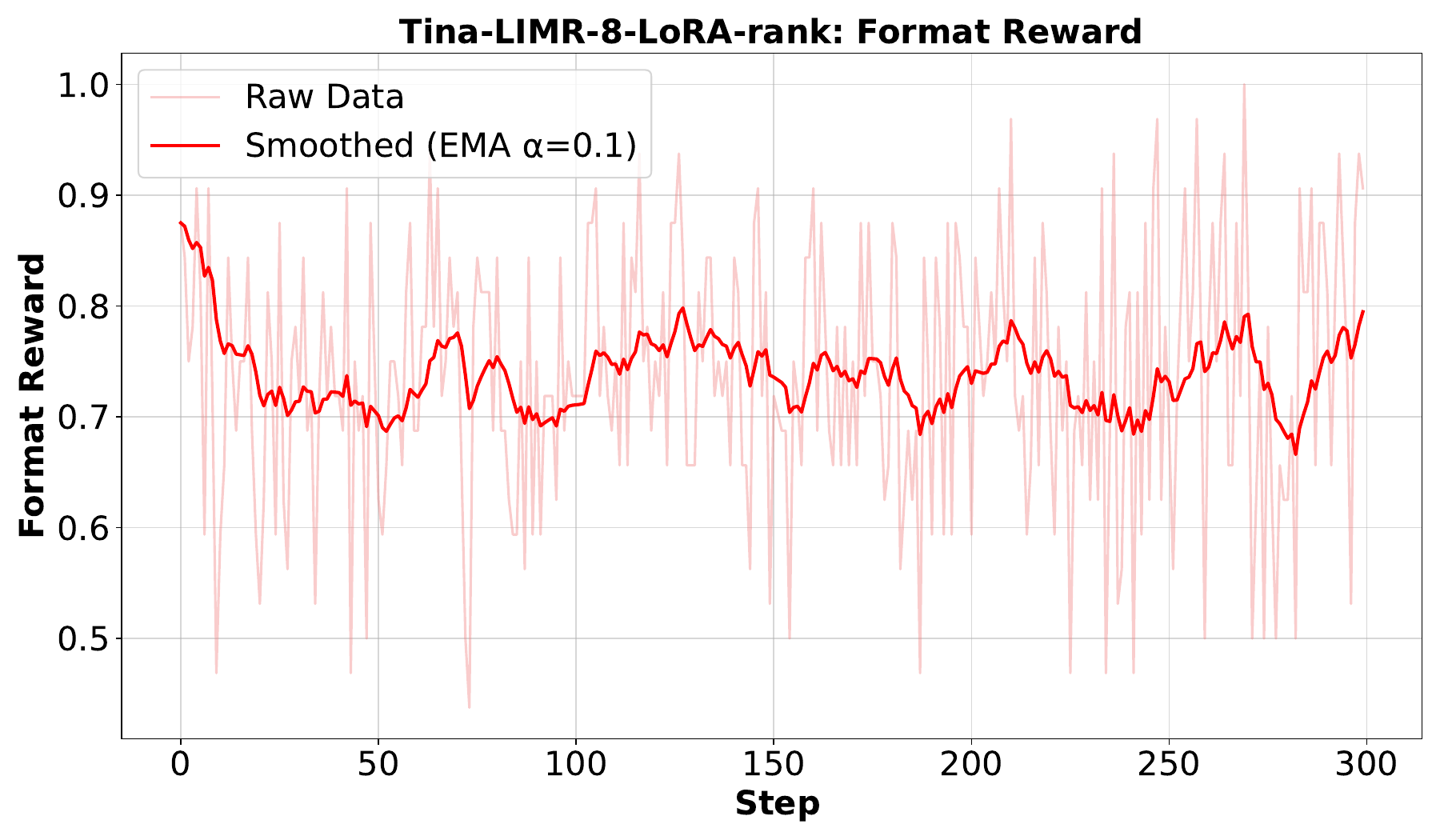}
    \includegraphics[width=.45\linewidth]{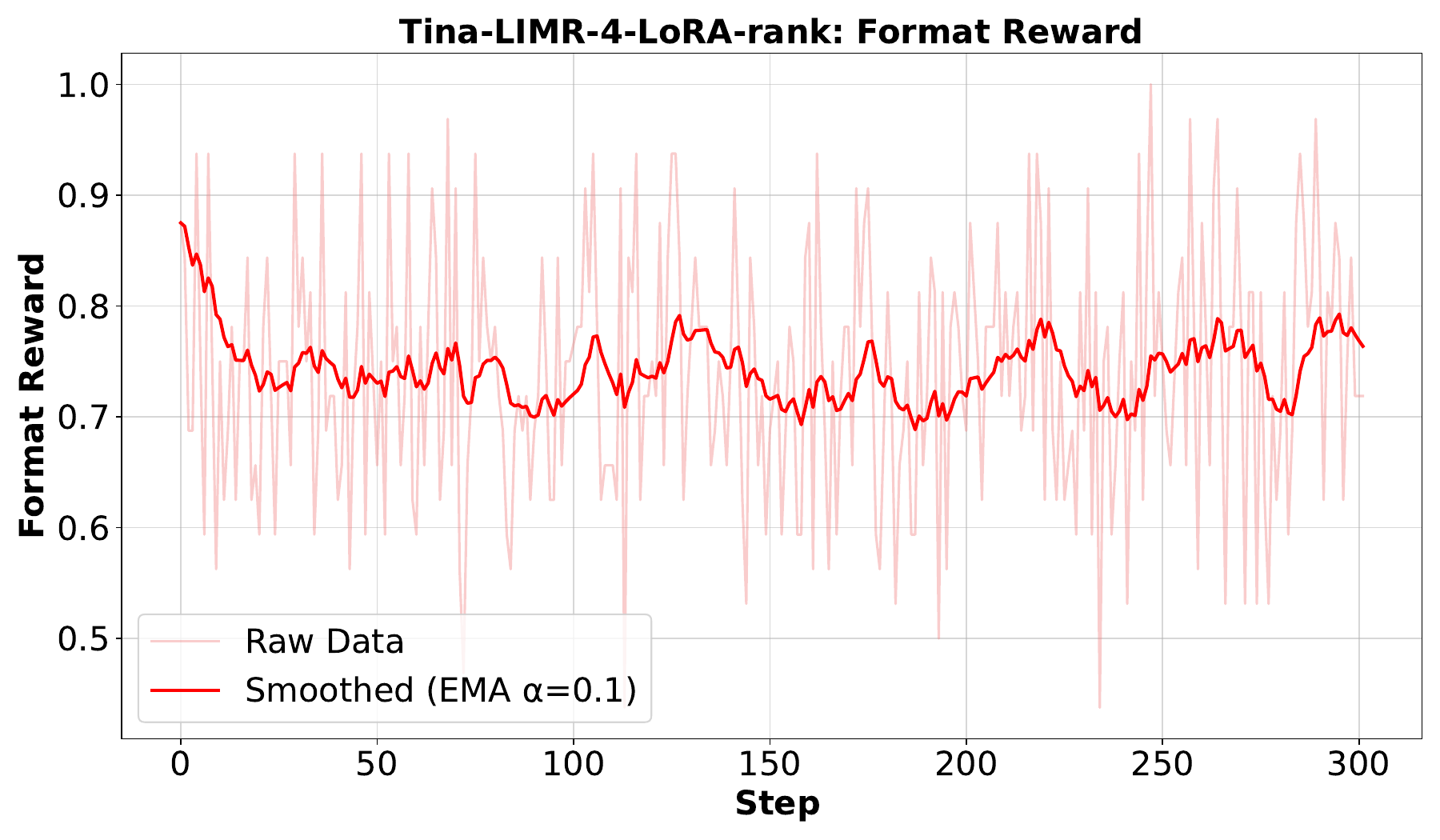}
    \includegraphics[width=.45\linewidth]{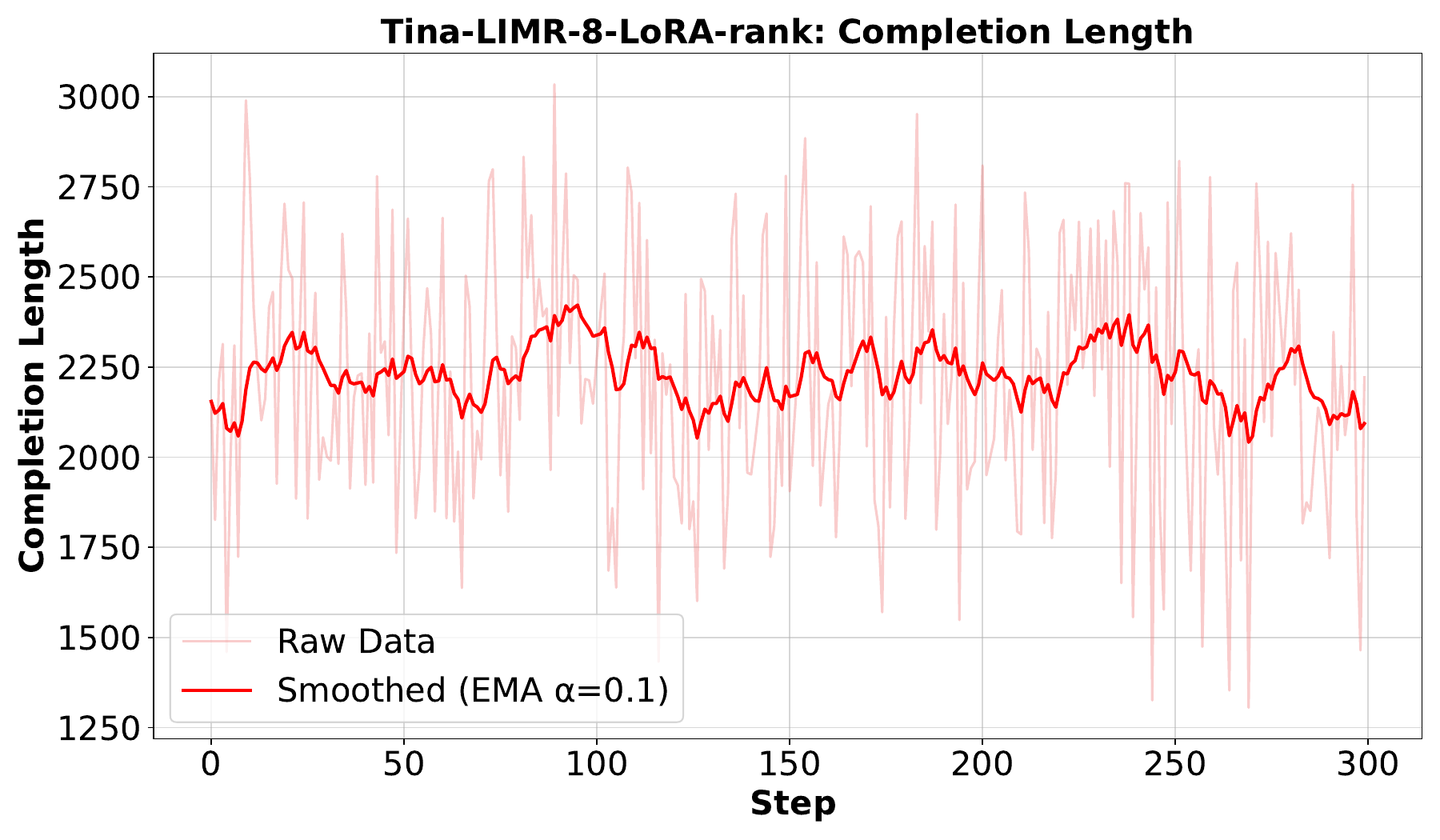}
    \includegraphics[width=.45\linewidth]{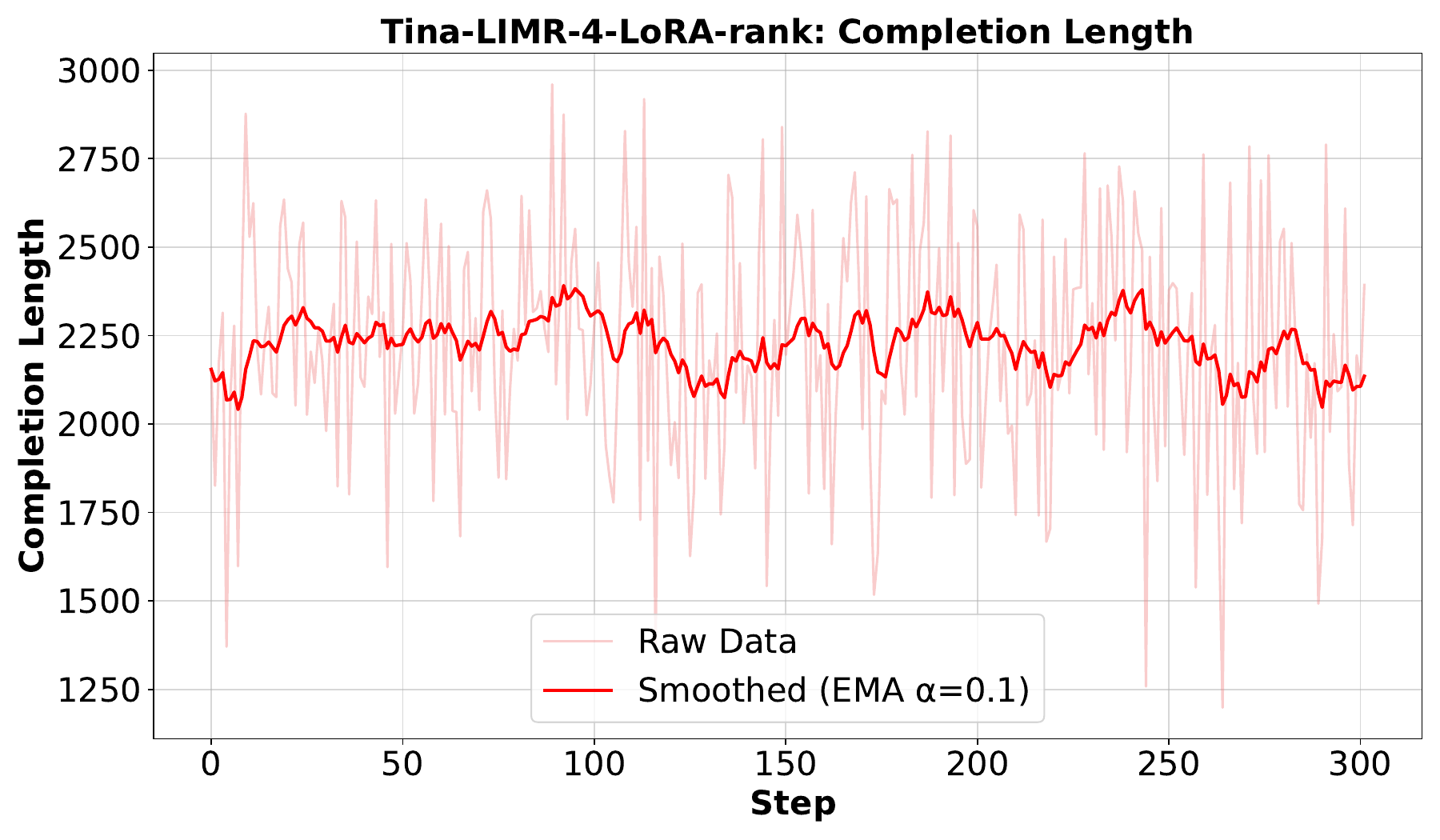}
    \caption{\textbf{Phase transition in \texttt{Tina-LIMR-8-LoRA-rank} and \texttt{Tina-LIMR-4-LoRA-rank}.} The raw data is from the Weights \& Biases training logs and smoothed via exponential moving average (EMA) with factor $0.1$.}
    \label{fig:full_phase_transit_7}
\end{figure}

\begin{figure}[h!]
    \centering
    \includegraphics[width=.45\linewidth]{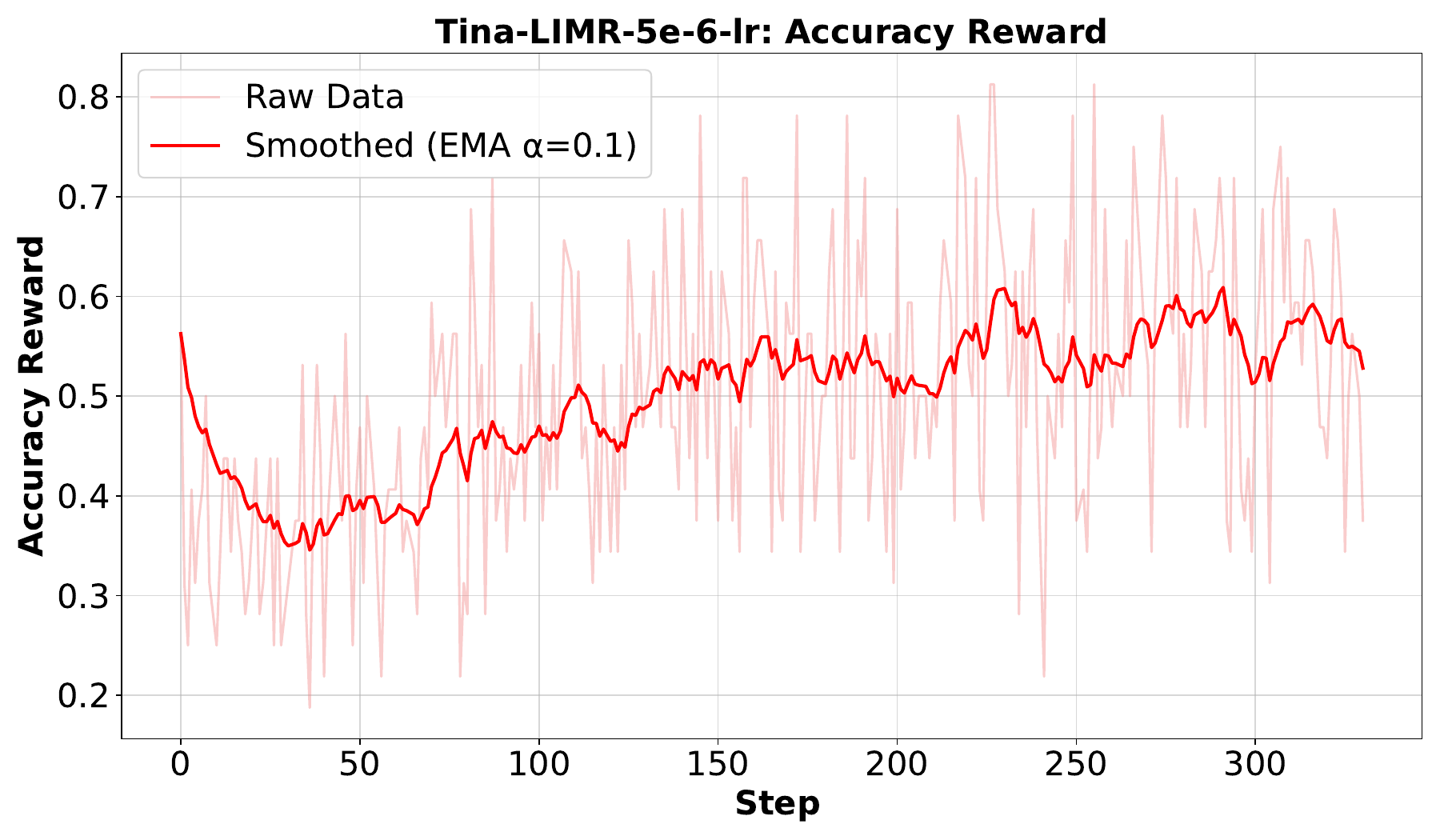}
    \includegraphics[width=.45\linewidth]{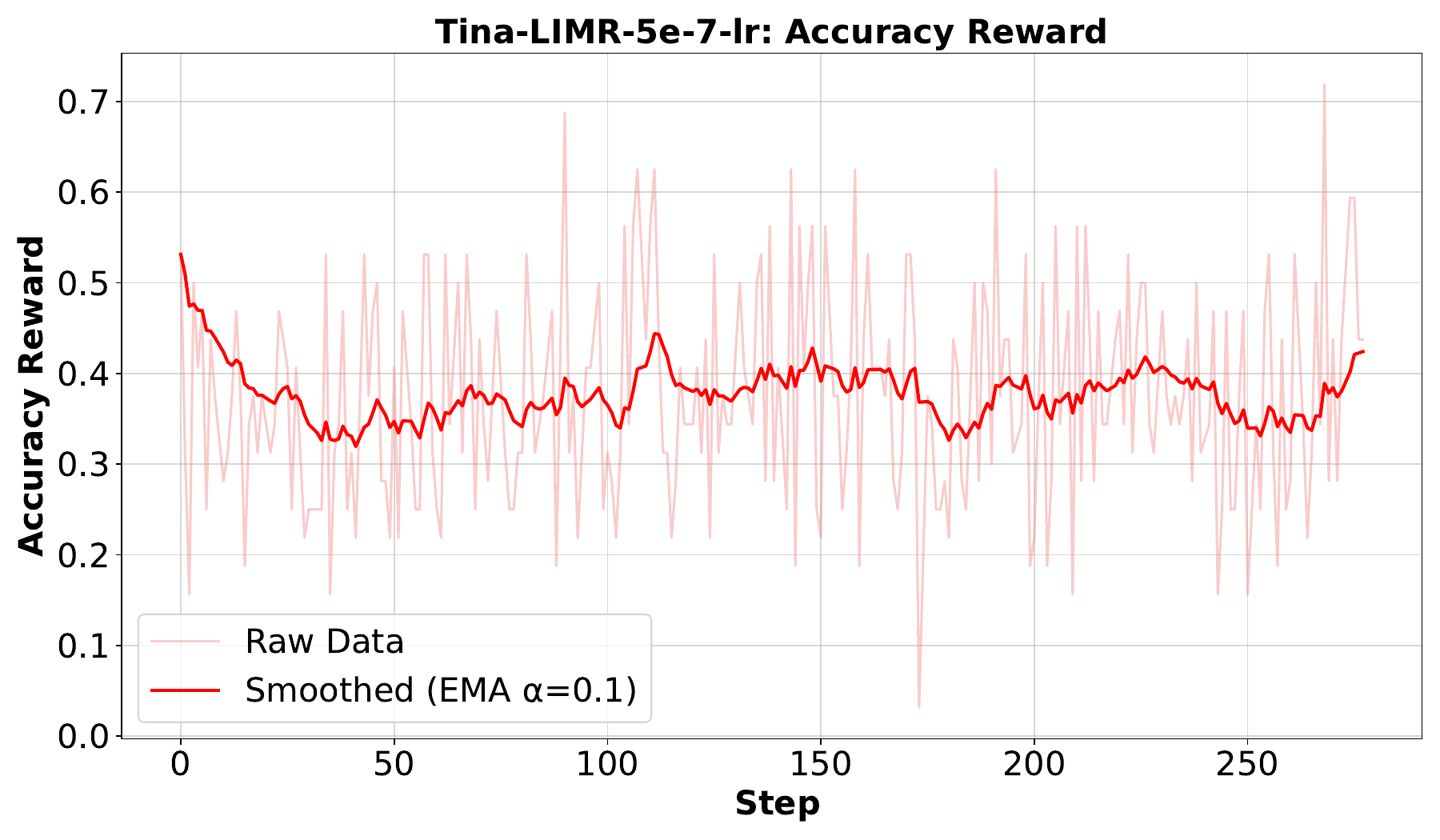}
    \includegraphics[width=.45\linewidth]{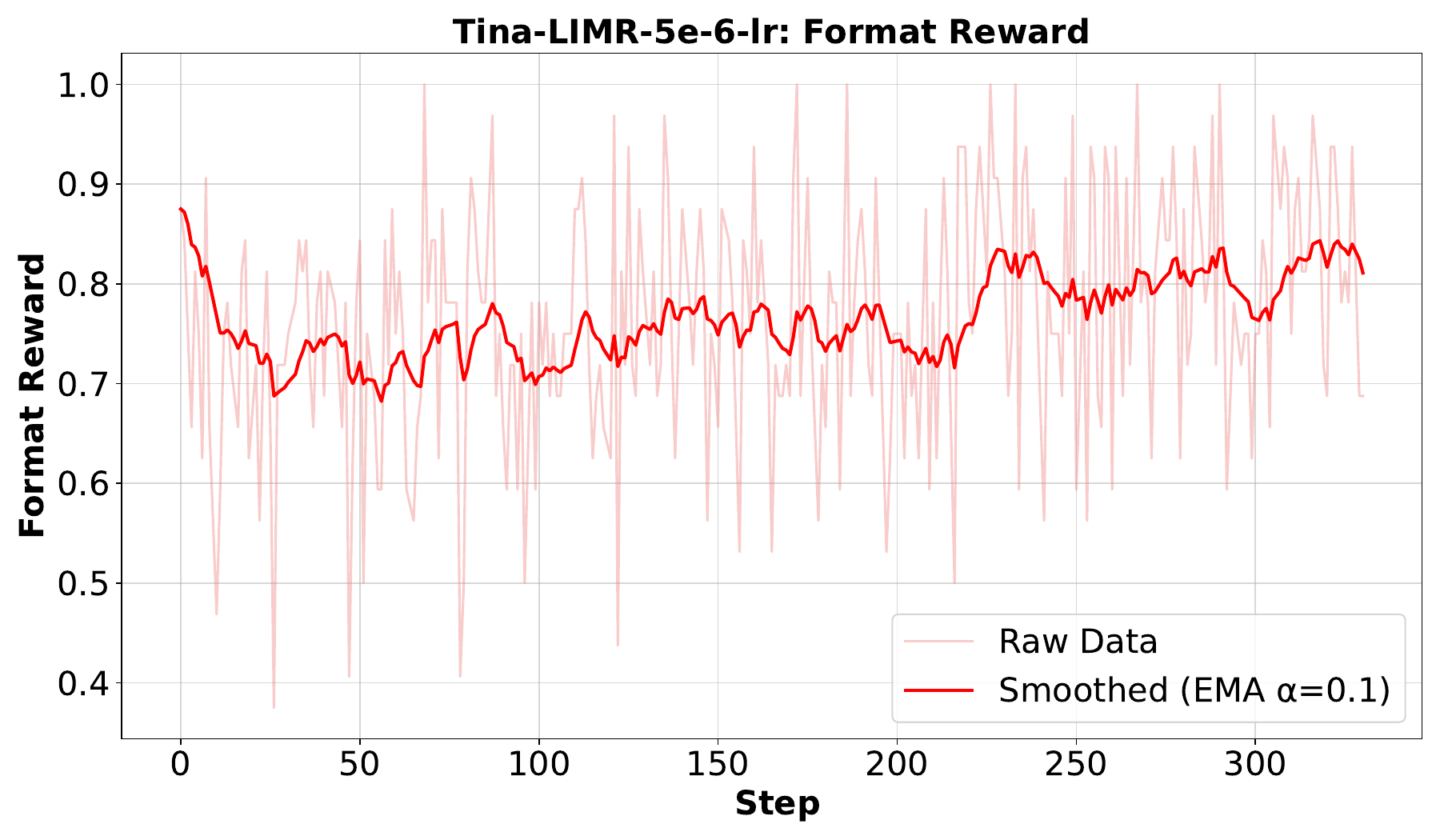}
    \includegraphics[width=.45\linewidth]{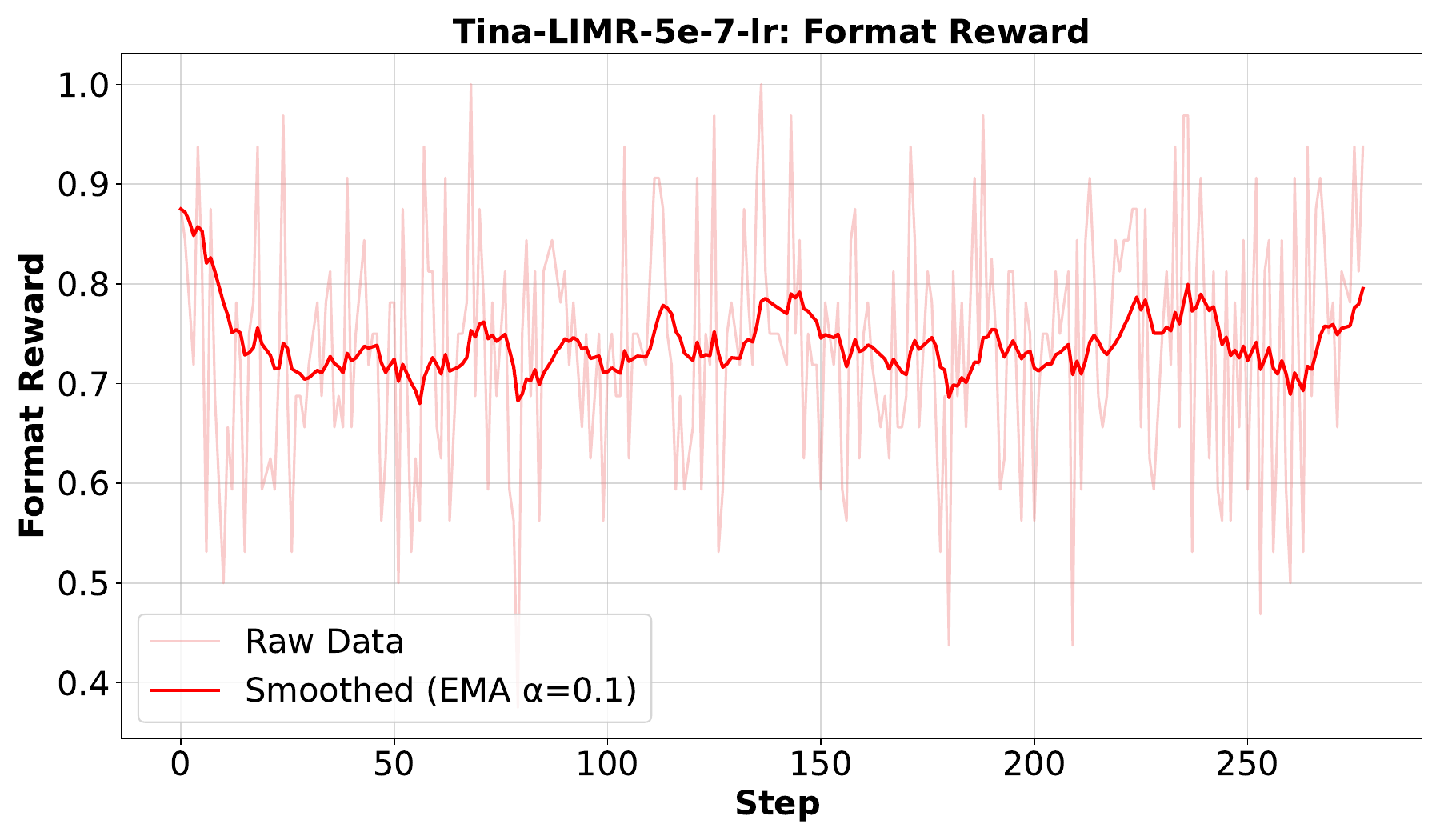}
    \includegraphics[width=.45\linewidth]{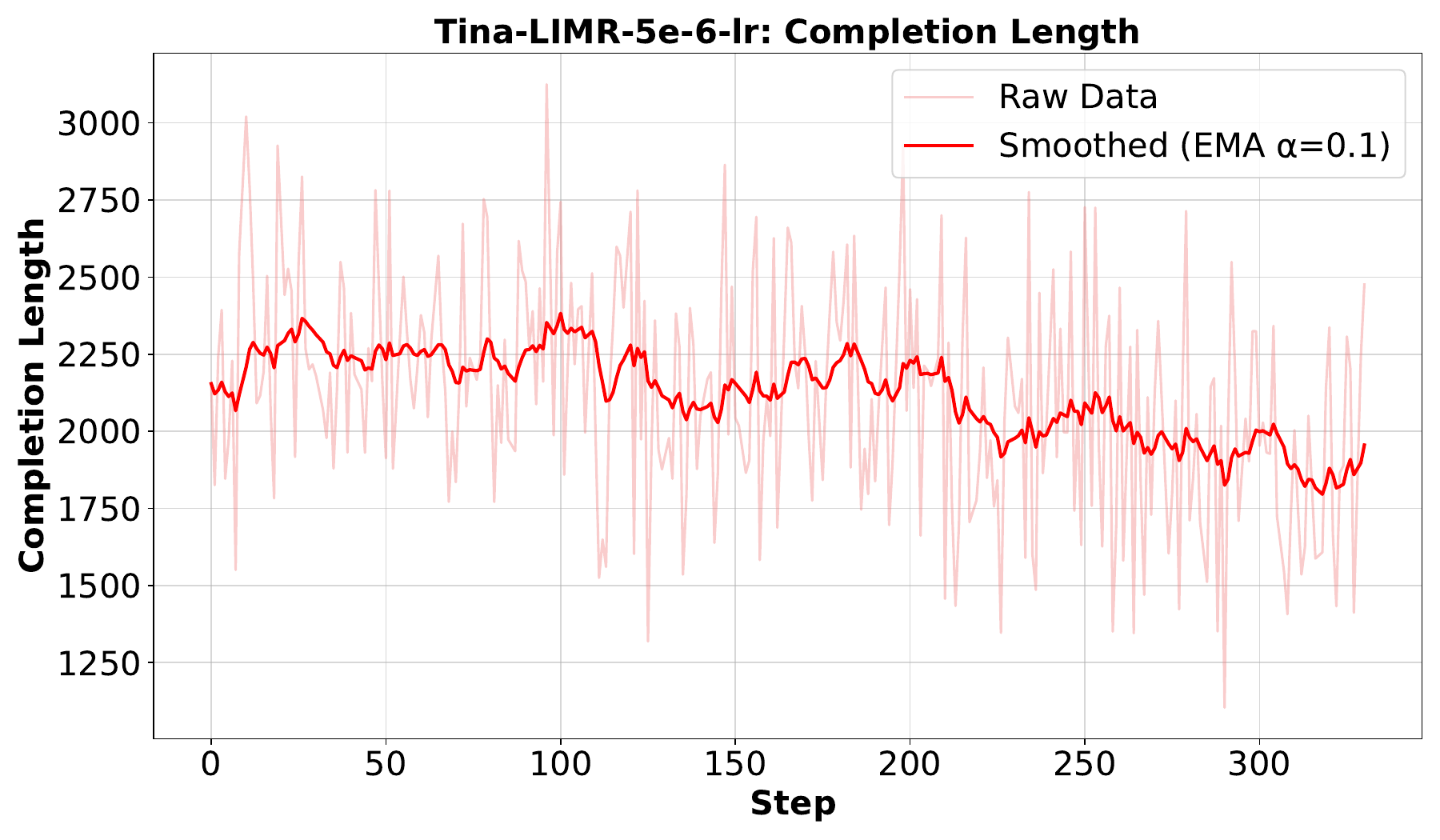}
    \includegraphics[width=.45\linewidth]{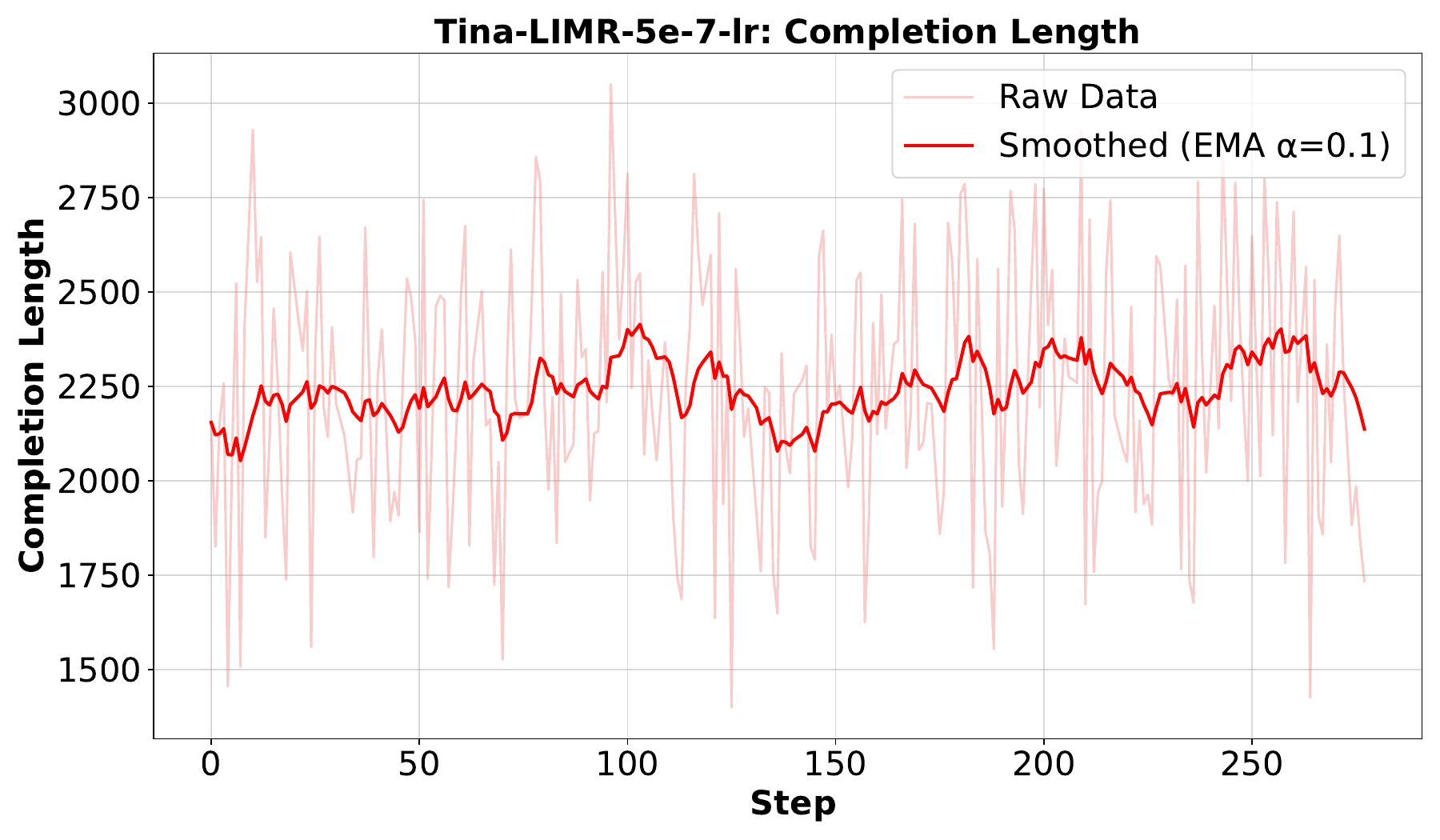}
    \caption{\textbf{Phase transition in \texttt{Tina-LIMR-5e-6-lr} and \texttt{Tina-LIMR-5e-7-lr}.} The raw data is from the Weights \& Biases training logs and smoothed via exponential moving average (EMA) with factor $0.1$.}
    \label{fig:full_phase_transit_8}
\end{figure}

In this section, we present all Tina models' training phase transitions along the training dynamics. Specifically, we observe clear phase transitions in the training of \texttt{Tina-DeepScaleR-1.5B-Preview}, \linebreak \texttt{Tina-STILL-3-1.5B-preview}, \texttt{Tina-Open-RS1}, \texttt{Tina-Open-RS2}, \texttt{Tina-Open-RS3}, and \linebreak \texttt{Tina-Open-RS3-GRPO}, as shown in Figures~\ref{fig:full_phase_transit_1},~\ref{fig:full_phase_transit_2}, and~\ref{fig:full_phase_transit_3}. 
For \texttt{Tina-OpenR1} and \texttt{Tina-Thoughts} (Figures~\ref{fig:full_phase_transit_4} and~\ref{fig:full_phase_transit_5}), the observation is similar, except the best-performing checkpoint is achieved after the training turning point, rather than before. However, we do not observe such a transition in all Tina variants on the LIMR dataset, as shown in Figures~\ref{fig:full_phase_transit_6},~\ref{fig:full_phase_transit_7}, and~\ref{fig:full_phase_transit_8}, possibly because its small data size leads to training periods which are too brief to extract meaningful information. 

\appendix
\end{document}